%% file: paper.tex
\documentclass{article}
\usepackage{ifthen}         
\newboolean{anonymous}
\setboolean{anonymous}{false}  
\newboolean{appendix} 
\setboolean{appendix}{true}  

\newboolean{showframe}
\setboolean{showframe}{true}  
\ifthenelse{\boolean{showframe}}{
    \overfullrule=5pt   
    \hfuzz=0pt          
}{}

\usepackage{ijcai26}

\input{packages.tex}
\input{commands.tex}

\newcommand{\AppendixOrPublic}[2]{\ifthenelse{\boolean{appendix}}{#1}{#2}}
\newcommand{\ProofAppendixPointer}{%
    \AppendixOrPublic
    {We give all proofs in Appendix~\ref{sec:appendix_proofs}. }
    {We give all proofs in the Appendix~\cite{LevYehudi26arxiv}. }%
}
\newcommand{\MisTreeAppendixDetails}{%
    \AppendixOrPublic
    {We give full MIS Tree derivations, numerically stable log-likelihood forms, and analogous immediate-reward updates in Appendix~\ref{sec:appendix_mis_tree_procedures}.}
    {We give full MIS Tree derivations, numerically stable log-likelihood forms, and analogous immediate-reward updates in the Appendix.}%
}
\newcommand{\AreaFormulaAppendixDetails}{%
    \AppendixOrPublic
    {We further discuss these assumptions in Appendix~\ref{sec:appendix_area_formula_assumptions}. }
    {We further discuss these assumptions in the Appendix. }%
}

\ifthenelse{\boolean{anonymous}}{\linenumbers}{}

\allowdisplaybreaks[4]

\title{Action-Gradient Monte Carlo Tree Search for \\ Non-Parametric Continuous (PO)MDPs}

\ifthenelse{\boolean{anonymous}}
{
    \author
    {
        Anonymous Authors
        \\
        \affiliations
        Anonymous Affiliations
    }
}
{
    \author
    {
        Idan Lev-Yehudi$^1$\and
        Michael Novitsky$^1$\and
        Moran Barenboim$^1$\and
        Ron Benchetrit$^2$\And
        Vadim Indelman$^{3,4}$
        \\
        \affiliations
        $^1$Technion Autonomous Systems Program (TASP), Technion {--} Israel Institute of Technology\\
        $^2$Faculty of Computer Science, Technion {--} Israel Institute of Technology\\
        $^3$Stephen B. Klein Faculty of Aerospace Engineering, Technion {--} Israel Institute of Technology\\
        $^4$Faculty of Data and Decision Sciences, Technion {--} Israel Institute of Technology\\
        \emails
        idanlev@campus.technion.ac.il,
        vadim.indelman@technion.ac.il,
    }
}

\begin{document}

\maketitle

\begin{abstract}
Online planning in continuous state, action, and observation spaces remains challenging for autonomous systems. While Monte Carlo Tree Search (MCTS) scales effectively via sampling, most continuous (PO)MDP solvers do not exploit gradient-based action optimization. We propose Action-Gradient MCTS (AGMCTS), a framework that combines global tree search with local gradient-based action refinement, while maintaining consistent value estimates. We provide three key theoretical contributions: (1) an action score gradient theorem for particle belief states; (2) the Multiple Importance Sampling (MIS) Tree that supports frequent action-branch updates by reusing prior samples without introducing estimator drift; and (3) tractable action score gradients for smooth generative models using the Area Formula. Empirical results demonstrate that AGMCTS outperforms state-of-the-art sample-based solvers in multiple challenging continuous MDP and POMDP benchmarks.
\end{abstract}

\section{Introduction}

Planning under uncertainty is central to AI and robotics, where state, action, and observation spaces are often continuous~\cite{Lauri22tro}.
Markov Decision Processes (MDPs) and Partially Observable Markov Decision Processes (POMDPs) model such problems, but exact solutions are generally intractable~\cite{Papadimitriou87math,Madani03AI}, motivating efficient approximation.

Online solvers compute actions on demand from the current state or belief~\cite{Kurniawati22ar}, often via tree search such as Monte Carlo Tree Search (MCTS)~\cite{Silver10nips} and Determinized Sparse Partially Observable Tree (DESPOT)~\cite{Ye17jair}.
Earlier continuous-POMDP offline methods often used fixed-dimensional parametric beliefs, e.g., Gaussian beliefs~\cite{Porta06jmlr}; recent continuous MDP/POMDP online methods improve scalability via progressive widening, including Double Progressive Widening (DPW) for MDPs and POMCPOW/PFT-DPW for POMDPs~\cite{Couetoux11iclio,Sunberg18icaps}.

Most sampling-based continuous POMDP planners keep sampled actions fixed.
VG-UCT~\cite{Lee20aaai} refines continuous-MDP actions, but does not correct value-estimate drift after action updates.
Action-Gradient MCTS (AGMCTS) integrates gradient steps into global MCTS via MIS, and is the first such hybrid strategy for non-parametric continuous POMDPs with sample-based beliefs.

\subsection{Contributions}

\paragraph{Theoretical.} We derive an action score gradient theorem for MDPs and POMDPs in online tree search with importance weighting.
We introduce the Multiple Importance Sampling (MIS) Tree and its \emph{action update} operation, allowing frequent gradient steps while maintaining value-estimate consistency.
We also compute action score gradients tractably for smooth generative models via the Area Formula.

\paragraph{Algorithmic.} We present AGMCTS (Algorithm~\ref{alg:AGMCTS}), blending MCTS action exploration with online gradient refinement for MDPs/POMDPs under DPW or alternative continuous-action selection heuristics.

\paragraph{Experimental.} Across continuous MDP and POMDP benchmarks, AGMCTS variants show notable gains in most domains and competitive performance overall.

\subsection{Related Work}

\paragraph{Online Optimization in MCTS.} \citeauthor{Lee20aaai}~[\citeyear{Lee20aaai}] have proposed VG-UCT for integrating gradient steps into MCTS in continuous MDPs.
Their Lipschitz-continuous transition and reward setting gives a deterministic-MDP convergence result; we give estimator gradients under weaker assumptions and address value-estimate drift from action updates.

Black-box optimization methods have been used to bias action selection in MCTS to more promising regions, e.g.
Voronoi optimistic optimization (VOO)~\cite{Kim20aaai} and Voronoi progressive widening (VPW)~\cite{Lim21cdc,Hoerger24ijrr}, kernel regression~\cite{Yee16ijcai} and Bayesian optimization~\cite{Morere18iros,Mern21aaai} methods.
These complement AGMCTS by improving action proposals rather than refinements.
 
\paragraph{(PO)MDP Action-Gradients.} An action-gradient corresponds to the single-step gradient of a policy with a deterministic immediate action.
Policy gradients in POMDPs typically assume stochastic policies~\cite{Baxter01jair}, memory-less policies~\cite{Azizzadenesheli18arxiv}, or finite action spaces~\cite{Hong24iclr}, while we handle belief-dependent policies with continuous actions.
\citeauthor{Silver14icml}~[\citeyear{Silver14icml}] studied deterministic policies in MDPs; we extend to non-parametric POMDPs.

\paragraph{Information Reuse in Planning.} \citeauthor{Leurent20acml}~[\citeyear{Leurent20acml}] considered information sharing in discrete MDPs by merging similar states.
\citeauthor{Novitsky25ral}~[\citeyear{Novitsky25ral}] have introduced previous knowledge utilization in POMDPs via Multiple Importance Sampling (MIS)~\cite{Veach95siggraph}.
We reuse samples across sibling action branches via the MIS Tree, avoiding retrieval from an external belief database.
POMDP planning has used MIS for sample reuse, e.g.~\cite{Farhi21arxiv}, mainly to compute observation likelihoods.
AdaOps and CMCGS cluster values of similar observations/states~\cite{Wu21nips,Kujanpaa24aamas}, whereas we recompute value estimates under updated actions.

\section{Background}
\label{sec:background}

\paragraph{MDPs and POMDPs.} We consider MDPs in the form $\langle \mathcal{S},\mathcal{A},p_{T},r,\gamma,L,b_0\rangle$.
The state and action spaces are $\mathcal{S}\subseteq\mathbb{R}^{n_s}$ and $\mathcal{A}\subseteq\mathbb{R}^{n_a}$.
The transition model $p_{T}(s^{\prime}{\mid} s,a)$ is the probability of arriving at state $s^{\prime}$ when taking action $a\in\mathcal{A}$ at state $s\in\mathcal{S}$.
The reward function $r(s,a,s^\prime)\in\mathbb{R}$ gives the immediate reward of transitioning from state $s$ to $s^\prime$ by taking action $a$, and the expected reward is $r(s,a)\bydef \mathbb{E}_{s^\prime{\mid} s,a}[r(s,a,s^\prime)]$.
The discount factor is $\gamma\in(0,1]$.
The MDP starts at time $0$ and terminates after $L\in\mathbb{N}\cup\{\infty\}$ steps, and if $L=\infty$ then we assume $\gamma < 1$.
The initial state is drawn from the distribution $s_0 \sim b_0$.

We assume $\pi=(\pi_t)_{t=0}^{L}$ is a (possibly time-dependent) deterministic policy, but we note our theoretical results generalize to stochastic policies as well.
The value function of a policy $\pi$ at time $t$ is the expected sum of discounted rewards until the horizon: $V_{t}^{\pi}(s_{t})\bydef \ExptFlat{s_{t+1:L}{\mid} s_t,\pi}{}{\sum_{i=t}^{L}\gamma^{i-t}r(s_{i},\pi_i(s_{i}),s_{i+1})}$. 
We denote next-step variables with primes, e.g. $s^\prime$ follows $s$.
We define the action-value function as $Q_{t}^{\pi}(s,a)\bydef \mathbb{E}_{s^\prime{\mid} s,a}[r(s,a,s^\prime)+\gamma V_{t+1}^{\pi}(s^\prime)]$.
Our goal is to compute a policy that maximizes the value function, i.e. find $\pi^{*}\in\argmax_{\pi} V_{0}^{\pi}(s_0)$.

A POMDP adds $\langle \mathcal{O},p_{O}\rangle$ to the MDP tuple.
The observation space is $\mathcal{O}\subseteq\mathbb{R}^{n_o}$, and the observation model $p_{O}(o{\mid} s)$ is the conditional probability of receiving an observation $o\in\mathcal{O}$ at $s\in\mathcal{S}$. Similarly to~\citeauthor{Sunberg18icaps}~[\citeyear{Sunberg18icaps}], we assume that we can both sample from and evaluate $p_{O}(o{\mid} s)$.

A history at time $t$ is defined as a sequence of $b_0$, followed by actions taken and observations received until $t$: $H_{t}\bydef(b_{0},a_{0},o_{1},\dots,a_{t-1},o_{t})$.
In partially observable settings, the agent has to maintain a probability distribution of the current state given past actions and observations, known as the belief.
The belief at time $t$ is $b_{t}(s_{t})\bydef p(s_{t}{\mid} H_{t})$.
We define $H_{t}^{-}\bydef(b_{0},a_{0},o_{1},\dots,a_{t-1})$, i.e. the history until time $t$ without the last measurement, and correspondingly the propagated belief $b_{t}^{-}(s_{t})\bydef p(s_{t}{\mid} H_{t}^{-})$.
It has been shown that optimal decision-making can be made given the belief, instead of considering the entire history~\cite{Kaelbling98ai}, meaning a POMDP is an MDP on the belief space~\cite{Aastrom65jmaa}.
Policies and value functions extend from MDPs to POMDPs by equivalent definitions on beliefs as states.

\paragraph{Importance Sampling.} Importance Sampling (IS)~\cite{Kloek78econometrica} is a technique in Monte-Carlo (MC) estimation where a proposal distribution $q$ is used to generate samples. The IS estimator for $g(x)=\mathbb{E}_{x\sim p}[f(x)]$ is $\hat{g}_{q}=N^{-1} \sum_{i=1}^{N}\rho^{p}_{q}(x^i)f(x^{i})$, for importance ratios $\rho^{p}_{q} (x) \bydef p(x) \slash q(x)$.
The IS estimator $\hat{g}_{q}$ is unbiased if $q(x)=0$ implies $p(x)=0$.
Often, the self-normalized IS (SNIS) estimator is formulated when we only have access to an unnormalized version of $p$, or for variance reduction purposes, and is given by $\tilde{g}_{q}=(\sum_{i=1}^{N}\rho^{p}_{q}(x^i))^{-1} \sum_{i=1}^{N}\rho^{p}_{q}(x^i)f(x^{i})$.
While biased, under weak assumptions it is consistent~\cite[9.2]{Owen13book}.
We denote SNIS estimators throughout with $\tilde{[\cdot]}$.

MIS~\cite{Veach95siggraph} is a method to leverage several sampling methods $\{q_i\}_{i=1}^{n}$, by calculating a weighted average of IS estimators with $n_i$ samples each:
The corresponding MIS estimator is $\hat{g}_{\text{MIS}}=\sum_{i=1}^{n} n_i^{-1}\sum_{j=1}^{n_i}w_i(x^{i,j})\rho^{p}_{q_i}(x^{i,j})f(x^{i,j})$.
If the weights $w_i$ satisfy requirements: \hypertarget{assum:mis-i}{(MIS-I)} $\sum_{i=1}^{n}w_i(x)=1$ whenever $f(x)\neq 0$; \hypertarget{assum:mis-ii}{(MIS-II)} $w_i(x)=0$ whenever $q_i(x)=0$; then $\hat{g}_{\text{MIS}}$ is unbiased.
Many weighting strategies exist; the balance heuristic $w_i(x)=n_i q_i(x)\slash \sum_k n_k q_k(x)$, has its variance provably bounded from the optimal weighting strategy for independent samples.
Analogously to SNIS, self-normalized MIS (SN-MIS) can be defined as $\tilde{g}_{\text{MIS}}=\beta^{-1} \sum_{i=1}^{n} \sum_{j=1}^{n_i}w_i(x^{i,j})\rho^{p}_{q_i}(x^{i,j})f(x^{i,j})$ for $\beta = \sum_{i=1}^{n}\sum_{j=1}^{n_i}w_i(x^{i,j})\rho^{p}_{q_i}(x^{i,j})$~\cite{Metelli20jmlr}.

\paragraph{Particle Beliefs.} Particle filters are often used to represent a belief non-parametrically~\cite[1.3.2]{Doucet00}.
We denote particle beliefs with $\bar{b}$.
The particle belief of $J$ particles is the ordered set of state-weight pairs $\bar{b} =((s^{j},\lambda^{j}))_{j=1}^{J}$ (following~\cite{Lim23jair}), and is defined as the discrete distribution $p(s{\mid} \bar{b})={\sum_{j=1}^{J}{\lambda^{j}\cdotp \delta(s-s^{j})}}\slash {\sum_{j=1}^{J}{\lambda^{j}}}$.
For computational simplicity, we assume throughout the paper that the bootstrap filter is used to update particle beliefs~\cite[1.3.2]{Doucet00}.
At each time step, the bootstrap filter advances sampled particles using the transition model, reweights them based on the observation model, and resamples to avoid weight degeneracy~\cite{Kong94}.

\paragraph{MCTS and DPW.} MCTS is an algorithm used to quickly explore large state spaces~\cite{Browne12ieee}. It iteratively repeats four steps to build a search tree that approximates the action-values, using a best-first strategy:
(1) \textit{Selection}: Starting from the root node, descend recursively until a node to which children can be added is found;
(2) \textit{Expansion}: A new node is added as a child, according to an action expansion strategy;
(3) \textit{Simulation}: A simulation is run from the new node according to the rollout (default) policy;
(4) \textit{Backpropagation}: The simulation result is backpropagated through the selected nodes to update the action-values.
Often UCT~\cite{Kocsis06ecml} is used at \textit{selection} to balance between exploration of new actions and exploitation of promising ones.
Double Progressive Widening (DPW)~\cite{Couetoux11iclio} is a technique to limit the branching factor from being infinite in continuous settings.
The number of children of a node is artificially limited to $kN^\alpha$ for $N$ visitations, for fixed $k>0$ and $0<\alpha<1$.

\section{Action Gradients and MIS Trees}

Here we present the theoretical basis of AGMCTS.
We derive action score gradients for local refinement, introduce MIS Trees for consistent value estimates, and discuss density computation for smooth black-box simulators.

\subsection{(PO)MDP Action Score Gradients}
\label{sec:action_gradients}
In POMDPs the posterior belief transition probability $p(b^{\prime} {\mid} b,a)$ is generally intractable.
We approach the belief update structure via the propagated belief density:
\begin{linenomath*}
    \begin{talign}
        \hspace{-0.08cm} p(b^{\prime} {\mid} b,a) = \int p(b^{\prime} {\mid} b^{\prime -},o^{\prime}) p(o^{\prime} {\mid} b^{\prime -}) p(b^{\prime -} {\mid} b,a) \dif b^{\prime -} \dif o^{\prime}.
        \label{eq:propagated_belief_ratio}
    \end{talign}
\end{linenomath*}
The propagated belief $b^{\prime -}$ is obtained after action $a$ but before observation $o^{\prime}$.
Exact beliefs make $p(b^{\prime -} {\mid} b,a)$ a Dirac delta because $b^{\prime -}$ is deterministic in $(b,a)$; for particle beliefs $\bar{b},\bar{b}^{\prime -}$ it is non-degenerate because particles are sampled through stochastic transitions:
\begin{lem}
    \label{lem:propagated_pb_likelihood}
    Let $\bar{b} =((s^{j},\lambda^{j}))_{j=1}^{J}$, $\bar{b}^{\prime -} =((s^{\prime -,j},\lambda^{\prime -,j}))_{j=1}^{J}$ be ordered particle beliefs. The probability that $\bar{b}^{\prime -}$ is a propagated belief in the bootstrap filter, given $\bar{b}$ and action $a$, is:
    \begin{linenomath*}
        \begin{talign}
            p(\bar{b}^{\prime -}{\mid} \bar{b},a)&=\prod_{j=1}^{J}p_T(s^{\prime -,j}{\mid} s^{j}, a),
            \label{eq:propagated_belief_likelihood}
        \end{talign}
    \end{linenomath*}
    whenever $\lambda^{\prime -,j}=\lambda^{j}$ for all $j=1,\dots,J$, and zero otherwise.
    When $p(\bar{b}^{\prime -}{\mid} \bar{b},a)>0$ it holds that:
    \begin{linenomath*}
        \begin{talign}
            \nabla_a \log p(\bar{b}^{\prime -}{\mid} \bar{b},a)&=\sum_{j=1}^{J} \nabla_a \log p_T(s^{\prime -,j}{\mid} s^{j}, a).
            \label{eq:propagated_belief_grad}
        \end{talign}
    \end{linenomath*}
\end{lem}
We use this to introduce importance ratios over the future propagated belief $\bar{b}^{\prime -}$ rather than the posterior belief $\bar{b}^{\prime}$.

In tree search, after updating action $a$ to $\breve{a}$, we estimate $Q_{t}^{\pi}(s,\breve{a})$ by reusing samples from $Q_{t}^{\pi}(s,a)$ instead of recomputing the subtree.
Because repeated updates create samples from multiple proposals, we use IS/MIS.
Equations \eqref{eq:mdp_reuse}-\eqref{eq:pomdp_reuse} give the single-proposal action-reuse form:
\begin{linenomath*}
    \begin{talign}
        Q_{t}^{\pi}(s,\breve{a}) &= \mathbb{E}_{s^{\prime}{\mid} q}
        [\rho^{p_T}_{q}(s^\prime)
        (r(s,\breve{a},s^{\prime})+\gamma V_{t+1}^{\pi}(s^{\prime}))],
        \label{eq:mdp_reuse} \\
        Q_{t}^{\pi}(\bar{b},\breve{a})&=
        \mathbb{E}_{\bar{b}^{\prime -},o^{\prime},\bar{b}^{\prime}{\mid} q}
        [\rho^{p}_{q}(\bar{b}^{\prime -})
        (r(\bar{b},\breve{a},\bar{b}^{\prime})+\gamma V_{t+1}^{\pi}(\bar{b}^{\prime}))],
        \label{eq:pomdp_reuse}
    \end{talign}
\end{linenomath*}
where $\rho^{p_T}_{q}(s^\prime) \bydef p_T(s^{\prime}{\mid} s, \breve{a}) \slash q(s^{\prime} {\mid} s, \breve{a})$, and $\rho^{p}_{q}(\bar{b}^{\prime -}) \bydef p(\bar{b}^{\prime -} {\mid} \bar{b},\breve{a}) \slash q(\bar{b}^{\prime -} {\mid} \bar{b}, \breve{a})$ are the importance ratios.
The ratios require proposal support to cover target support.
We use $q(s^{\prime}{\mid} s,\breve{a})\coloneqq p_T(s^{\prime}{\mid}s,a)$, reusing samples from the previous action $a$, but generally other choices are possible too.
In Section~\ref{sec:mis_tree}, we extend Equations \eqref{eq:mdp_reuse}-\eqref{eq:pomdp_reuse} to an MIS setting with multiple proposals, where assumptions (\hyperlink{assum:mis-i}{MIS-I}) and (\hyperlink{assum:mis-ii}{MIS-II}) suffice for unbiasedness (Section~\ref{sec:background}).

We introduce our main results of action score gradients.
\begin{thm}
    \label{thm:mdp_action_grad}
    Assume: (i) $\rho^{p_T}_{q}(s^\prime)$, $\rho^{p}_{q}(\bar{b}^{\prime -})$ are well-defined; (ii) $r$, $V$ are bounded; (iii) there exist integrable $g_r,g_T$ w.r.t. the MDP/POMDP measures such that $\normflat{\nabla_a r}\leq g_r$ and $\normflat{\nabla_a \log p_T}\leq g_T$. For POMDPs, assume the assumptions of Lemma~\ref{lem:propagated_pb_likelihood} hold.
    Then, the action score gradient satisfies:
    \begin{linenomath*}
        \begin{talign}
            &\nabla_{\breve{a}}Q_{t}^{\pi}(s,\breve{a})
            =\mathbb{E}_{s^{\prime}{\mid} q}[\rho^{p_T}_{q}(s^\prime)
            [\nabla_{\breve{a}}\log p_{T}(s^\prime{\mid} s,\breve{a})
            \nonumber \\
            &\quad(r(s,\breve{a},s^{\prime})+\gamma V_{t+1}^{\pi}(s^{\prime}) - B(s))+\nabla_{\breve{a}} r(s,\breve{a},s^{\prime})]],
            \label{eq:mdp_action_grad}
            \\
            &\nabla_{\breve{a}}Q_{t}^{\pi}(\bar{b},\breve{a})
            = \mathbb{E}_{\bar{b}^{\prime -},o^{\prime},\bar{b}^{\prime}{\mid} q}
            [\rho^{p}_{q}(\bar{b}^{\prime -})
            [\nabla_{\breve{a}}\log p(\bar{b}^{\prime -}{\mid} \bar{b},\breve{a})
            \nonumber \\
            &\quad(r(\bar{b},\breve{a},\bar{b}^{\prime})+\gamma V_{t+1}^{\pi}(\bar{b}^{\prime}) - B(\bar{b}))+\nabla_{\breve{a}} r(\bar{b}, \breve{a}, \bar{b}^{\prime})]],
            \label{eq:pomdp_action_grad}
        \end{talign}
        where $B(s)$ (resp. $B(\bar{b})$) is any baseline independent of $\breve{a}$ for estimator variance reduction~\cite{Schulman16iclr}, e.g. $B(s)=V_{t}^{\pi}(s)$.
\end{linenomath*}
\end{thm}
The proof follows policy-gradient arguments~\cite{Sutton99nips}. 
\ProofAppendixPointer
The result allows computing action gradients with samples taken from the proposal $q$.
The gradient favors immediate reward and likely high-value successors; since $s^\prime$ is integrated out, $V_{t+1}^{\pi}(s^\prime)$ has no direct $\breve{a}$-gradient.
Compared with recursive estimates like VG-UCT~\cite{Lee20aaai}, these score gradients are branch-local in the MIS Tree, and apply to belief-dependent policies via Equation~\eqref{eq:propagated_belief_ratio}.

\subsection{MIS Trees for Action-Adaptive MCTS}
\label{sec:mis_tree}

\begin{figure}[t]
    \centering
    \makebox[0.99\columnwidth][c]{
    \begin{subfigure}[b]{0.3\columnwidth}
    \centering
    \includegraphics[width=\columnwidth]{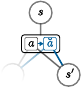}
    \end{subfigure}
    \hfill
    \begin{subfigure}[b]{0.3\columnwidth}
    \centering
    \includegraphics[width=\columnwidth]{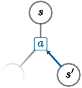}
    \end{subfigure}
    \hfill
    \begin{subfigure}[b]{0.3\columnwidth}
    \centering
    \includegraphics[width=\columnwidth]{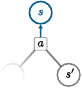}
    \end{subfigure}
    }
    \caption
    {
    Update operations in the MIS Tree, left to right, highlighted in cyan: (1) \textit{Action update}, a novel operation, computes a consistent action-value estimate for a new action $\breve{a}$ reusing previous successor states; (2) \textit{Action backpropagation}; (3) \textit{State backpropagation}.
    }
    \label{fig:mis_operations}
\end{figure}

\begin{algorithm*}[t]
    \caption{AGMCTS. Highlighted lines mark AGMCTS-specific MIS and action-optimization steps.}
    \label{alg:AGMCTS}
    \vspace*{-1.1\baselineskip}
    \begin{multicols}{2}
    \textbf{procedure}\;\textsc{Simulate}$(s,d)$
    \begin{algorithmic}[1]
    \IF {$d=0$}
    \STATE $v\leftarrow$ \textsc{Rollout}$(s)$
    \COMMENT{\textit{Rollout until terminal state}}
    \STATE \begingroup \AGHighlightColor \textsc{UpdateTerminal}($s,v$)
    \COMMENT {\textit{Eq. \eqref{eq:posterior_terminal_update}}} \endgroup
    \STATE \textbf{return} $v$
    \ENDIF
    \STATE $a\leftarrow$ \textsc{ActionProgWiden}$(s)$ 
    \COMMENT{\textit{Vanilla or VPW}}
    \STATE \begingroup \AGHighlightColor $addBranch \leftarrow $ \textsc{ActionOpt} $(s,a,d)$ \endgroup
    \IF {\begingroup \AGHighlightColor $\absvalflat{\mathcal{S}_{sa}}\leq k_{o}\cdot n(s,a)^{\alpha_{o}}$ OR $addBranch$ \endgroup}
    \STATE $s^{\prime},r \sim G(s,a)$
    \begingroup \AGHighlightColor
    \COMMENT {$\bar{b}^{\prime -},\bar{b}^{\prime},r \sim G(\bar{b},a)$ \textit{in POMDPs}}
    \endgroup
    \STATE $\mathcal{S}_{sa}\leftarrow \mathcal{S}_{sa}\cup\left\{ s^{\prime}\right\}$ 
    \STATE $v\leftarrow\text{\textsc{Rollout}}(s^{\prime},d-1)$
    \ELSE
    \STATE $s^{\prime} \sim \textnormal{Unif}(\mathcal{S}_{sa})$, $r\leftarrow$ \textsc{Reward}$(s,a,s^{\prime})$
    \STATE $v\leftarrow\text{\textsc{Simulate}}(s^{\prime},d-1)$
    \ENDIF
    \STATE \begingroup \AGHighlightColor \textsc{UpdateMIS}($s,a,s^\prime,r$)
    \COMMENT {\textit{Eqs. \eqref{eq:action_backprop_update_1}, \eqref{eq:action_backprop_update_2}, \eqref{eq:posterior_nonterminal_update}}} \endgroup
    \end{algorithmic}
    \AGHighlightColor
    \textbf{procedure}\;$\text{\textsc{ActionOpt}}(s,a,d)$
    \begin{algorithmic}[1]
    \STATE $addBranch\leftarrow$ FALSE
    \FORALL {$k=1,\dots,K_{\text{opt}}$} 
    \STATE $g_a^q \leftarrow \hat{\nabla}_{a}\tilde{Q}(s,a)$
    \COMMENT {\textit{Eq. \eqref{eq:mdp_grad_est_linearized}}}
    \STATE $\breve{a} \leftarrow \text{\textsc{Opt}}(s,a,g_a^q)$
    \COMMENT {\textit{Adam/other}}
    \STATE $\breve{a} \leftarrow $\textsc{ClipNorm}$(\breve{a},T_{d_a}^{\max})$
    \STATE \textsc{ActionUpdate}$(s, a, \breve{a})$
    \COMMENT {\textit{Eq. \eqref{eq:branch_imp_ratio}-\eqref{eq:imp_ratio_linear_update}, \eqref{eq:posterior_nonterminal_update}}}
    \STATE $addBranch \leftarrow addBranch \lor (\forall s^{\prime,i}\in \mathcal{S}_{s\breve{a}}: \rho_{\breve{a}}^{i}(s^{\prime,i})\leq T_{\rho}^{add})$
    \ENDFOR
    \STATE \textbf{return} $addBranch$
    \end{algorithmic}
    \end{multicols}
    \vspace*{-0.7\baselineskip}
\end{algorithm*}

We introduce the MIS Tree, a novel search-tree whose action-value estimates are MIS estimators.
It supports standard backpropagation plus \textit{action update}: replacing an action label while reusing previous successor states for a consistent updated action-value estimate.
Figure~\ref{fig:mis_operations} illustrates these operations.
We formulate the MIS Tree recursively over state nodes $s$ and action nodes $(s,a)$.
For POMDPs, we use a particle-filter tree (PFT)~\cite{Sunberg18icaps}, replacing $s$ by ordered particle belief states $\bar{b}$ throughout.

Each state node $s$ has a set of child actions $\mathcal{A}_s\bydef\{a^i\}_{i}$.
Each action node $(s,a)$ has a set of child successor states
$\mathcal{S}_{sa}\bydef\{s^{\prime,i}\}_{i}$.
The visitation counts of state and action nodes are $n(s)$ and $n(s,a)$.
We store for each successor $s^{\prime,i}$ the proposal action $a_{\textnormal{prop}}^i$, namely the action under which $s^{\prime,i}$ was originally sampled.
This stored proposal action may differ from the current action label $a$ after action updates.
We also store the value estimate $\hat{V}(s^{\prime,i})$, defined by:
\begin{linenomath*}
    \begin{talign}
        \hat{V}(s) \bydef n(s)^{-1}\sum_{i=1}^{\absvalflat{\mathcal{A}_{s}}}n(s,a^i)\hat{Q}(s,a^i).
        \label{eq:state_value_estimator}
    \end{talign}
\end{linenomath*}
The visitation counts satisfy:
\begin{linenomath*}
    \begin{talign}
        n(s) = \sum_{i=1}^{\absvalflat{\mathcal{A}_{s}}}n(s,a^i), \quad
        n(s,a)=\sum_{i=1}^{\absvalflat{\mathcal{S}_{sa}}}n(s^{\prime,i})_{+1}. \label{eq:visits_nodes_mcts}
    \end{talign}
\end{linenomath*}
Here, $n(s^{\prime})_{+1}\bydef n(s^{\prime})+1$ is the post-expansion count contribution of successor $s^\prime$.
The $+1$ accounts for the initial rollout value stored when the successor node is created, while $n(s^\prime)$ counts later visits.
For action nodes, $\hat{Q}(s,a) \bydef \hat{r}(s,a)+\gamma\hat{V}_{f}(s,a)$, where $\hat r$ estimates immediate reward and $\hat V_f$ estimates next-step value.
Our key observation is that this decomposition lets \textit{action update} recompute $\hat V_f$ from older successor samples and their stored value estimates.

Generally in MIS, each proposal distribution $q_i$ has $n_i$ samples.
In our situation, we assume that each $s^{\prime,i}\in \mathcal{S}_{sa}$ was sampled from $p_T(\cdot{\mid} s,a_{\textnormal{prop}}^i)$, with its associated single value estimate $V(s^{\prime,i})$.
Therefore, we adapt the MIS definition to combine successor-specific estimators, based on the associated proposal action, branch importance ratio $\rho_{a}^{i}(s^{\prime})$ and MIS weights $w_i(s^{\prime})$:
\begin{linenomath*}
    \begin{talign}
        \hat{V}_{f}(s,a) &\bydef \sum_{i=1}^{\absvalflat{\mathcal{S}_{sa}}} w_i(s^{\prime,i}) \rho_{a}^{i}(s^{\prime,i}) \hat{V}(s^{\prime,i}), \\
        \rho_{a}^{i}(s^{\prime}) &\bydef p_T(s^{\prime}{\mid} s,a)\slash p_T(s^{\prime} {\mid} s,a_{\textnormal{prop}}^i), \label{eq:branch_imp_ratio}
    \end{talign}
\end{linenomath*}
and for POMDPs $\rho_{a}^{i}(\bar{b}^{\prime}) \bydef p(\bar{b}^{\prime -}{\mid} \bar{b},a)\slash p(\bar{b}^{\prime -} {\mid} \bar{b},a_{\textnormal{prop}}^i)$.
Analogously, replace $\hat V(s^{\prime,i})$ with $r(s,a,s^{\prime,i})$ to obtain $\hat r(s,a)$.

The following theorem shows that the MIS conditions suffice for unbiasedness of the MIS Tree:
\begin{thm}
    \label{thm:mis_tree}
    If $w_i(s^{\prime})$ satisfy requirements (\hyperlink{assum:mis-i}{MIS-I}) and (\hyperlink{assum:mis-ii}{MIS-II}) (see Section~\ref{sec:background}), then the MIS Tree yields unbiased value and action-value estimates for a given tree structure.
\end{thm}

The MIS balance heuristic requires $O(\absvalflat{\mathcal{S}_{sa}})$ density evaluations per recursive sample update~\cite{Novitsky25ral}.
For efficiency, we use self-normalized MIS (SN-MIS)~\cite{Metelli20jmlr} with naive weights $w_i(s^{\prime})\propto n(s^{\prime, i})_{+1}$, which we found to perform well in practice:
\begin{linenomath*}
    \begin{talign}
        \tilde{V}_{f}(s,a) &\bydef \sum_{i=1}^{\absvalflat{\mathcal{S}_{sa}}} n(s^{\prime,i})_{+1} \rho_{a}^{i}(s^{\prime,i}) \hat{V}(s^{\prime,i}) \slash \eta(s,a),
        \label{eq:snmis_vf}
        \\
        \eta(s,a) &\bydef \sum_{i=1}^{\absvalflat{\mathcal{S}_{sa}}} n(s^{\prime,i})_{+1} \rho_{a}^{i}(s^{\prime,i}). \label{eq:snmis_eta}
    \end{talign}
\end{linenomath*}
Using SN-MIS, all MCTS operations are $O(1)$ except $O(\absvalflat{\mathcal{S}_{sa}})$ \textit{action update}.
\MisTreeAppendixDetails

\paragraph{Action Update.} Action update replaces $(s,a)$ by $(s,\breve{a})$, recomputes all ratios, \eqref{eq:snmis_vf}, \eqref{eq:snmis_eta}, and immediate rewards in $O(\absvalflat{\mathcal{S}_{sa}})$ time.

In POMDPs, the explicit computation of the new transition likelihood $p(\bar{b}^{\prime -}{\mid} \bar{b},\breve{a})$ via \eqref{eq:propagated_belief_likelihood} requires $J$ evaluations of the transition density, for $J$ particles in $\bar{b}$ and $\bar{b}^{\prime -}$, which may be computationally expensive.
As $p(\bar{b}^{\prime -}{\mid} \bar{b},\breve{a})$ is a product of probabilities, a direct MC estimate based on a subset of particles results in a large bias.
We resort to estimating changes in $\log p(\bar{b}^{\prime -}{\mid} \bar{b},\breve{a})$. 
For small $\normflat{\breve{a} - a}$ we approximate\footnote{This is obtained by the first-order approximation $f(x+\delta x) \approx f(x) + \nabla f(x)^T \delta x$, applied to $\log \rho_{a}^{i}(s^{\prime})$ where $\delta a = \breve{a} - a$.}
\begin{linenomath*}
    \begin{talign}
        \log \rho_{\breve{a}}^{i}(\bar{b}^{\prime -}) - \log \rho_{a}^{i}(\bar{b}^{\prime -}) &\approx (\nabla_a \log \rho_{a}^{i}(\bar{b}^{\prime -}))^T \delta a
        \nonumber \\
        = (\nabla_a \log p(\bar{b}^{\prime -}{\mid} \bar{b},a))^T \delta a &= (\textnormal{Eq. } \eqref{eq:propagated_belief_grad})^T \delta a, \label{eq:imp_ratio_linear_update}
    \end{talign}
\end{linenomath*}
and due to Equation \eqref{eq:propagated_belief_grad} being a sum, it admits an unbiased MC estimate by subsampling particles.

\paragraph{Action Backpropagation.} Let $s^{\prime,i}$ be an updated node. Hence, we updated $n(s^{\prime,i})$ to $n^{\prime}(s^{\prime,i})$ and $\hat{V}(s^{\prime,i})$ to $\hat{V}^{\prime}(s^{\prime,i})$. We update $n(s,a)$ by \eqref{eq:visits_nodes_mcts} and perform
\begin{linenomath*}
    \begin{talign}
        \eta^{\prime}(s,a)&=\eta(s,a)+\rho_{a}^{i}(s^{\prime,i}) (n{^\prime}(s^{\prime,i}) - n(s^{\prime,i})),
        \label{eq:action_backprop_update_1} \\
        \tilde{V}_{f}^{\prime}(s,a)&=(\eta^{\prime}(s,a))^{-1}\big(\eta(s,a)\tilde{V}_{f}(s,a)
        \nonumber \\
        &+ \rho_{a}^{i}(s^{\prime,i})(n{^\prime}(s^{\prime,i})\hat{V}^{\prime}(s^{\prime,i}) - n(s^{\prime,i})\hat{V}(s^{\prime,i}))\big).
        \label{eq:action_backprop_update_2}
    \end{talign}
\end{linenomath*}
Assuming a general $n^{\prime}(s^{\prime})$ also supports cases where we delete branches in the tree, i.e. $n^{\prime}(s^{\prime}) < 1$. For \textit{state expansion}, the same equations apply by initializing $n(s^{\prime})=0$.

\paragraph{State Backpropagation.} Let $s$ be a state node.
If $s$ is at the maximum tree depth, $\hat{V}(s)$ is based only on rollouts. We update the running average with the new rollout value $v^{\prime}$,
\begin{linenomath*}
    \begin{talign}
        \hspace{-0.01cm}\hat{V}^{\prime}(s)=\hat{V}(s)+(n^{\prime}(s)-n(s))(v^{\prime}-\hat{V}(s)) \slash n^{\prime}(s)_{+1},
        \label{eq:posterior_terminal_update}
    \end{talign}
\end{linenomath*}
If $s$ is not a terminal state, let its recently updated action-value child be $(s,a)$, with updated $n^{\prime}(s,a)$ and $\tilde{Q}^{\prime}(s,a)$. We update $n^{\prime}(s)$ by \eqref{eq:visits_nodes_mcts} and perform
\begin{linenomath*}
    \begin{talign}
        \hat{V}^{\prime}(s)&=(n^{\prime}(s))^{-1}(n(s)\hat{V}(s) 
        \nonumber\\
        &\quad + n^{\prime}(s,a)\tilde{Q}^{\prime}(s,a) - n(s,a)\tilde{Q}(s,a)).
        \label{eq:posterior_nonterminal_update}
    \end{talign}
\end{linenomath*}

\subsection{Densities of Smooth Generative Models via the Area Formula}
\label{sec:area_formula}

Thus far we have assumed computable $p_T(s^{\prime}{\mid} s,a)$ and $\nabla_a\log p_T(s^{\prime}{\mid} s,a)$.
In practice, many simulators provide instead $s^{\prime}=f(s,a,\xi)$ with noise $\xi\sim p_{\xi}(\cdot {\mid} s,a)\in\mathbb{R}^{n_\xi}$.
If $n_\xi=n_s$ and $\xi\mapsto f(s,a,\xi)$ is bijective and differentiable, the change-of-variables formula states that for $s^\prime=f(s,a,\xi^{*})$:
\begin{linenomath*}
    \begin{align}
        \textstyle p_T(s^\prime {\mid} s,a)=p_{\xi}(\xi^{*}{\mid} s,a) \left|{D_\xi f(s,a,\xi^{*})}\right|^{-1},
    \end{align}
\end{linenomath*}
where $D_\xi f$ is the Jacobian matrix of $f$ w.r.t. $\xi$.
However, it is often the case that $n_{\xi}<n_s$ and $\xi\mapsto f(s,a,\xi)$ is an embedding into an $n_{\xi}$-dimensional manifold rather than a bijection.
Then $p_T(s^\prime {\mid} s,a)$ is a density with respect to the induced manifold measure, not Lebesgue measure, and the Area Formula~\cite[Theorem 3.2.3]{Federer69book} applies.
\citeauthor{Negro22}~[\citeyear{Negro22}] gives the following locally Lipschitz form for $f$ and the induced density $p_T$:
\begin{manualtheorem}{4.1}~\cite{Negro22}
If $f$ is locally Lipschitz, and it holds that $\operatorname{rank}(D_\xi f)=n_\xi$ a.e., then the induced probability measure over $s^\prime$ is absolutely continuous w.r.t. the Hausdorff measure $\mathcal{H}^{n_\xi}$ on $\mathbb{R}^{n_s}$, and its Radon-Nikodym derivative is
\begin{linenomath*}
    \begin{align}
        \textstyle p_T(s^{\prime}{\mid} s,a)=\sum_{\Xi^{*}}p_{\xi}(\xi^{*}{\mid} s,a)(J_{n_{\xi}}f(s,a,\xi^{*}))^{-1},
        \label{eq:area_formula}
    \end{align}
\end{linenomath*}
for $\Xi^{*}=\{\xi^{*}:s^{\prime}=f(s,a,\xi^{*})\}$, and $0$ when $\Xi^{*}=\emptyset$.
The $n_{\xi}$-dimensional Jacobian is given by the Cauchy-Binet formula: $J_{n_{\xi}}f(s,a,\xi)=\sqrt{\det((D_\xi f)^{\intercal} \cdot (D_\xi f))}$.
\end{manualtheorem}
\AreaFormulaAppendixDetails

Robotics simulators often integrate continuous-time dynamics with stochasticity injected at the input or output.
In both cases below, once $p_T$ is computable, $\nabla_a\log p_T$ follows by chain rule and automatic differentiation.

\paragraph{Input Noise to Simulator.} Let the input noise be described by the function $h(s,a,\xi)=(\tilde{s},\tilde{a})$.
The simulator then produces a successor state via $s^{\prime}=f(\tilde{s},\tilde{a})$.
If for a new action $\breve{a}$ there exists a unique $\xi^{*}$ satisfying $h(s,\breve{a},\xi^{*})=(\tilde{s}, \tilde{a})$, i.e. such that the input to the simulator remains unchanged, then we can evaluate the sensitivity of the output state with respect to the noise: 
$\textstyle D_\xi f(s,\breve{a},\xi^{*})
= D_{(\tilde{s}, \tilde{a})} f \cdot \frac{\partial h}{\partial \xi} |_{s, \breve{a}, \xi^{*}}$.
Moreover, the Jacobian matrix $D_{(\tilde{s}, \tilde{a})} f$ can be cached from the original sample $(s,a,\xi,s^\prime)$, allowing to compute the density without requiring additional simulator queries.

\paragraph{Noise in Simulator Output.} Let the simulator output be $h(s,a)$, and the next state be given by $s^{\prime}=g(h(s,a),\xi)$.
If we can find all $\xi^{*}$ such that $s^{\prime}=g(h(s,\breve{a}),\xi^{*})$, then we compute the Jacobian via $D_\xi f(s,\breve{a},\xi^{*})=\frac{\partial g}{\partial \xi}|_{h(s,\breve{a}),\xi^{*}}$, requiring a single forward simulation $h(s,\breve{a})$ for all $\xi^{*}$.

\section{AGMCTS Algorithm}
\label{sec:agmcts}

AGMCTS combines MCTS with frequent action-gradient optimization, and correcting online value estimates via MIS.
It uses a forward simulator $G(s,a)$ with assumed access to $p_T$ and $p_O$, controls branching via DPW, and extends to POMDPs via ordered particle-belief states like PFT-DPW~\cite{Sunberg18icaps}.
Algorithm~\ref{alg:AGMCTS} keeps the standard simulation loop; highlighted components add \textsc{ActionOpt} and MIS \textit{action update} before expansion so gradients influence expansion.
\AppendixOrPublic{We detail the implementation in Appendix~\ref{sec:appendix_agmcts}.}{We detail the implementation in the Appendix.}

\paragraph{Monte Carlo Gradient Estimation.} In POMDPs, we estimate \eqref{eq:propagated_belief_grad} with $K_b^{\nabla}$ state particles via 
$ \nabla_a \log p(\bar{b}^{\prime -}{\mid} \bar{b},a)\approx (J \slash K_{b}^{\nabla})\sum_{l=1}^{K_{b}^{\nabla}} \nabla_a \log p_T(s^{-,j_l}{\mid} s^{j_l}, a), $
where indices $j_l$ are sampled uniformly from $[1,\dots,J]$, and $J$ is the particle count of $\bar{b}$ and $\bar{b}^{\prime -}$.
We choose a baseline function of the current value estimate $B(s)=\hat{V}(s)$ ($B(\bar{b})=\hat{V}(\bar{b})$ in POMDPs), effectively estimating advantage gradients. 
As we linearize the weight updates as in Equation \eqref{eq:imp_ratio_linear_update}, $\nabla_a \rho_{a}^{i}(s^{\prime,i})$ has to be computed for all $s^{\prime,i}\in \mathcal{S}_{sa}$.
To save computations, we sum over all successor states $s^{\prime}\in \mathcal{S}_{sa}$ when computing $\hat{\nabla}\tilde{Q}(s,a)$:
\begin{linenomath*}
    \begin{talign}
        \hat{\nabla}_{a}\tilde{Q}(s,a)
        =\sum_{i=1}^{\absval{\mathcal{S}_{sa}}} n(s^{\prime,i})_{+1} \rho_{a}^{i}(s^{\prime,i}) \cdot 
        ( \nabla_{a}\log p_{T}(s^{\prime,i} {\mid} s,a)
        \nonumber \\
        \cdot(r(s,a,s^{\prime,i})+\gamma\hat{V}(s^{\prime,i}) - B(s))+\nabla_{a}r(s,a,s^{\prime,i}) ) \slash \eta(s,a),
        \label{eq:mdp_grad_est_linearized}
    \end{talign}
\end{linenomath*}
and we cache $\nabla_{a}\log p_{T}(s^{\prime,i} {\mid} s,a)$ for later \textit{action updates}.

\paragraph{Gradient Optimization.} We used Adam~\cite{Kingma15iclr} for its step-size normalization, which is crucial for high-variance gradients.
We perform $K_{\text{opt}}$ consecutive gradient updates to control the accuracy-compute trade-off, and additional clipping of $\normflat{\breve{a}-a}$ to $T_{d_a}^{\max}$ to stabilize updates.

\paragraph{Threshold for Adding Successor Nodes.} If $\rho_{\breve{a}}^{i}(s^{\prime,i})<T_{\rho}^{add}$ for all $s^{\prime,i}\in \mathcal{S}_{s\breve{a}}$, we force a new successor sample, with optional low-relevance deletion.

\paragraph{Action Sampling Heuristics.} We use uniform action sampling and Voronoi progressive widening (VPW)~\cite{Lim21cdc}; VPW biases samples through Voronoi partitions, demonstrating that AGMCTS can benefit from different action proposal mechanisms.

\begin{figure*}[h]
    \includegraphics[width=0.95\textwidth]{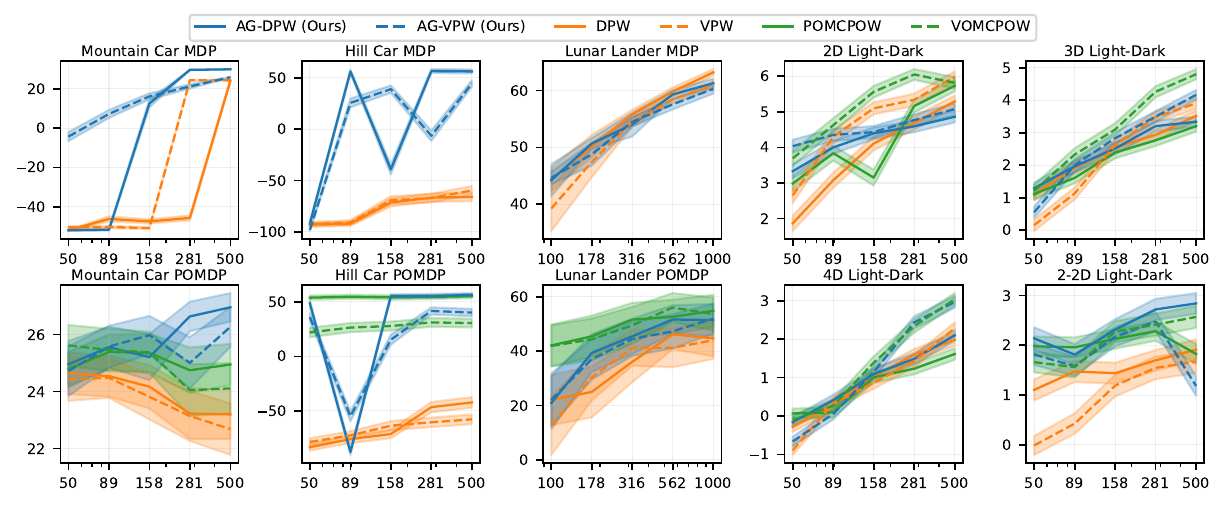}
    \caption{ 
        Average returns over 1000 seeds, shaded areas show $\pm 2$ standard error. The x-axis shows the simulations budget per planning session. Ours: AGMCTS algorithms Action-Gradient DPW (AG-DPW) for MDPs, and its particle-filter tree (PFT) adaptation for POMDPs.
        Orange: DPW for MDPs and PFT-DPW for POMDPs. Green: POMCPOW (POMDP domains only).
        We test both vanilla action sampling (solid lines) and VPW (dashed lines), resulting in 4 MDP and 6 POMDP algorithms.
    }
    \label{fig:mdp_pomdp_grid_results}
\end{figure*}

\section{Experiments}
\label{sec:experiments}

We evaluate AGMCTS on continuous MDP/POMDP benchmarks. MDP baselines are DPW/VPW; POMDP baselines are PFT-DPW/PFT-VPW and POMCPOW/VOMCPOW from POMDPs.jl~\cite{egorov2017pomdps}.
AGMCTS also uses POMDPs.jl.
VPW and result analysis follow the codebase of~\cite{Lim21cdc}; \AppendixOrPublic{we give additional hyperparameters, result tables, and implementation details in Appendix~\ref{sec:appendix_experiments}.}{we give additional hyperparameters, result tables, and implementation details in the Appendix.}

\subsection{Benchmark Domains}

\paragraph{\boldmath{$d$}D-Continuous Light-Dark POMDP.} This domain generalizes the Light-Dark POMDP~\cite{Platt10rss} to arbitrary dimension $d$ and combines a narrow goal-centered reward peak, a surrounding moat penalty, a non-Gaussian initial belief, and observation noise that changes sharply across the state space.
Here $\mathcal{S}=\mathcal{O}=\mathbb{R}^d$, $b_0=S^{d-1}_{0.5}$, goal $s_g=(\boldsymbol{0}_{d-1},2.5)$, beacon $s_b=(2.5,\boldsymbol{0}_{d-1})$, $\gamma=0.99$, terminal radius $T_g=0.2$, and horizon $L=6$.
Actions lie in $B_{1.5}(\boldsymbol{0}_d)$, transitions are $s'\sim\mathcal{N}(s+a,\sigma_T^2\mathbf{I})$ with $\sigma_T=0.025$, and observations are relative beacon measurements $o\sim\mathcal{N}(s-s_b,\sigma_O(\normflat{s-s_b})^2\mathbf{I})$ with $\sigma_O(x)=\min\{15,0.01(x+x^8)\}$.
The reward depends only on the successor state and combines three terms: a sharp positive bump at the goal, a surrounding moat penalty that discourages random wandering near the target, and a small quadratic distance penalty. With $\Delta_g=\normflat{s'-s_g}$, the constants are $R_{\textnormal{goal}}=10$, $R_{\textnormal{moat}}=2$, and $R_{\textnormal{dist}}=0.02$, giving $R_{\textnormal{goal}}e^{-\frac{1}{2}(\Delta_g/(0.5T_g))^2}-R_{\textnormal{moat}}e^{-\frac{1}{2}((\Delta_g-5T_g)/T_g)^2}-R_{\textnormal{dist}}\Delta_g^2$.
Rollouts move toward $s_g$ but add $\mathcal{N}(\boldsymbol{0},0.1^2\mathbf{I})$ action noise, providing a noisy goal-directed default policy. We benchmark $d=2,3,4$.

\paragraph{Two-Agent $d$D-Continuous Light-Dark POMDP.} This variant uses $\mathcal{S}=\mathcal{O}=\mathbb{R}^{2d}$, and $\mathcal{A}=B_{1.5}(\boldsymbol{0}_d)\times B_{1.5}(\boldsymbol{0}_d)$.
Each agent has the same transition and individual reward structure, with the positive goal term obtained only when both agents are within $T_g$.
The observation model couples the agents: agent 1 observes its relative beacon position, while agent 2 observes its relative position to agent 1 with the same distance-dependent covariance.
Thus agent 2 receives position information through agent 1 rather than directly through the fixed beacon, while the reward grants success jointly.
We benchmark $d=2$.

\paragraph{Mountain Car and Hill Car MDP/POMDP.} These domains are based on Mountain Car~\cite{Singh96ml} and Car on the Hill~\cite{Ernst05jmlr}; $\mathcal{S}=\mathbb{R}^2$ is position and velocity, and $\mathcal{A}=[-a_{\max},a_{\max}]$ accelerates/brakes the car while avoiding the left boundary.
In POMDPs, velocity is unobserved and $z\sim\mathcal{N}(x,0.03^2)$.
These domains have momentum dynamics and threshold termination: the car may move away from the goal to gain speed, while overshooting the safe speed range causes termination.
The reward is $100$ at $x\geq x_{\max}$, $-100$ at $x<x_{\min}$ or $\absval{v}\geq v_{\max}$, and $-0.1$ otherwise; $\gamma=0.99$.
Rollouts use the standard momentum heuristic, pushing with $a_{\max}$ when $v>0$ and $-a_{\max}$ otherwise, which provides a useful but myopic default.
Our modified Mountain Car instance uses $x_{\min}=-1.5$, $x_{\max}=0.5$, $v_{\max}=0.05$, $a_{\max}=1.0$, horizon $L=200$, $v'=v+0.001\tilde{a}-0.0025\cos(3x)$, and $x'=x+v'$.
Hill Car uses horizon $L=30$ and integrates nonlinear continuous-time dynamics for $\Delta t=0.1$, with the same noisy-clamped action interface used for AGMCTS density evaluation.

The two car domains differ in where complexity appears.
Mountain Car has a long horizon under a tight velocity bound, so momentum-building policies can terminate by exceeding the speed limit.
Hill Car has a shorter horizon, but each transition is obtained by integrating nonlinear continuous-time dynamics.
In both domains, additive action-input noise $\xi\sim\mathcal{N}(0,0.1^2)$ is clamped as $\tilde{a}=\textnormal{clip}(a+\xi,-a_{\max},a_{\max})$ before simulation, yielding transition distributions with continuous interior components and discrete clipping masses.

\paragraph{Lunar Lander MDP/POMDP.} We use Lunar Lander as in~\cite{Mern21aaai,Lim21cdc}, with state $(x,y,\theta,\dot{x},\dot{y},\omega)$ and action $(F_x,T,\delta)\in[0,15]\times[-5,5]\times[-1,1]$.
Actions couple translation, orientation, and angular rate through the thrust direction and torque, and rewards combine threshold penalties with a landing reward.
Here $T$ is main thrust, $F_x$ corrective thrust, and $\delta$ the offset producing torque $\tau=-\delta F_x$.
With $m=1$, $I=10$, and $\Delta t=0.4$, the deterministic model computes $f_x=\cos\theta F_x-\sin\theta T$, $f_y=\cos\theta T+\sin\theta F_x$, $\dot{x}'=\dot{x}+f_x\Delta t$, $\dot{y}'=\dot{y}+(f_y-9)\Delta t$, $\omega'=\omega+\tau\Delta t/I$, $x'=x+\dot{x}\Delta t$, $y'=y+\dot{y}\Delta t$, and $\theta'=\theta+\omega\Delta t$.
Gaussian output noise with scales $(0.1,0.1,0.01)$ is added to $\dot{x}'$, $\dot{y}'$, and $\omega'$, so transition densities are computed by solving for the output-noise variables under the Area Formula.
In the POMDP variant, $o=(\tilde{\omega},\tilde{\dot{x}},\tilde{h})$ with noise scales $(1.0,0.01,0.1)$ and $\tilde{h}$ centered at $y/\cos\theta$.
Rewards are $-1000$ if $x\geq15$ or $\theta\geq0.5$, $100-x-\dot{y}^2$ if $y\leq1$, and $-1$ otherwise.
We use $\gamma=0.99$ and horizon $L=35$.
The state distribution is a Gaussian with mean $(0,50,0,0,-10,0)$, and the rollout is a proportional stabilizer $(-0.1\dot{x},-0.1\dot{y},0)$ under noisy partial observations.

\subsection{Performance Evaluation}
\label{sec:experiments_eval}

\begin{table}[t]
    \centering
    \small
    \setlength{\tabcolsep}{2.5pt}
    \begin{tabular}{@{}llrrrr@{}}
        \toprule
        Domain & Setting & DPW & VPW & AG-DPW & AG-VPW \\
        \midrule
        2D Light-Dark & POMDP & .027 & .027 & .316 & .191 \\
        3D Light-Dark & POMDP & .068 & .073 & .315 & .154 \\
        4D Light-Dark & POMDP & .151 & .155 & .302 & .318 \\
        2-2D Light-Dark & POMDP & .160 & .173 & .447 & .464 \\
        \midrule
        \multirow{2}{*}{Mountain Car} & MDP & .016 & .016 & .148 & .152 \\
        & POMDP & .042 & .044 & .928 & .904 \\
        \multirow{2}{*}{Hill Car} & MDP & .026 & .029 & .744 & 1.028 \\
        & POMDP & .360 & .347 & 1.636 & 1.109 \\
        \multirow{2}{*}{Lunar Lander} & MDP & .008 & .009 & .203 & .216 \\
        & POMDP & .103 & .102 & .504 & .508 \\
        \bottomrule
    \end{tabular}

    \vspace{0.4em}

    \begin{tabular}{@{}lrr@{}}
        \toprule
        POMDP domain & POMCPOW & VOMCPOW \\
        \midrule
        2D Light-Dark & .033 & .033 \\
        3D Light-Dark & .060 & .063 \\
        4D Light-Dark & .141 & .170 \\
        2-2D Light-Dark & .141 & .169 \\
        Mountain Car & .116 & .134 \\
        Hill Car & .348 & .297 \\
        Lunar Lander & .136 & .144 \\
        \bottomrule
    \end{tabular}
    \caption{Mean planning time in seconds at the maximum simulation budget of each domain. In POMDP domains, POMCPOW/VOMCPOW use a larger simulation budget to match runtimes similar to PFT-DPW/PFT-VPW.}
    \label{tab:runtime_summary}
\end{table}

We tune hyperparameters with cross-entropy maximization at the maximum simulation budget.
This includes UCT/DPW parameters, and Adam step size for AG variants, while other AG settings are manual.
VPW Voronoi parameters use reasonable values from~\cite{Lim21cdc}.
The CE procedure runs 50 iterations; each iteration samples 150 parameter settings, evaluates each on 40 seeds, and refits a smoothed Gaussian proposal from the 30 elite settings.
All reported means use 1000 shared seeds.
Maximum planning budgets are $500$ simulations for Light-Dark and Car domains and $1000$ for Lunar Lander.
These simulation-budget comparisons intentionally hold the number of model queries fixed, giving a clearer test of whether action-gradient refinement improves the same underlying tree-search procedures rather than conflating this effect with implementation-dependent wall-clock costs.
For wall-clock matching, POMCPOW receives a larger state-trajectory budget matched to PFT-DPW runtime.

Transition-density computations match the domain structure: closed-form Gaussian densities in Light-Dark, clipped input-noise inversion in Car domains, and output-noise inversion in Lunar Lander.
AG variants use $K_{\textnormal{opt}}\in\{2,3\}$ Adam updates with linearized importance-weight updates~\eqref{eq:imp_ratio_linear_update}. 

Figure~\ref{fig:mdp_pomdp_grid_results} shows mean performance per planning budget. At maximum budget, an AG variant was preferable to its matched non-AG DPW/VPW-family baseline in 6/10 scenarios (all Hill/Mountain Car, 4D and 2x2D Light-Dark).
Against matched baselines (AG-DPW vs. DPW, AG-VPW vs. VPW, and the corresponding PFT variants), AG improved significantly in 10/20 cases, was insignificant in 7/20, and underperformed in 3/20.

By best algorithm per scenario, AGMCTS led in Mountain/Hill Car MDPs, was best or statistically tied in 5/7 POMDPs, and underperformed in 2/7. Overall, it was competitive with the wall-clock-matched POMCPOW/VOMCPOW baselines and often improved over PFT-DPW/PFT-VPW.
The largest gains appear in the Mountain/Hill Car domains, while Light-Dark and Lunar Lander show more mixed outcomes. 
The results suggest that AGMCTS is effective in domains
where small action changes can lead to large changes in states and rewards, as in the very long-horizon Mountain Car or very narrow goal in the Light-Dark POMDPs.
The performance difference in Lunar Lander is mostly insignificant, which may be due to the combination of large action dimensionality and sparse rewards, reducing gradient signal quality.

On the other hand, AG variants were $10\text{--}35\mathord{\times}$ slower than DPW/VPW in MDPs and $2\text{--}20\mathord{\times}$ slower than PFT-DPW/VPW in POMDPs, as shown in Table~\ref{tab:runtime_summary}.
AG variants have higher per-simulation cost because action updates revise successor weights and evaluate transition-density terms; these single-threaded timings therefore report implementation cost rather than a wall-clock optimum, since particle-belief updates in AGMCTS and PFT-DPW can be parallelized over particles.
Thus, the practical trade-off is whether improved action quality per model query offsets this density and gradient overhead within an application-relevant latency budget.

\section{Conclusions}
\label{sec:conclusions}

We introduced action-gradient continuous-POMDP tree search with MIS-consistent action-branch updates.
Using the Area Formula to obtain transition log-probabilities, AGMCTS improved several continuous MDP/POMDP benchmarks and was competitive overall, showing that online gradient refinement can strengthen MCTS.
Future work includes convergence analysis and faster, less tuning-sensitive general planning algorithms.

\paragraph{Practical Considerations.} AGMCTS requires transition-density access and tuning in addition to the per-simulation overhead.
These costs should be weighed against application latency and modeling constraints.

\ifthenelse{\boolean{anonymous}}%
{}
{ 
\section*{Acknowledgments}
This work was supported by the Israel Ministry of Innovation, Science and Technology.
}

\bibliographystyle{named}
\bibliography{../../../../../../references/refs,more_refs}

\appendix

\ifthenelse{\boolean{appendix}}
{
    \clearpage
    \input{appendix_content.tex}
}
{}

\end{document}

%% file: packages.tex
\usepackage{times}
\usepackage{soul}
\usepackage{url}
\usepackage[hidelinks]{hyperref}
\usepackage[utf8]{inputenc}
\usepackage[font=small]{caption}
\usepackage{graphicx}
\usepackage{amsmath}
\usepackage{amsthm}
\usepackage{booktabs}
\usepackage[switch]{lineno}

\usepackage{makecell}

\usepackage{algorithm}
\usepackage[noend]{algorithmic}

\usepackage{amssymb}
\usepackage{mathtools}
\usepackage[titletoc,title]{appendix}
\usepackage{bbm}
\usepackage{cancel}
\usepackage{enumitem}
\usepackage{multirow}
\usepackage{multicol}
\usepackage{placeins}
\usepackage{romannum}
\usepackage{subcaption}
\usepackage{tabularx}
\usepackage{wrapfig}
\usepackage{xspace}
\usepackage[dvipsnames, svgnames]{xcolor}
\usepackage{colortbl}

\newlength{\algoboxheight}
\usepackage{siunitx}

%% file: commands.tex
\theoremstyle{plain}
\newtheorem{thm}{\protect\theoremname}
\providecommand{\theoremname}{Theorem}

\theoremstyle{plain}

\providecommand{\claimname}{Claim}

\theoremstyle{plain}
\newtheorem{lem}{\protect\lemmaname}
\providecommand{\lemmaname}{Lemma}

\theoremstyle{plain}

\providecommand{\corollaryname}{Corollary}

\theoremstyle{definition}

\providecommand{\definitionname}{Definition}

\theoremstyle{remark}

\theoremstyle{plain}
\newtheorem{manualtheoreminner}{Theorem}[thm]
\newenvironment{manualtheorem}[1]{%
  \renewcommand\themanualtheoreminner{#1}%
  \manualtheoreminner
}{\endmanualtheoreminner}

\theoremstyle{plain}
\newtheorem{manuallemmainner}{Lemma}[lem]
\newenvironment{manuallemma}[1]{%
  \renewcommand\themanuallemmainner{#1}%
  \manuallemmainner
}{\endmanuallemmainner}

\theoremstyle{plain}

\theoremstyle{plain}
\newtheorem{theoremrpt}{\protect\theoremname}
\providecommand{\theoremname}{Theorem}

\theoremstyle{plain}
\newtheorem{lemmarpt}{\protect\lemmaname}
\providecommand{\lemmaname}{Lemma}

\theoremstyle{plain}

\providecommand{\corollaryname}{Corollary}

\newenvironment{talign}
{\align}
{\endalign}

\newenvironment{talign*}
{\csname align*\endcsname}
{\csname endalign*\endcsname}

\definecolor{AGHighlight}{RGB}{0,92,150}
\newcommand{\AGHighlightColor}{\color{AGHighlight}}

\DeclareMathOperator*{\argmax}{arg\,max}

\newcommand{\bydef}{\triangleq}

\newcommand{\dif}{\mathop{}\!\mathrm{d}}

\newcommand{\absval}[1]{\left|{#1}\right|}
\newcommand{\absvalflat}[1]{\lvert{#1}\rvert}

\newcommand{\normflat}[1]{\lVert{#1}\rVert}

\newcommand{\CondCurlyBrackets}[1]{\expandafter\ifx\expandafter\relax\detokenize{#1}\relax\else\left(#1\right)\fi}
\newcommand{\CondRectBrackets}[1]{\expandafter\ifx\expandafter\relax\detokenize{#1}\relax\else[#1]\fi}
\newcommand{\CondRectBracketsHigh}[1]{\expandafter\ifx\expandafter\relax\detokenize{#1}\relax\else\left[#1\right]\fi}
\newcommand{\CondColon}[1]{\expandafter\ifx\expandafter\relax\detokenize{#1}\relax\else:{#1}\fi}

\newcommand{\Expt}{\mathbb{E}}
\newcommand{\ExptFlat}[3]{\ensuremath{\Expt_{#1}^{#2}\CondRectBrackets{#3}}}

\newcommand{\tallet}[1]{\scalebox{1.2}{\MakeUppercase{#1}}}

\makeatletter
\@ifundefined{linenomath*}{
  \newenvironment{linenomath*}{}{}
}{}
\makeatother

%% file: appendix_content.tex
\allowdisplaybreaks[3]

\section{Proofs}
\label{sec:appendix_proofs}

\subsection{Lemma~\ref{lem:propagated_pb_likelihood_rpt}}

\begin{lemmarpt}
\label{lem:propagated_pb_likelihood_rpt}
Let $\bar{b} =((s^{j},\lambda^{j}))_{j=1}^{J}$, $\bar{b}^{\prime -} =((s^{\prime -,j},\lambda^{\prime -,j}))_{j=1}^{J}$ be ordered particle beliefs. The probability that $\bar{b}^{\prime -}$ is a propagated belief in the bootstrap filter, given $\bar{b}$ and action $a$, is:
\begin{linenomath*}
    \begin{talign}
        p(\bar{b}^{\prime -}{\mid} \bar{b},a)&=\prod_{j=1}^{J}p_T(s^{\prime -,j}{\mid} s^{j}, a),
        \label{eq:propagated_belief_likelihood_rpt}
    \end{talign}
\end{linenomath*}
whenever $\lambda^{\prime -,j}=\lambda^{j}$ for all $j=1,\dots,J$, and zero otherwise.
When $p(\bar{b}^{\prime -}{\mid} \bar{b},a)>0$ it holds that:
\begin{linenomath*}
    \begin{talign}
        \nabla_a \log p(\bar{b}^{\prime -}{\mid} \bar{b},a)&=\sum_{j=1}^{J} \nabla_a \log p_T(s^{\prime -,j}{\mid} s^{j}, a).
        \label{eq:propagated_belief_grad_rpt}
    \end{talign}
\end{linenomath*}
\end{lemmarpt}

\begin{proof}
In the bootstrap filter algorithm, each particle is sampled $s^{\prime -,j}\sim p_T(\cdot{\mid} s^j,a)$ independently for each $j=1,\ldots,J$.
Therefore:
\begin{linenomath*}
    \begin{gather}
        p(\bar{b}^{\prime -}{\mid} \bar{b},a)
        =p(((s^{\prime -,j},\lambda^{\prime -,j}))_{j=1}^{J} {\mid} ((s^{l},\lambda^{l}))_{l=1}^{J}, a),
    \end{gather}
\end{linenomath*}
due to the beliefs being ordered, and the independent sampling, we can factorize the terms into a product:
\begin{linenomath*}
    \begin{align}
        &=\prod_{i=1}^{J}p(((s^{\prime -,j},\lambda^{\prime -,j})) {\mid} ((s^{l},\lambda^{l}))_{l=1}^{J}, a) \\
        &=\prod_{i=1}^{J}p(((s^{\prime -,j},\lambda^{\prime -,j})) {\mid} (s^{j},\lambda^{j}), a).
    \end{align}
\end{linenomath*}
Since $\lambda^{\prime -,j}=\lambda^{j}$, we can drop the weights to obtain
\begin{linenomath*}
    \begin{gather}
        p(\bar{b}^{\prime -}{\mid} \bar{b},a)=\prod_{j=1}^{J}p_T(s^{\prime -,j}{\mid} s^{j}, a).
    \end{gather}
\end{linenomath*}
\end{proof}

\subsection{Theorem~\ref{thm:mdp_action_grad_rpt}}
\label{sec:appendix_action_score_gradient}

We first highlight that the required "well-behaved" conditions for the theorem are for changing the order of derivatives and integrals by the Leibniz integral rule.
Sufficient conditions are characterized by~\citeauthor[Theorem 2.27]{Folland99book}~[\citeyear{Folland99book}]:
\begin{manualtheorem}{2.27}
    \label{thm:folland_2.27}
Suppose that $f: X\times [a,b]\to \mathbb{C}(-\infty < a < b < \infty)$ and that $f(\cdot, t):X\to \mathbb{C}$ is integrable for each $t\in[a,b]$. Let $F(t)=\int_X f(x,t) \dif \mu (x)$.
\begin{enumerate}
    \item Suppose that there exists $g\in L^{1}(\mu)$ such that $\absvalflat{f(x,t)}\leq g(x)$ for all $x, t$. 
    If $\lim_{t\to t_0}f(x,t)=f(x,t_0)$ for every x, then $lim_{t\to t_0}F(t)=F(t_0)$; in particular, if $f(x,\cdot)$ is continuous for each x, then F is continuous.
    \item Suppose that $\partial f \slash \partial t$ exists and there is a $g\in L^{1}(\mu)$ such that $\absvalflat{\partial f \slash \partial t (x,t)}\leq g(x)$ for all $x, t$. Then $F$ is differentiable and $F^{\prime}=\int(\partial f \slash \partial t)(x,t)\dif \mu(x)$.
\end{enumerate}
\end{manualtheorem}
More general versions of the theorem have been proven~\cite{Cheng06}.
Since such generalizations are not central to our contribution, we present Theorem~\ref{thm:mdp_action_grad_rpt} under a set of sufficient but non-minimal assumptions.
We note that Lebesgue integrability may hold for a very broad class of functions, including functions with a countable set of discontinuities, and densities of common distributions such as regular and clipped Gaussians. This is important for many MDP and POMDP problem formulations, in which sparse rewards and discontinuous transition and observation models are often used.

For clarity, we present the results for a generic action $a\in\mathcal{A}$.
Unlike the main text, where $\breve{a}$ denotes an updated action relative to the sampling action, such a distinction is unnecessary here due to the use of general proposal distributions.

\begin{theoremrpt}
\label{thm:mdp_action_grad_rpt}
Assume: (i) $\rho^{p_T}_{q}(s^\prime)$, $\rho^{p}_{q}(\bar{b}^{\prime -})$ are well-defined; (ii) $r$, $V$ are bounded; (iii) there exist integrable $g_r,g_T$ w.r.t. the MDP/POMDP measures such that $\normflat{\nabla_a r}\leq g_r$ and $\normflat{\nabla_a \log p_T}\leq g_T$. For POMDPs, assume the assumptions of Lemma~\ref{lem:propagated_pb_likelihood} hold.
Then, from the dominated-convergence form of Leibniz integral rule~\cite[Theorem 2.27]{Folland99book}, the action score gradient satisfies:
\begin{linenomath*}
    \begin{align}
        &\nabla_{a}Q_{t}^{\pi}(s,a)
        =\mathbb{E}_{s^{\prime}{\mid} q}
        \Big[
            \rho^{p_T}_{q}(s^\prime)[\nabla_{a}\log p_{T}(s^\prime{\mid} s,a)
            \nonumber \\
            &\quad(r(s,a,s^{\prime})+\gamma V_{t+1}^{\pi}(s^{\prime}) - B(s))+\nabla_{a} r(s,a,s^{\prime})]
        \Big],
        \label{eq:mdp_action_grad_rpt}
        \\
        &\nabla_{a}Q_{t}^{\pi}(\bar{b},a)
        = \mathbb{E}_{\bar{b}^{\prime -},o^{\prime},\bar{b}^{\prime}{\mid} q}
        \Big[
            \rho^{p}_{q}(\bar{b}^{\prime -})
            [\nabla_{a}\log p(\bar{b}^{\prime -}{\mid} \bar{b},a)
            \nonumber \\
            &\quad(r(\bar{b},a,\bar{b}^{\prime})+\gamma V_{t+1}^{\pi}(\bar{b}^{\prime}) - B(\bar{b}))+\nabla_{a} r(\bar{b}, a, \bar{b}^{\prime})]
        \Big],
        \label{eq:pomdp_action_grad_rpt}
    \end{align}
    where $B(s)$ (resp. $B(\bar{b})$) is any baseline independent of $a$ for estimator variance reduction~\cite{Schulman16iclr}, e.g. $B(s)=V_{t}^{\pi}(s)$.
\end{linenomath*}
\end{theoremrpt}

\begin{proof}
Recall the definition of the importance weights 
\begin{linenomath*}
    \begin{gather}
        \rho^{p_T}_{q}(s^\prime) = \frac{p_T(s{^\prime}{\mid} s,a)}{q(s{^\prime}{\mid} s,a)},
        \quad
        \rho^{p}_{q}(\bar{b}^{\prime -}) = \frac{p(\bar{b}^{\prime -}{\mid} \bar{b},a)}{q(\bar{b}^{\prime -}{\mid} \bar{b},a)}.
    \end{gather}
\end{linenomath*}

\paragraph{MDP Case, $B(s)=0$.}
In this case, the action-value function is given by:
\begin{linenomath*}
    \begin{align}
        &\nabla_{a}Q_{t}^{\pi}(s,a)
        \nonumber \\
        &\begin{multlined}[b][0.8\displaywidth]
            = \nabla_{a}\mathbb{E}_{s^{\prime}{\mid} q}\left[\rho^{p_T}_{q}\left(s^{\prime}\right)(r\left(s,a,s^{\prime}\right)+\gamma V_{t+1}^{\pi}\left(s^{\prime}\right))\right] 
        \end{multlined}
        \\
        &\begin{multlined}[b][0.8\displaywidth]
            = \nabla_{a}\int_{s^{\prime}}q(s^{\prime}{\mid} s,a)\frac{p_{T}(s^{\prime}{\mid} s,a)}{q(s^{\prime}{\mid} s,a)}
            \\
            \cdot (r\left(s,a,s^{\prime}\right)+\gamma V_{t+1}^{\pi}\left(s^{\prime}\right)) \dif s^\prime
        \end{multlined}
        \\
        &\begin{multlined}[b][0.8\displaywidth]
            = \nabla_{a}\int_{s^{\prime}}p_{T}(s^{\prime}{\mid} s,a)(r\left(s,a,s^{\prime}\right)+\gamma V_{t+1}^{\pi}\left(s^{\prime}\right)) \dif s^\prime.
        \end{multlined}
    \end{align}
\end{linenomath*}

From the assumption that $r$ and $V$ are bounded, hence their sum is also bounded by some $M>0$:
\begin{gather}
    \absval{r\left(s,a,s^{\prime}\right)+\gamma V_{t+1}^{\pi}\left(s^{\prime}\right)} \leq M. \label{eq:mdp_obj_bounded}
\end{gather}

For each $a_{i}$ (a coordinate of $a\in\mathcal{A}$), the partial derivative satisfies
\begin{linenomath*}
    \begin{align}
        & \frac{\partial}{\partial a_i} \Big(p_{T}(s^{\prime}{\mid} s,a)(r\left(s,a,s^{\prime}\right)+\gamma V_{t+1}^{\pi}\left(s^{\prime}\right))\Big)
        \nonumber \\
        &\begin{multlined}[b][0.8\displaywidth]
            = \frac{\partial}{\partial a_i}p_{T}(s^{\prime}{\mid} s,a)\cdot (r\left(s,a,s^{\prime}\right)+\gamma V_{t+1}^{\pi}\left(s^{\prime}\right))
            \\
            + p_{T}(s^{\prime}{\mid} s,a) \cdot \frac{\partial}{\partial a_i}r\left(s,a,s^{\prime}\right),
        \end{multlined}
        \label{eq:deriv_of_mdp_obj_1}
    \end{align}
\end{linenomath*}
where importantly, $\frac{\partial}{\partial a_i}V_{t+1}^{\pi}\left(s^{\prime}\right)=0$ due to $s^\prime$ being an integration variable that is independent of $a$.
We can assume that $p_T(s^{\prime}{\mid} s,a) > 0$, as the set of $s^\prime$ where $p_T(s^{\prime}{\mid} s,a) = 0$ does not contribute to the integral.
Thus, we can apply the log-gradient transform to obtain:
\begin{linenomath*}
    \begin{align}
        &\eqref{eq:deriv_of_mdp_obj_1}
        \nonumber \\
        &\begin{multlined}[b][0.8\displaywidth]
            = p_{T}(s^{\prime}{\mid} s,a) \cdot \frac{\partial}{\partial a_i}\log p_{T}(s^{\prime}{\mid} s,a)\cdot (r(s,a,s^{\prime})
            \\
            +\gamma V_{t+1}^{\pi}(s^{\prime}))
            + p_{T}(s^{\prime}{\mid} s,a) \cdot \frac{\partial}{\partial a_i}r\left(s,a,s^{\prime}\right),
        \end{multlined}
        \\
        &\begin{multlined}[b][0.8\displaywidth]
            = p_{T}(s^{\prime}{\mid} s,a) \cdot \big(\frac{\partial}{\partial a_i}\log p_{T}(s^{\prime}{\mid} s,a)
            \\
            \cdot (r(s,a,s^{\prime})+\gamma V_{t+1}^{\pi}(s^{\prime}))
            + \frac{\partial}{\partial a_i}r(s,a,s^{\prime})\big).
        \end{multlined}
        \label{eq:deriv_of_mdp_obj_2}
    \end{align}
\end{linenomath*}
By our assumptions, it follows that there exist integrable $g_1(s^\prime)$, $g_2(s^\prime)$ w.r.t. $p_T$, for which it holds that $\frac{\partial}{\partial a_{i}}\log p_{T}(s^{\prime}{\mid} s,a)\leq g_1(s^\prime)$ and $\frac{\partial}{\partial a_{i}}r\left(s,a,s^{\prime}\right)\leq g_2(s^\prime)$ for all $s$, $a$, $s^\prime$.
Therefore, there exists an integrable envelope $g(s^\prime)\bydef M g_1(s^\prime) + g_2(s^\prime)$ for \eqref{eq:deriv_of_mdp_obj_2}:
\begin{linenomath*}
    \begin{align}
        &\begin{multlined}[b][0.8\displaywidth]
            \Big| p_{T}(s^{\prime}{\mid} s,a) \cdot \big(\frac{\partial}{\partial a_i}\log p_{T}(s^{\prime}{\mid} s,a)
            \\
            \cdot (r(s,a,s^{\prime})+\gamma V_{t+1}^{\pi}(s^{\prime}))
            + \frac{\partial}{\partial a_i}r(s,a,s^{\prime})\big) \Big|
        \end{multlined}
        \nonumber \\
        &\leq p_{T}(s^{\prime}{\mid} s,a) \left( M g_1(s^\prime) + g_2(s^\prime)\right)
        \\
        &= p_{T}(s^{\prime}{\mid} s,a) g(s^\prime).
    \end{align}
\end{linenomath*}
Summation and scalar multiplication preserve integrability, and it follows that $g(s^\prime)$ is integrable w.r.t. $p_T$~\cite{Folland99book}.
Therefore, from Theorem~\ref{thm:folland_2.27} (clause 2), we can apply the Leibniz integral rule (in vector form):
\begin{linenomath*}
    \begin{align}
        & \nabla_{a}\int_{s^{\prime}}p_{T}(s^{\prime}{\mid} s,a)(r(s,a,s^{\prime})+\gamma V_{t+1}^{\pi}(s^{\prime})) \dif s^\prime
        \\
        &= \int_{s^{\prime}} \nabla_{a} \big( p_{T}(s^{\prime}{\mid} s,a)(r(s,a,s^{\prime})+\gamma V_{t+1}^{\pi}(s^{\prime})) \big) \dif s^\prime
        \nonumber \\
        &\begin{multlined}[b][0.8\displaywidth]
            = \int_{s^\prime} p_{T}(s^{\prime}{\mid} s,a) \big( \nabla_{a}\log p_{T}(s^{\prime}{\mid} s,a)
            \\
            \cdot (r(s,a,s^{\prime})
            +\gamma V_{t+1}^{\pi}(s^{\prime}))
            + \nabla_{a}r(s,a,s^{\prime}) \big) \dif s^\prime.
        \end{multlined}
        \label{eq:deriv_of_mdp_obj_3}
    \end{align}
\end{linenomath*}
By bringing back the importance ratio, we obtain the required result \eqref{eq:mdp_action_grad_rpt}:
\begin{linenomath*}
    \begin{align}
    & \eqref{eq:deriv_of_mdp_obj_3}
    \nonumber \\
    &\begin{multlined}[b][0.8\displaywidth]
        = \int_{s^\prime} q(s^{\prime} {\mid} s,a) \frac{p_T(s{^\prime}{\mid} s,a)}{q(s^{\prime} {\mid} s,a)} \big( \nabla_{a}\log p_{T}(s^\prime{\mid} s,a)
        \\
        (r(s,a,s^{\prime}) + \gamma V_{t+1}^{\pi}(s^{\prime})) + \nabla_{a}r(s,a,s^\prime) \big) \dif s^\prime 
    \end{multlined}
    \\
    &\begin{multlined}[b][0.8\displaywidth]
        = \mathbb{E}_{s^{\prime}{\mid} q}\Big[\rho^{p_T}_{q}(s^\prime)
        [\nabla_{a}\log p_{T}(s^\prime{\mid} s,a)
        (r(s,a,s^{\prime})
        \\
        +\gamma V_{t+1}^{\pi}(s^{\prime}))+\nabla_{a}r(s,a,s^\prime)]\Big] .
    \end{multlined}
    \end{align}
\end{linenomath*}

\paragraph{MDP Case, $B(s)\neq0$.}
Let $F_t^\pi(s,a)\bydef Q_t^\pi(s,a) - B(s)$.
On the one hand, it holds that:
\begin{linenomath*}
    \begin{align}
        &\nabla_a F_t^\pi(s,a)
        \nonumber \\
        &= \nabla_a Q_t^\pi(s,a) - \nabla_a B(s)
        \\
        &= \nabla_a Q_t^\pi(s,a),
    \end{align}
\end{linenomath*}
as $B(s)$ is independent of $a$.
On the other hand, we can apply the steps of the previous case to $F_t^\pi(s,a)$, resulting in:
\begin{linenomath*}
    \begin{align}
        &\nabla_{a}F_{t}^{\pi}(s,a)
        \nonumber \\
        &\begin{multlined}[b][0.8\displaywidth]
            = \nabla_{a}\mathbb{E}_{s^{\prime}{\mid} s,a}\Big[(r\left(s,a,s^{\prime}\right)+\gamma V_{t+1}^{\pi}\left(s^{\prime}\right) - B(s))\Big]
        \end{multlined}
        \\
        &\begin{multlined}[b][0.8\displaywidth]
            = \mathbb{E}_{s^{\prime}{\mid} s,a}\Big[\nabla_{a}\log p_T(s^\prime{\mid} s,a)
            \\
            \cdot (r\left(s,a,s^{\prime}\right)+\gamma V_{t+1}^{\pi}\left(s^{\prime}\right) - B(s)) + \nabla_{a}r\left(s,a,s^{\prime}\right)\Big]
        \end{multlined}
        \\
        &\begin{multlined}[b][0.8\displaywidth]
            = \mathbb{E}_{s^{\prime}{\mid} q}
            \Big[
                \rho^{p_T}_{q}(s^\prime)[\nabla_{a}\log p_{T}(s^\prime{\mid} s,a)
                \\
                \cdot (r(s,a,s^{\prime})+\gamma V_{t+1}^{\pi}(s^{\prime}) - B(s))+\nabla_{a} r(s,a,s^{\prime})]
            \Big].
        \end{multlined}
    \end{align}
\end{linenomath*}
By equating the two expressions for $\nabla_a F_t^\pi(s,a)$, we obtain the required result.

\paragraph{POMDP case.}
We shall prove for $B(\bar{b}=0)$, as the case $B(\bar{b}\neq 0)$ follows similarly to the MDP case.

We begin by showing propagated belief decomposition applied to the action-value function:
\begin{linenomath*}
    \begin{align}
        & Q_t^\pi(\bar{b},a) 
        \nonumber \\
        &= \mathbb{E}_{\bar{b}^\prime {\mid} \bar{b},a}\left[r(\bar{b},a,\bar{b}^\prime) + \gamma V_{t+1}^\pi(\bar{b}^\prime)\right]
        \\
        &= \int_{\bar{b}^\prime} p(\bar{b}^\prime {\mid} \bar{b},a) (r(\bar{b},a,\bar{b}^\prime) + \gamma V_{t+1}^\pi(\bar{b}^\prime)) \dif \bar{b}^\prime,
    \end{align}
\end{linenomath*}
and we continue by marginalizing over the propagated belief $\bar{b}^{\prime -}$ and the observation $o^{\prime}$:
\begin{linenomath*}
    \begin{align}
        &\begin{multlined}[b][0.8\displaywidth]
            = \int_{\bar{b}^{\prime -}} \int_{o^{\prime}} \int_{\bar{b}^\prime} p(\bar{b}^{\prime -}, o^{\prime}, \bar{b}^\prime {\mid} \bar{b},a) (r(\bar{b},a,\bar{b}^\prime)
            \\
            + \gamma V_{t+1}^\pi(\bar{b}^\prime)) \dif \bar{b}^\prime \dif o^{\prime} \dif \bar{b}^{\prime -}.
        \end{multlined}
    \end{align}
\end{linenomath*}
By its definition, the propagated belief $\bar{b}^{\prime -}$ is the generated belief by conditioning on the history without the last observation.
Therefore, by using the chain rule and removing variables that are conditionally independent, we obtain:
\begin{linenomath*}
    \begin{align}
        & Q_t^\pi(\bar{b},a)
        \nonumber \\
        &\begin{multlined}[b][0.8\displaywidth]
            = \int_{\bar{b}^{\prime -}} \int_{o^{\prime}} \int_{\bar{b}^\prime} p(\bar{b}^\prime {\mid} o^{\prime},\bar{b}^{\prime -},\cancel{\bar{b},a}) p(o^{\prime} {\mid} \bar{b}^{\prime -},\cancel{\bar{b},a})
            \\
            p(\bar{b}^{\prime -} {\mid} \bar{b},a) (r(\bar{b},a,\bar{b}^\prime) + \gamma V_{t+1}^\pi(\bar{b}^\prime)) \dif \bar{b}^\prime \dif o^{\prime} \dif \bar{b}^{\prime -}
        \end{multlined}
        \\
        &\begin{multlined}[b][0.8\displaywidth]
            = \int_{\bar{b}^{\prime -}} \int_{o^{\prime}} \int_{\bar{b}^\prime} p(\bar{b}^\prime {\mid} o^{\prime},\bar{b}^{\prime -}) p(o^{\prime} {\mid} \bar{b}^{\prime -}) p(\bar{b}^{\prime -} {\mid} \bar{b},a)
            \\
            (r(\bar{b},a,\bar{b}^\prime) + \gamma V_{t+1}^\pi(\bar{b}^\prime)) \dif \bar{b}^\prime \dif o^{\prime} \dif \bar{b}^{\prime -}
        \end{multlined}
        \\
        &= \mathbb{E}_{\bar{b}^{\prime -} {\mid} \bar{b},a} \mathbb{E}_{o^{\prime} {\mid} \bar{b}^{\prime -}} \mathbb{E}_{\bar{b}^\prime {\mid} \bar{b}^{\prime -},o^{\prime}}
        \left[r(\bar{b},a,\bar{b}^\prime) + \gamma V_{t+1}^\pi(\bar{b}^\prime)\right].
    \end{align}
\end{linenomath*}
We introduce the importance ratio $\rho^{p}_{q}(\bar{b}^{\prime -})\bydef \frac{p(\bar{b}^{\prime -} {\mid} \bar{b},a)}{q(\bar{b}^{\prime -} {\mid} \bar{b},a)}$ over the propagated belief:
\begin{linenomath*}
    \begin{align}
        & Q_t^\pi(\bar{b},a)
        \nonumber \\
        &\begin{multlined}[b][0.8\displaywidth] =
            \mathbb{E}_{\bar{b}^{\prime -} {\mid} \bar{b},a} \mathbb{E}_{o^{\prime} {\mid} \bar{b}^{\prime -}} \mathbb{E}_{\bar{b}^{\prime} {\mid} \bar{b}^{\prime -},o^{\prime}} \Big[
            \\
            \frac{q(\bar{b}^{\prime -} {\mid} \bar{b},a)}{q(\bar{b}^{\prime -} {\mid} \bar{b},a)} (r(\bar{b},a,\bar{b}^\prime) + \gamma V_{t+1}^\pi(\bar{b}^\prime)) \Big],
        \end{multlined}
        \\
        &\begin{multlined}[b][0.8\displaywidth] =
            \mathbb{E}_{\bar{b}^{\prime -} {\mid} q} \mathbb{E}_{o^{\prime} {\mid} \bar{b}^{\prime -}} \mathbb{E}_{\bar{b}^\prime {\mid} \bar{b}^{\prime -},o^{\prime}}[
            \rho^{p}_{q}(\bar{b}^{\prime -})
            (r(\bar{b},a,\bar{b}^\prime) + \gamma V_{t+1}^\pi(\bar{b}^\prime)) ],
        \end{multlined}
    \end{align}
\end{linenomath*}
Similarly to the derivation of the MDP case, we can apply the Leibniz integral rule to obtain the result:
\begin{linenomath*}
    \begin{align}
        & \nabla_{a}Q_{t}^{\pi}(\bar{b},a)
        \nonumber \\
        &\begin{multlined}[b][0.8\displaywidth] =
            \mathbb{E}_{\bar{b}^{\prime -} {\mid} q} \mathbb{E}_{o^{\prime} {\mid} \bar{b}^{\prime -}} \mathbb{E}_{\bar{b}^\prime {\mid} \bar{b}^{\prime -},o^{\prime}} \Big[\rho^{p}_{q}(\bar{b}^{\prime -})
            \Big(\nabla_{a}\log p(\bar{b}^{\prime -}{\mid} \bar{b},a)
            \\
            \cdot (r(\bar{b},a,\bar{b}^{\prime})+\gamma V_{t+1}^{\pi}(\bar{b}^{\prime}))
            +\nabla_{a}r(\bar{b},a,\bar{b}^{\prime})\Big)\Big].
        \end{multlined}
    \end{align}
\end{linenomath*}
\end{proof}

\subsection{Theorem~\ref{thm:action_grad_state_reward_rpt}}

This theorem was not presented in the main paper due to space limitations.
It is an alternate form of Theorem~\ref{thm:mdp_action_grad_rpt} for MDPs and state-reward POMDPs, and it was used in the gradient calculation of some of the experimental domains, hence it is brought here for completeness.

Theorem~\ref{thm:action_grad_state_reward_rpt} modifies the immediate reward gradient for MDPs and state-reward POMDPs, by basing it on state samples from the updated action rather than scoring previous value estimates.
It may give more accurate gradient estimates, albeit possibly more expensive to compute.

We refer to a \textit{state-reward POMDP} when the reward function can be expressed as $r(b,a,b^\prime)=\mathbb{E}_{s,s^\prime{\mid} b,a,b^\prime}[r(s,a,s^\prime)]$. 

We highlight that as an extension of Theorem~\ref{thm:mdp_action_grad_rpt}, it requires the same conditions.
Additionally, for the simplification of the immediate reward expectation, we make use of Fubini's theorem for changing integration orders.
Sufficient conditions are that $r(s,a,s^\prime)$ is always integrable w.r.t. the POMDP measure.

\begin{manualtheorem}{1 Extended (Immediate Reward Gradient)}
    \label{thm:action_grad_state_reward_rpt}
    Under the conditions of Theorem~\ref{thm:mdp_action_grad_rpt}, the action-gradient of the MDP action-value function is given by:
    \begin{linenomath*}
        \begin{align}
            & \nabla_{a}Q_{t}^{\pi}(s,a) 
            \nonumber \\
            &= \mathbb{E}_{s^{\prime}{\mid} s,a} 
                \Big[ 
                    \nabla_{a}\log p_T(s^{\prime}{\mid} s,a) \cdot r(s,a,s^{\prime})
                    + \nabla_{a}r(s,a,s^{\prime})
                \Big] 
            \nonumber \\
            &\begin{multlined}[b][0.8\displaywidth]
                + \mathbb{E}_{s^{\prime}{\mid} q} 
                \Big[ 
                    \rho^{p_T}_{q}(s^\prime)
                    \cdot \nabla_{a}\log p_T(s^{\prime}{\mid} s,a) 
                    \cdot ( \gamma V_{t+1}^{\pi}(s^{\prime}) - B(s) )
                    \Big] .
            \end{multlined}
                \label{eq:mdp_state_grad_rpt}
        \end{align}
    \end{linenomath*}
    For a state-reward POMDP, the action-gradient of the action-value function is given by:
    \begin{linenomath*}
        \begin{align}
            &\nabla_{a}Q_{t}^{\pi}(\bar{b},a) 
            \nonumber \\
            &\begin{multlined}[b][0.8\displaywidth] =
                \mathbb{E}_{s{\mid} \bar{b}}\mathbb{E}_{s^{\prime}{\mid} s,a} 
                \Big[
                    \nabla_{a}\log p_T(s^{\prime}{\mid} s,a) \cdot r(s,a,s^{\prime})
                    \\
                    + \nabla_{a}r(s,a,s^{\prime})
                \Big]
            \end{multlined}
            \nonumber \\
            &\begin{multlined}[b][0.8\displaywidth]
                + \mathbb{E}_{\bar{b}^{\prime -}{\mid} q}\mathbb{E}_{o{\mid} \bar{b}^{\prime -}}\mathbb{E}_{\bar{b}^{\prime}{\mid} \bar{b}^{\prime -},o} 
                \Big[
                    \rho^{p}_{q}(\bar{b}^{\prime -}) 
                    \\
                    \cdot \nabla_{a}\log p(\bar{b}^{\prime -}{\mid} \bar{b},a)\cdot  (\gamma V_{t+1}^{\pi}(\bar{b}^{\prime}) - B(\bar{b}))
                \Big].
            \end{multlined}
            \label{eq:pomdp_state_grad_rpt}
        \end{align}
    \end{linenomath*}
\end{manualtheorem}

\begin{proof}
    The central part of this proof is the decomposition of the belief action-value function to an expectation over state trajectories of immediate rewards, and an expectation over belief trajectories of future values.
    Afterwards, we apply the Leibniz integral rule similarly to Theorem~\ref{thm:mdp_action_grad_rpt}.
    We show the derivation of the POMDP case, as the MDP case is a simplified version of it.

    Similarly to Theorem~\ref{thm:mdp_action_grad_rpt}, for any baseline $B(\bar{b})$ that is independent of $a$, we have:
    \begin{linenomath*}
        \begin{align}
            & \nabla_{a}(Q_{t}^{\pi}(\bar{b},a) - B(\bar{b}))
            \nonumber \\
            &= \nabla_{a}Q_{t}^{\pi}(\bar{b},a)
            \\
            &= \nabla_{a}\mathbb{E}_{\bar{b}^\prime {\mid} \bar{b},a}\left[r(\bar{b},a,\bar{b}^\prime) + \gamma V_{t+1}^\pi(\bar{b}^\prime) - B(\bar{b})\right],
        \end{align}
    \end{linenomath*}
    as $B(\bar{b})$ is independent of $a$ and it holds that $\nabla_a B(\bar{b}) = 0$.
    We rewrite the belief action-value function:
    \begin{linenomath*}
        \begin{align}
            &Q_t^\pi(\bar{b},a) - B(\bar{b})
            \nonumber \\
            &= \mathbb{E}_{\bar{b}^\prime {\mid} \bar{b},a}\left[r(\bar{b},a,\bar{b}^\prime) + \gamma V_{t+1}^\pi(\bar{b}^\prime)\right] - B(\bar{b}) \\
            &= \mathbb{E}_{\bar{b}^\prime {\mid} \bar{b},a}\left[r(\bar{b},a,\bar{b}^\prime)\right] + \mathbb{E}_{\bar{b}^\prime {\mid} \bar{b},a}\left[\gamma V_{t+1}^\pi(\bar{b}^\prime) - B(\bar{b})\right].
        \end{align}
    \end{linenomath*}
    Expanding the first expectation, and using the chain rule and then Fubini's theorem to cancel the expectation over the observation and posterior belief, we obtain:
    \begin{linenomath*}
        \begin{align}            
            & \mathbb{E}_{\bar{b}^\prime {\mid} \bar{b},a}\left[r(\bar{b},a,\bar{b}^\prime)\right]
            \nonumber \\
            &= \mathbb{E}_{\bar{b}^{\prime -} {\mid} \bar{b},a} \mathbb{E}_{o {\mid} \bar{b}^{\prime -}} \mathbb{E}_{\bar{b}^\prime {\mid} \bar{b}^{\prime -},o} \left[ \mathbb{E}_{s,s^\prime{\mid} \bar{b},a,\bar{b}^\prime}[r(s,a,s^\prime)]\right]
            \\
            &\begin{multlined}[b][0.8\displaywidth]
                = \int_{\bar{b}^{\prime -}} \int_{o} \int_{\bar{b}^\prime} \int_{s} \int_{s^\prime} p(s,s^\prime, o, \bar{b}^{\prime -},\bar{b}^\prime {\mid} \bar{b},a)
                \\
                r(s,a,s^\prime) \dif s^\prime \dif s \dif \bar{b}^\prime \dif o \dif \bar{b}^{\prime -}        
            \end{multlined} 
            \\
            &\begin{multlined}[b][0.8\displaywidth]
                = \int_{\bar{b}^{\prime -}} \int_{o} \int_{\bar{b}^\prime} \int_{s} \int_{s^\prime} p(o,\bar{b}^\prime,\bar{b}^{\prime -} {\mid} s,s^\prime,\bar{b},a) p(s,s^\prime {\mid} \bar{b},a)            
                \\
                r(s,a,s^\prime) \dif s^\prime \dif s \dif \bar{b}^\prime \dif o \dif \bar{b}^{\prime -}
            \end{multlined}
            \\
            &\begin{multlined}[b][0.8\displaywidth]
                = \int_{s} \int_{s^\prime} p(s,s^\prime {\mid} \bar{b},a) r(s,a,s^\prime)
                \\
                \cancel{\left(\int_{\bar{b}^{\prime -}} \int_{o} \int_{\bar{b}^\prime} p(o,\bar{b}^\prime,\bar{b}^{\prime -} {\mid} s,s^\prime,\bar{b},a) \dif \bar{b}^\prime \dif o \dif \bar{b}^{\prime -} \right)}
                \dif s^\prime \dif s
            \end{multlined}
            \\
            &= \int_{s} \int_{s^\prime} p(s^\prime {\mid} s,a,\cancel{\bar{b}}) p(s{\mid} \bar{b},\cancel{a}) r(s,a,s^\prime) \dif s^\prime \dif s
            \\
            &= \mathbb{E}_{s{\mid} \bar{b}}\mathbb{E}_{s^{\prime}{\mid} s,a} [r(s,a,s^{\prime})].
        \end{align}
    \end{linenomath*}
    Now, we apply $\nabla_{a}$ on the obtained terms.
    The gradient of the immediate reward expectation, following steps similar to Theorem~\ref{thm:mdp_action_grad_rpt}, is given by:
    \begin{linenomath*}
        \begin{align}
            & \nabla_{a}\mathbb{E}_{s{\mid} \bar{b}}\mathbb{E}_{s^{\prime}{\mid} s,a} \left[r(s,a,s^{\prime})\right] \nonumber \\
            &\begin{multlined}[b][0.8\displaywidth] =
                \mathbb{E}_{s{\mid} \bar{b}}\mathbb{E}_{s^{\prime}{\mid} s,a} \Big[\Big(\nabla_{a} \log p_T(s^\prime {\mid} s,a)
                \cdot r(s,a,s^{\prime}) 
                \\
                + \nabla_{a}r(s,a,s^{\prime})\Big)\Big].
            \end{multlined}
        \end{align}
    \end{linenomath*}
    Next, the gradient of the future value expectation, when adding the importance ratio, is calculated by:
    \begin{linenomath*}
        \begin{align}
            & \nabla_{a} \mathbb{E}_{\bar{b}^\prime {\mid} \bar{b},a}\Big[\gamma V_{t+1}^\pi(\bar{b}^\prime)- B(\bar{b}) \Big]
            \nonumber \\
            &\begin{multlined}[b][0.8\displaywidth]
                = \nabla_{a} \mathbb{E}_{\bar{b}^{\prime -}{\mid} \bar{b}, a}\mathbb{E}_{o{\mid} \bar{b}^{\prime -}}\mathbb{E}_{\bar{b}^{\prime}{\mid} \bar{b}^{\prime -},o} \Big[\frac{q(\bar{b}^{\prime -}{\mid} \bar{b},a)}{q(\bar{b}^{\prime -}{\mid} \bar{b},a)} 
                \\
                \cdot (\gamma V_{t+1}^\pi(\bar{b}^\prime)- B(\bar{b})) \Big]
            \end{multlined}
            \\
            &\begin{multlined}[b][0.8\displaywidth]
                =\mathbb{E}_{\bar{b}^{\prime -}{\mid} q}\mathbb{E}_{o{\mid} \bar{b}^{\prime -}}\mathbb{E}_{\bar{b}^{\prime}{\mid} \bar{b}^{\prime -},o} \Big[\rho^{p}_{q}(\bar{b}^{\prime -})
                \\
                \cdot \nabla_{a}\log p(\bar{b}^{\prime -}{\mid} \bar{b},a) \cdot (\gamma V_{t+1}^\pi(\bar{b}^\prime)- B(\bar{b})) \Big].
            \end{multlined}
        \end{align}
    \end{linenomath*}
    Combining the two results, we obtain the final expression \eqref{eq:pomdp_state_grad_rpt}.
\end{proof}

\subsection{Theorem~\ref{thm:mis_tree_rpt}}

We first start giving the formal statement of Lemma 9.1 by~\citeauthor{Veach95siggraph}~[\citeyear{Veach95siggraph}]:
\begin{manuallemma}{9.1}
    Let $F$ be any estimator of the form
    \begin{linenomath*}
        \begin{gather}
            F=\sum_{i=1}^{n}\frac{1}{n_i} \sum_{j=1}^{n_i} w_i(X_{i,j})\frac{f(X_{i,j})}{p_i(X_{i,j})}, \nonumber
        \end{gather}
    \end{linenomath*}
    where $n_i\geq 1$ for all $i$, and the weighting functions $w_i$ satisfy conditions 
    \begin{linenomath*}
        \begin{gather}
            \text{(\hyperlink{assum:mis-i}{MIS-I})}\quad \sum_{i=1}^{n}w_i(x)=1 \quad \text{whenever $f(x)\neq 0$}, \nonumber \\
            \text{(\hyperlink{assum:mis-ii}{MIS-II})}\quad w_i(x)=0 \quad \text{whenever $p_i(x)= 0$} \nonumber.
        \end{gather}
    \end{linenomath*}
    
\end{manuallemma}
The proof follows from the linearity of the expectation.

An important corollary brought by~\citeauthor{Veach95siggraph}~[\citeyear{Veach95siggraph}] is that these conditions imply that at any point where $f(x)\neq 0$, at least one of the $p_i(x)$ must be positive.
Thus, it is not necessary for every $p_i$ to sample the whole domain, and it is allowable for some $p_i$ to be specialized sampling techniques that concentrate on specific regions of the integrand.

We recall that our target function to estimate is given by the equations:
\begin{linenomath*}
    \begin{align}
        Q_{t}^{\pi}(s,a) = \mathbb{E}_{s^{\prime}{\mid} q}
        [\rho^{p_T}_{q}(s^\prime)
        (r(s,a,s^{\prime})+\gamma V_{t+1}^{\pi}(s^{\prime}))] \\
        = \mathbb{E}_{s^{\prime}{\mid} q}
        [\frac{p_T(s{^\prime}{\mid} s,a)}{q(s{^\prime}{\mid} s,a)}
        (r(s,a,s^{\prime})+\gamma V_{t+1}^{\pi}(s^{\prime}))].
        \label{eq:mdp_reuse_rpt}
    \end{align}
\end{linenomath*}
Hence, for each successor branch in $s^{\prime,i}\in \mathcal{S}_{sa}$, the target function to estimate is the same $p_T(s^{\prime,i}{\mid} s,a)(r(s,a,s^{\prime,i})+\gamma V_{t+1}^{\pi}(s^{\prime,i}))$, but with the proposal distribution $q(s^{\prime,i} {\mid} s,a) = p_T(s^{\prime,i} {\mid} s,a_{\textnormal{prop}}^{i})$.
We denote the branch importance ratio:
\begin{align}
    \rho_{a}^{i}(s^{\prime}) = \frac{p_T(s^{\prime}{\mid} s,a)}{p_T(s^{\prime} {\mid} s,a_{\textnormal{prop}}^{i})},
\end{align}
and we define the following MIS estimators of a single sample of each proposal distribution:
\begin{linenomath*}
    \begin{gather}
        \hat{V}_{f}(s,a) \bydef \sum_{i=1}^{\absval{\mathcal{S}_{sa}}} w_i(s^{\prime,i}) \rho_{a}^{i}(s^{\prime,i}) \hat{V}(s^{\prime,i}) , \\
        \hat{r}(s,a) \bydef \sum_{i=1}^{\absval{\mathcal{S}_{sa}}} w_i(s^{\prime,i}) \rho_{a}^{i}(s^{\prime,i}) r(s, a, s^{\prime,i}),
    \end{gather}
\end{linenomath*}
for weighting functions $w_i(s^{\prime,i})$ that may in general depend on $s$, $a$, $s^\prime$ and $a$.

We've defined the value estimate of each state node recursively as:
\begin{linenomath*}
    \begin{gather}
        \hat{V}(s) \bydef \sum_{a \in C(s)}\frac{n(s,a)}{n(s)}\hat{Q}(s,a),
    \end{gather}
\end{linenomath*}
unless $s^\prime$ is at the maximum tree depth, in which case its value estimate is the mean of the rollout values:
\begin{linenomath*}
    \begin{gather}
        \hat{V}(s)=\frac{1}{n(s)_{+1}}\sum_{i=1}^{n(s)_{+1}}v^{i}, \\
    \end{gather}
\end{linenomath*}
for all $v^{i}$ rollout values from the terminal node $s$.

\begin{theoremrpt}
    \label{thm:mis_tree_rpt}
    If $w_i(s^{\prime})$ satisfy requirements (\hyperlink{assum:mis-i}{MIS-I}) and (\hyperlink{assum:mis-ii}{MIS-II}) (see Section~\ref{sec:background}), then the MIS Tree yields unbiased value and action-value estimates for a given tree structure.
\end{theoremrpt}

\begin{proof}
    The proof goes by a mutual induction over $\hat{V}$ and $\hat{Q}$, proving that a node's value or action-value estimate is unbiased if all its children have unbiased estimates.

    \paragraph{Base case: leaf state nodes are unbiased.}
    As stated in the tree construction, $\hat{V}(s)$ when $s$ is a leaf node is the mean of the rollout values, and therefore:
    \begin{linenomath*}
        \begin{align}
            &\mathbb{E}_{s\sim \pi_{rollout}}\left[\hat{V}(s)\right]
            \nonumber \\
            &= \mathbb{E}_{s\sim \pi_{rollout}}\left[\frac{1}{n(s)_{+1}}\sum_{i=1}^{n(s)_{+1}}v^{i}\right]
            \\
            &= \frac{1}{n(s)_{+1}}\sum_{i=1}^{n(s)_{+1}}\mathbb{E}_{s\sim \pi_{rollout}}\left[v^{i}\right] 
            \\
            &= V^{\pi_{rollout}}(s).
        \end{align}
    \end{linenomath*}
    Hence, it is unbiased.

    \textbf{Inductive step 1: Value estimates $\hat{V}$ are unbiased.} \\
    \textbf{Induction hypothesis}: Assume that $\hat{Q}(s,a^{i})$ is unbiased, i.e. $\mathbb{E}\left[\hat{Q}(s,a^{i})\right] = Q(s,a^{i})$, for all $a^{i} \in \mathcal{A}_{s}$.

    The value estimate is defined as the mean of the action-value estimates weighted by the number of times each action was selected.
    Therefore, for the policy representing the current visitation counters, i.e. defined by $\pi_{visits}(s) \stackrel{\triangle}{\sim} n(s,a)\slash n(s)$, the value estimate satisfies:
    \begin{linenomath*}
        \begin{align}
            &\mathbb{E}\left[\hat{V}(s)\right]
            \nonumber \\
            &= \mathbb{E}\left[\sum_{i=1}^{\absval{\mathcal{A}_{s}}}\frac{n(s,a^{i})}{n(s)}\hat{Q}(s,a^{i})\right] 
            \\
            &= \sum_{i=1}^{\absval{\mathcal{A}_{s}}}\frac{n(s,a^{i})}{n(s)}\mathbb{E}\left[\hat{Q}(s,a^{i})\right]
            \\
            &= \sum_{i=1}^{\absval{\mathcal{A}_{s}}} p(a^{i} {\mid} \pi_{visits}(s)) Q(s,a^{i})
            \\
            &= \mathbb{E}_{a\sim \pi_{visits}(s)} \left[Q(s,a)\right]
            \\
            &= V^{\pi_{visits}}(s),
        \end{align}
    \end{linenomath*}
    where we used the definition of $\pi_{visits}$ and the induction hypothesis.
    
    This explains the importance of "given tree structure" in the theorem statement - when changing the visitation counters and the according action selection policy at each state node, it results in a different weighting of action-value estimates, and hence could introduce a bias when considering the action-value function of a different policy at the root node of the search tree.

    \textbf{Inductive step 2: Action-value estimates $\hat{Q}$ are unbiased.} \\
    \textbf{Induction hypothesis}: Assume that $\hat{V}(s^\prime)$ is unbiased, i.e. $\mathbb{E}\left[\hat{V}(s^{\prime,i})\right] = V(s^{\prime,i})$, for all $s^{\prime,i} \in \mathcal{S}_{sa}$.

    From assumption (\hyperlink{assum:mis-ii}{MIS-II}), we have that $w_i(s^{\prime}) = 0$ whenever $p_T(s^{\prime} {\mid} s,a_{\textnormal{prop}}^{i}) = 0$.
    Therefore, we can ignore the subset of the integration domain in which $p_T(s^{\prime} {\mid} s,a_{\textnormal{prop}}^{i}) = 0$ when calculating the expectation of $\hat{V}_{f}(s,a)$.
    From assumption (\hyperlink{assum:mis-i}{MIS-I}), we have that $\sum_{i=1}^{\absval{\mathcal{S}_{sa}}} w_i(s^{\prime}) = 1$ whenever the target function is non-zero, i.e. whenever $p_T(s^{\prime}{\mid} s,a)(r(s,a,s^{\prime})+\gamma V_{t+1}^{\pi}(s^{\prime})) \neq 0$.
    Thus, it holds that at least one of the proposal distributions $p_T(s^{\prime} {\mid} s,a_{\textnormal{prop}}^{i})$ is non-zero at those points, and hence the estimator covers the whole domain of the target function. Thus, by using the tower property and the induction hypothesis, the expected future-value estimate is:
    \begin{linenomath*}
        \begin{align}
            &\mathbb{E}\left[\hat{V}_{f}(s,a)\right]
            \nonumber \\
            &= \mathbb{E}\left[\sum_{i=1}^{\absval{\mathcal{S}_{sa}}} w_i(s^{\prime,i}) \rho_{a}^{i}(s^{\prime,i}) \hat{V}(s^{\prime,i})\right]
            \\
            &\stackrel{(1)}{=} \mathbb{E}\left[\mathbb{E}\left[ \sum_{i=1}^{\absval{\mathcal{S}_{sa}}} w_i(s^{\prime,i}) \rho_{a}^{i}(s^{\prime,i}) \hat{V}(s^{\prime,i}) \;\bigg|\; s^{\prime,i} \right]\right]
            \\
            &\stackrel{(2)}{=} \mathbb{E}\left[ \sum_{i=1}^{\absval{\mathcal{S}_{sa}}} w_i(s^{\prime,i}) \rho_{a}^{i}(s^{\prime,i}) \mathbb{E}\left[ \hat{V}(s^{\prime,i}) \;\bigg|\; s^{\prime,i} \right]\right]
            \\
            &\stackrel{(3)}{=} \mathbb{E}\left[ \sum_{i=1}^{\absval{\mathcal{S}_{sa}}} w_i(s^{\prime,i}) \rho_{a}^{i}(s^{\prime,i}) V(s^{\prime,i}) \right],
            \label{eq:mis_w_rho_v_1}
        \end{align}
    \end{linenomath*}
    where transitions (1) and (2) are due to the tower property of expectations, and transition (3) follows from conditional independence and the induction hypothesis.

    The expectation is calculated over all random variables $s^{\prime,i}$.
    We enumerate them $s^{\prime,1},\dots,s^{\prime,n}$ and corresponding weighting functions $w_1, \dots, w_n$.
    Writing the explicit integral:
    \begin{linenomath*}
        \begin{align}
            &\eqref{eq:mis_w_rho_v_1} 
            \nonumber\\ 
            &= \int_{s^{\prime,1},\dots,s^{\prime,n}}
            p(s^{\prime,1}, \dots, s^{\prime,n} {\mid} s, a_{\textnormal{prop}}^1, \dots, a_{\textnormal{prop}}^n )
            \nonumber \\
            &\qquad \sum_{i=1}^{\absval{\mathcal{S}_{sa}}} w_i(s^{\prime,i}) \rho_{a}^{i}(s^{\prime,i}) V(s^{\prime,i})
            \dif s^{\prime,1} \cdots \dif s^{\prime,n}.
            \label{eq:mis_w_rho_v_2}
        \end{align}
    \end{linenomath*}
    We simplify $p(s^{\prime,1}, \dots, s^{\prime,n} {\mid} s, a_{\textnormal{prop}}^1, \dots, a_{\textnormal{prop}}^n )=\prod_{j=1}^{n} p_T(s^{\prime,j} {\mid} s,a_{\textnormal{prop}}^j)$, as the state samples are independent.
    Hence, distributing the product over the sum yields:
    \begin{linenomath*}
        \begin{align}
            &\eqref{eq:mis_w_rho_v_2}
            \nonumber \\
            &= \int_{s^{\prime,1},\dots,s^{\prime,n}} \sum_{i=1}^{\absval{\mathcal{S}_{sa}}} \Bigg( \bigg( \prod_{j=1}^{n} p_T(s^{\prime,j} {\mid} s,a_{\textnormal{prop}}^j \bigg) w_i(s^{\prime,i})
            \nonumber \\
            &\qquad \qquad \cdot \rho_{a}^{i}(s^{\prime,i}) V(s^{\prime,i}) \Bigg) 
            \dif s^{\prime,1} \cdots \dif s^{\prime,n}.
            \label{eq:mis_w_rho_v_3}
        \end{align}
    \end{linenomath*}
    Switching order of integral and summation (linearity of the integral), we obtain:
    \begin{linenomath*}
        \begin{align}
            &\eqref{eq:mis_w_rho_v_3}
            \nonumber \\
            &= \sum_{i=1}^{\absval{\mathcal{S}_{sa}}} \int_{s^{\prime,1},\dots,s^{\prime,n}} \bigg( \prod_{j=1}^{n} p_T(s^{\prime,j} {\mid} s,a_{\textnormal{prop}}^j) \bigg) w_i(s^{\prime,i})
            \nonumber \\
            &\qquad \qquad \cdot \rho_{a}^{i}(s^{\prime,i}) V(s^{\prime,i}) \dif s^{\prime,1} \cdots \dif s^{\prime,n}.
            \label{eq:mis_w_rho_v_4}
        \end{align}
    \end{linenomath*}
    Now, for the $i$'th summand, the integration over transitions densities $j\neq i$ cancel (they sum to 1 as proper probability densities):
    \begin{linenomath*}
        \begin{align}
            &\eqref{eq:mis_w_rho_v_4}
            \nonumber \\
            &= \sum_{i=1}^{\absval{\mathcal{S}_{sa}}} \int_{s^{\prime,i}} p_T(s^{\prime,i} \mid s,a_{\textnormal{prop}}^{i}) w_i(s^{\prime,i}) \rho_{a}^{i}(s^{\prime,i}) V(s^{\prime,i}) 
            \nonumber \\
            &\begin{multlined}[b][0.8\displaywidth] 
                \cancel{\bigg( \int_{s^{\prime,1},\dots,s^{\prime,i-1},s^{\prime,i+1},\dots,s^{\prime,n}} \prod_{j\neq i} p_T(s^{\prime,j} \mid s,a_{\textnormal{prop}}^j)}
                \\
                \cancel{\dif s^{\prime,1} \cdots \dif s^{\prime,i-1} \dif s^{\prime,i + 1} \cdots \dif s^{\prime,n} \bigg) }
            \end{multlined}
            \cdot \dif s^{\prime,i}.
            \label{eq:mis_w_rho_v_5}
        \end{align}
    \end{linenomath*}
    We substitute the definition of $\rho_{a}^{i}(s^{\prime,i})$ to obtain:
    \begin{linenomath*}
        \begin{align}
            &\eqref{eq:mis_w_rho_v_5}
            \nonumber \\
            &\begin{multlined}[b][0.8\displaywidth] =
                \sum_{i=1}^{\absval{\mathcal{S}_{sa}}} \int_{s^{\prime,i}} \cancel{p_T(s^{\prime,i} \mid s,a_{\textnormal{prop}}^{i})} w_i(s^{\prime,i})
                \\
                \cdot \frac{p_T(s^{\prime,i}\mid s,a)}{\cancel{p_T(s^{\prime,i} \mid s,a_{\textnormal{prop}}^{i})}} V(s^{\prime,i}) d s^{\prime,i}
            \end{multlined}
            \\
            &= \sum_{i=1}^{\absval{\mathcal{S}_{sa}}} \int_{s^{\prime,i}} w_i(s^{\prime,i}) p_T(s^{\prime,i}\mid s,a) V(s^{\prime,i}) \dif s^{\prime,i}.
            \label{eq:mis_w_rho_v_6}
        \end{align}
    \end{linenomath*}
    We can now rename all $s^{\prime,i}$ to $s^\prime$ as they share the same integration domain. Hence, we can switch again the order of summation and integration. By distributive laws, we combine shared terms $p_T(s^{\prime}\mid s,a) V(s^{\prime})$, and we finish the proof:
    \begin{linenomath*}
        \begin{align}
            &\eqref{eq:mis_w_rho_v_6} 
            \nonumber \\
            &= \int_{s^{\prime}} \sum_{i=1}^{\absval{\mathcal{S}_{sa}}} \left( w_i(s^{\prime}) p_T(s^{\prime}\mid s,a) V(s^{\prime}) \right) d s^{\prime}.
            \\
            &=\int_{s^{\prime}} \underbrace{\left(\sum_{i=1}^{\absval{\mathcal{S}_{sa}}} w_i(s^{\prime}) \right)}_{= 1 \text{ (\hyperlink{assum:mis-i}{MIS-I})}} p_T(s^{\prime}\mid s,a) V(s^{\prime}) d s^{\prime}
            \\
            &= \int_{s^\prime} p_T(s^{\prime}\mid s,a) V(s^{\prime}) d s^\prime 
            \\
            &= \mathbb{E} \left[ V(s^{\prime}) \right].
        \end{align}
    \end{linenomath*}
    \emph{Note regarding bias in POMDPs:} when using self-normalized particle-belief approximations instead of exact belief representations, a bias in the action-value estimate arises, which increases recursively with the tree depth.
    This bias can be bounded with finite-sample concentration inequalities.
    For an in-depth analysis of this bias we refer to \citeauthor{Lim23jair}~[\citeyear{Lim23jair}].
\end{proof}

\section{MIS Tree Procedures}
\label{sec:appendix_mis_tree_procedures}

In this section we show the derivation of the MIS Tree update equations, and also show their numerically stable forms using log likelihoods for probability values.
We denote the regular and the weighted log-sum-exp function as $LSE$, differentiated by having one or two vector inputs, and they are defined as 
\begin{linenomath*}
    \begin{gather}
        LSE(\boldsymbol{x}_{1,\ldots,n})\bydef\log\left(\sum_{i=1}^{n}\exp(\boldsymbol{x}_i)\right), \label{eq:LSE_regular} \\
        LSE(\boldsymbol{x}_{1,\ldots,n}, \boldsymbol{y}_{1,\ldots,n})\bydef\log\left(\sum_{i=1}^{n}\exp(\boldsymbol{x}_i)\cdotp \boldsymbol{y}_i\right) \label{eq:LSE_weighted}.
    \end{gather}
\end{linenomath*}
To handle negative $\boldsymbol{y}_{1,\ldots,n}$, \eqref{eq:LSE_weighted} can be defined to compute the log of absolute value and sign separately of the log argument, such that the sign is multiplied back after exponentiation if required.
We also denote a vector with brackets $[\cdot]$.

We slightly modify the notations to an enumerated form in this subsection to later match vector notations.
In these notations, the visitation counter relationships are:
\begin{linenomath*}
    \begin{gather}
        n(s)=\sum_{i=1}^{\absval{\mathcal{A}_s}}n(s,a^{i}),\quad
        n(s,a)=\sum_{i=1}^{\absval{\mathcal{S}_{sa}}}n(s^{\prime,i})_{+1}, \label{eq:visits_nodes_mcts_rpt}
    \end{gather}
\end{linenomath*}
the state value estimator:
\begin{linenomath*}
    \begin{gather}
        \hat{V}(s) \bydef \sum_{i=1}^{\absval{\mathcal{A}_s}}\frac{n(s,a^{i})}{n(s)}\tilde{Q}(s,a^{i}),
        \label{eq:state_value_estimator_rpt}
    \end{gather}
\end{linenomath*}
and the self normalized MIS estimator:
\begin{linenomath*}
    \begin{align}
        \eta(s,a) &\bydef \sum_{i=1}^{\absvalflat{\mathcal{S}_{sa}}} n(s^{\prime,i})_{+1} \rho_{a}^{i}(s^{\prime,i}),
        \label{eq:snmis_eta_rpt}
        \\
        \tilde{V}_{f}(s,a) &\bydef \sum_{i=1}^{\absvalflat{\mathcal{S}_{sa}}} n(s^{\prime,i})_{+1} \rho_{a}^{i}(s^{\prime,i}) \hat{V}(s^{\prime,i}) \slash \eta(s,a),
        \label{eq:snmis_vf_rpt}
        \\
        \tilde{r}(s,a) &\bydef \sum_{i=1}^{\absvalflat{\mathcal{S}_{sa}}} n(s^{\prime,i})_{+1} \rho_{a}^{i}(s^{\prime,i}) r(s,a,s^{\prime,i}) \slash \eta(s,a),
        \label{eq:snmis_r_rpt}
        \\
        \tilde{Q}(s,a) &\bydef \tilde{r}(s,a) + \gamma \tilde{V}_{f}(s,a).
        \label{eq:snmis_q_rpt}
    \end{align}
\end{linenomath*}
For brevity in this section, we denote $M\bydef\absvalflat{\mathcal{S}_{sa}}$.
We define the following vectors for each a given action node $(s,a)$ in the MDP case, or $(\bar{b},a)$ in the particle-belief MDP (for POMDPs) case:
\begin{enumerate}
    \item Child state visitation counters. \\
    MDP:
    \begin{linenomath*}
        \begin{gather}
            \boldsymbol{n}\bydef(n(s^{\prime,i})_{+1})_{i=1}^{M},
        \end{gather} 
    \end{linenomath*}
    particle-belief MDP:
    \begin{linenomath*}
        \begin{gather}
            \boldsymbol{n}\bydef(n(\bar{b}^{\prime,i})_{+1})_{i=1}^{M}.
        \end{gather}
    \end{linenomath*}
    \item Target transition likelihoods.\\
    MDP:
    \begin{linenomath*}
        \begin{gather}
            \boldsymbol{p}=(p_T(s^{\prime,i}{\mid} s,a))_{i=1}^{M},
        \end{gather} 
    \end{linenomath*}
    particle-belief MDP:
    \begin{linenomath*}
        \begin{gather}
            \boldsymbol{p}=(p(\bar{b}^{\prime -,i}{\mid} \bar{b},a))_{i=1}^{M}.
        \end{gather}
    \end{linenomath*}
    \item Proposal transition likelihoods.\\
    MDP:
    \begin{linenomath*}
        \begin{gather}
            \boldsymbol{q}=(p_T(s^{\prime,i}{\mid} s,a_{\textnormal{prop}}^{i}))_{i=1}^{M},
        \end{gather} 
    \end{linenomath*}
    particle-belief MDP:
    \begin{linenomath*}
        \begin{gather}
            \boldsymbol{q}=(p(\bar{b}^{\prime -,i}{\mid} \bar{b},a_{\textnormal{prop}}^{i}))_{i=1}^{M}.
        \end{gather}
    \end{linenomath*}
    \item Immediate rewards.\\
    MDP:
    \begin{linenomath*}
        \begin{gather}
            \boldsymbol{r}=(r(s,a,s^{\prime,i}))_{i=1}^{M}
        \end{gather} 
    \end{linenomath*}
    particle-belief MDP:
    \begin{linenomath*}
        \begin{gather}
            \boldsymbol{r}=(r(\bar{b},a,\bar{b}^{\prime,i}))_{i=1}^{M}.
        \end{gather}
    \end{linenomath*}
    \item Future value estimates.\\
    MDP:
    \begin{linenomath*}
        \begin{gather}
            \boldsymbol{V}=(\hat{V}(s^{\prime,i}))_{i=1}^{M}
        \end{gather} 
    \end{linenomath*}
    particle-belief MDP:
    \begin{linenomath*}
        \begin{gather}
            \boldsymbol{V}=(\hat{V}(\bar{b}^{\prime,i}))_{i=1}^{M}
        \end{gather}
    \end{linenomath*}
\end{enumerate}
From here on, the particle-belief MDP case is the same up to different notations as the MDP case, and we will show only the latter.
Using these notations the immediate reward and future value estimates can be compactly written:
\begin{linenomath*}
    \begin{align}
        \eta(s,a)&=\boldsymbol{1}^{T}(\boldsymbol{n}\odot\boldsymbol{p}\oslash\boldsymbol{q}), \\
        \tilde{r}(s,a)&=\eta(s,a)^{-1}\boldsymbol{1}^{T}(\boldsymbol{n}\odot\boldsymbol{p}\odot\boldsymbol{r}\oslash\boldsymbol{q}), \\
        \tilde{V}_{f}(s,a)&=\eta(s,a)^{-1}\boldsymbol{1}^{T}(\boldsymbol{n}\odot\boldsymbol{p}\odot\boldsymbol{V}\oslash\boldsymbol{q}),
    \end{align}
\end{linenomath*}
where $\odot$ and $\oslash$ are the Hadamard (element-wise) product and division operations respectively.
Similarly, we denote Hadamard addition and subtraction as $\oplus$ and $\ominus$.
We use a shortened notation $l(\cdot)\bydef \log$, which can be kept in any base.

We store the following values:
\begin{enumerate}
    \item $l(\eta)$.
    \item Log likelihoods $l(\boldsymbol{p})$ and $l(\boldsymbol{q})$.
    \item $\boldsymbol{n}^\prime$ and previous visitation counters $\boldsymbol{n}$.
    \item $\boldsymbol{V}^\prime$ and previous future value estimates $\boldsymbol{V}$.
\end{enumerate}

The equations for $\tilde{r}(s,a)$ are analogous to $\tilde{V}_{f}(s,a)$, and we will not show their derivation explicitly.

\paragraph{Action Backpropagation.}
Let $s^{\prime,j}$ be an updated node. Hence, we updated $\boldsymbol{n}_j$ to $\boldsymbol{n}^{\prime}_j$ and $\boldsymbol{V}_j$ to $\boldsymbol{V}^{\prime}_j$. We update $n(s,a)$ by \eqref{eq:visits_nodes_mcts_rpt} and perform
\begin{linenomath*}
    \begin{align}
        \eta^{\prime}(s,a)&=\eta(s,a)+\frac{\boldsymbol{p}_{j}}{\boldsymbol{q}_{j}}\left(\boldsymbol{n}_{j}^{\prime}-\boldsymbol{n}_{j}\right),
        \label{eq:action_backprop_update_eta}
        \\
        \tilde{V}_{f}^{\prime}(s,a)&=\frac{\left(\eta(s,a)\tilde{V}_{f}(s,a)+\frac{\boldsymbol{p}_{j}}{\boldsymbol{q}_{j}}(\boldsymbol{n}_{j}^{\prime}\boldsymbol{V}_{j}^{\prime}-\boldsymbol{n}_{j}\boldsymbol{V}_{j})\right)}{\eta^{\prime}(s,a)} .
        \label{eq:action_backprop_update_v}
    \end{align}
\end{linenomath*}
Therefore, the log-form update equations are:
\begin{linenomath*}
    \begin{align}
        l(\eta^{\prime}(s,a))&=LSE\Big([l(\eta(s,a)),
        \nonumber \\
        &\qquad l(\boldsymbol{p}_{j})-l(\boldsymbol{q}_{j})+l(\boldsymbol{n}_{j}^{\prime}-\boldsymbol{n}_{j})]\Big)
        \label{eq:action_backprop_update_eta_log}
        \\
        \tilde{V}_{f}^{\prime}(s,a)&=\exp\Big(LSE\big([l(\eta(s,a)),l(\boldsymbol{p}_{j})-l(\boldsymbol{q}_{j})],
        \nonumber \\
        &\qquad [\tilde{V}_{f}(s,a),\boldsymbol{n}_{j}^{\prime}\boldsymbol{V}_{j}^{\prime}-\boldsymbol{n}_{j}\boldsymbol{V}_{j}]\big)-l(\eta^{\prime}(s,a))\Big).
        \label{eq:action_backprop_update_v_log}
    \end{align}
\end{linenomath*}

\begin{proof}
    We start with the update for $\eta^{\prime}(s,a)$:
    \begin{linenomath*}
        \begin{align}
            &\eta^{\prime}(s,a)-\eta(s,a) 
            \nonumber \\
            &=\sum_{i=1}^{M}\frac{\boldsymbol{n}_{i}^{\prime}\boldsymbol{p}_{i}}{\boldsymbol{q}_{i}}-\sum_{i=1}^{M}\frac{\boldsymbol{n}_{i}\boldsymbol{p}_{i}}{\boldsymbol{q}_{i}}
            \\
            &=\frac{\boldsymbol{p}_{j}}{\boldsymbol{q}_{j}}(\boldsymbol{n}_{j}^{\prime}-\boldsymbol{n}_{j}),
            \label{eq:action_backprop_update_eta_proof}
        \end{align}
    \end{linenomath*}
    due to all terms being equal except for $\boldsymbol{n}_{j}^{\prime}\neq\boldsymbol{n}_{j}$.
    Similarly, for $\tilde{V}_{f}(s,a)$:
    \begin{linenomath*}
        \begin{align}
            &\tilde{V}_{f}^{\prime}(s,a)-\frac{\eta(s,a)}{\eta^{\prime}(s,a)}\tilde{V}_{f}(s,a)
            \nonumber \\
            &=\frac{1}{\eta^{\prime}(s,a)}\sum_{i=1}^{M}\frac{\boldsymbol{n}_{i}^{\prime}\boldsymbol{p}_{i}\boldsymbol{V}_{i}^{\prime}}{\boldsymbol{q}_{i}}-\frac{\eta(s,a)}{\eta^{\prime}(s,a)}\frac{1}{\eta(s,a)}\sum_{i=1}^{M}\frac{\boldsymbol{n}_{i}\boldsymbol{p}_{i}\boldsymbol{V}_{i}}{\boldsymbol{q}_{i}}
            \\
            &=\frac{1}{\eta^{\prime}(s,a)}\left(\sum_{i=1}^{M}\frac{\boldsymbol{n}_{i}^{\prime}\boldsymbol{p}_{i}\boldsymbol{V}_{i}^{\prime}}{\boldsymbol{q}_{i}}-\sum_{i=1}^{M}\frac{\boldsymbol{n}_{i}\boldsymbol{p}_{i}\boldsymbol{V}_{i}}{\boldsymbol{q}_{i}}\right)
            \\
            &=\frac{1}{\eta^{\prime}(s,a)}\frac{\boldsymbol{p}_{j}}{\boldsymbol{q}_{j}}(\boldsymbol{n}_{j}^{\prime}\boldsymbol{V}_{j}^{\prime}-\boldsymbol{n}_{j}\boldsymbol{V}_{j}),
            \label{eq:action_backprop_update_v_proof}
        \end{align}
    \end{linenomath*}
    due to all terms being equal except for $\boldsymbol{V}^{\prime}_j\neq\boldsymbol{V}_j$, $\boldsymbol{n}^{\prime}_j\neq\boldsymbol{n}_j$.
\end{proof}

\paragraph{State Expansion.}
This case is the same as \textit{action backpropagation}, where the updated index is a new summand at index $j=M+1$.
The same equations apply with $n(s^{\prime,j})=0$.

\begin{proof}
    We append new terms to all vectors: $\boldsymbol{n}_{M+1}$, $\boldsymbol{p}_{M+1}$, $\boldsymbol{V}_{M+1}$, $\boldsymbol{q}_{M+1}$.
    Update to $\eta(s,a)$:
    \begin{linenomath*}
        \begin{align}
            &\eta^{\prime}(s,a)-\eta(s,a)
            \nonumber \\
            &=\sum_{i=1}^{M+1}\frac{\boldsymbol{n}_{i}^{\prime}\boldsymbol{p}^{\prime}_{i}}{\boldsymbol{q}^{\prime}_{i}}-\sum_{i=1}^{M}\frac{\boldsymbol{n}_{i}\boldsymbol{p}_{i}}{\boldsymbol{q}_{i}}
            \\
            &=\frac{\boldsymbol{n}_{M+1}^{\prime}\boldsymbol{p}^{\prime}_{M+1}}{\boldsymbol{q}^{\prime}_{M+1}},
        \end{align}
    \end{linenomath*}
    due to all terms being equal except for index $M+1$.
    This corresponds to \eqref{eq:action_backprop_update_eta_proof} with $\boldsymbol{n}_{M+1}=0$, after initializing $\boldsymbol{p}^{\prime}_{M+1}=\boldsymbol{p}_{M+1}$ and $\boldsymbol{q}^{\prime}_{M+1}=\boldsymbol{q}_{M+1}$
    Similarly, for $\hat{V}_{f}(s,a)$:
    \begin{linenomath*}
        \begin{align}
            &\tilde{V}_{f}^{\prime}(s,a)-\frac{\eta(s,a)}{\eta^{\prime}(s,a)}\tilde{V}_{f}(s,a)
            \nonumber \\
            &=\frac{1}{\eta^{\prime}(s,a)}\sum_{i=1}^{M+1}\frac{\boldsymbol{n}_{i}^{\prime}\boldsymbol{p}_{i}\boldsymbol{V}_{i}^{\prime}}{\boldsymbol{q}_{i}}-\frac{\eta(s,a)}{\eta^{\prime}(s,a)}\frac{1}{\eta(s,a)}\sum_{i=1}^{M}\frac{\boldsymbol{n}_{i}\boldsymbol{p}_{i}\boldsymbol{V}_{i}}{\boldsymbol{q}_{i}}
            \\
            &=\frac{1}{\eta^{\prime}(s,a)}\left(\sum_{i=1}^{M+1}\frac{\boldsymbol{n}_{i}^{\prime}\boldsymbol{p}_{i}\boldsymbol{V}_{i}^{\prime}}{\boldsymbol{q}_{i}}-\sum_{i=1}^{M}\frac{\boldsymbol{n}_{i}\boldsymbol{p}_{i}\boldsymbol{V}_{i}}{\boldsymbol{q}_{i}}\right)
            \\
            &=\frac{1}{\eta^{\prime}(s,a)}\frac{\boldsymbol{p}^{\prime}_{M+1}}{\boldsymbol{q}^{\prime}_{M+1}}\boldsymbol{n}_{M+1}^{\prime}\boldsymbol{V}_{M+1}^{\prime},
        \end{align}
    \end{linenomath*}
    which corresponds to the expression in \eqref{eq:action_backprop_update_v_proof} with $\boldsymbol{n}_{M+1}=0$.
\end{proof}

\paragraph{Action Update.}
Let $\breve{a}$ be the new action. Here, we have to calculate $\rho_{\breve{a}}^{i}(s^{\prime})$ for all $s^{\prime,i} \in \mathcal{S}_{sa}$ in $O(M)$ time. We update
\begin{linenomath*}
    \begin{align}
        \rho_{\breve{a}}^{i}(s^{\prime}) &= p_T(s^{\prime}{\mid} s,\breve{a}) \slash p_T(s^{\prime}{\mid} s,a_{\textnormal{prop}}^{i}),
        \label{eq:action_update_rpt}
    \end{align}
\end{linenomath*}
and recalculate \eqref{eq:snmis_vf_rpt} and \eqref{eq:snmis_eta_rpt} with the new $\rho_{\breve{a}}^{i}(s^{\prime})$, which are also $O(M)$ time operations.
For the immediate reward, new rewards must be computed for the new triplets $(s,\breve{a},s^{\prime})$, and the update is analogous.

Therefore, the log-form equations are:
\begin{linenomath*}
    \begin{align}
        l(\eta^{\prime}(s,\breve{a}))&=LSE(l(\boldsymbol{n})\oplus l(\boldsymbol{p}^{\prime})\ominus l(\boldsymbol{q}))
        \\
        \tilde{V}_{f}^{\prime}(s,\breve{a})&=\exp\left(LSE(l(\boldsymbol{p}^{\prime})\ominus l(\boldsymbol{q}),\boldsymbol{n}\odot\boldsymbol{V})-l(\eta^{\prime}(s,\breve{a}))\right)
    \end{align}
\end{linenomath*}

In case we linearize the weight updates (as in Equation \eqref{eq:imp_ratio_linear_update}), instead of re-evaluating the density, we update the stored log-likelihoods $l(\boldsymbol{p})$ using the gradients.
Let $\delta a = \breve{a} - a$. 
The updated log-target likelihoods are approximated by:
\begin{linenomath*}
    \begin{align}
        l(\boldsymbol{p}^{\prime}_{i}) \approx l(\boldsymbol{p}_{i}) + \nabla_a \log p_T(s^{\prime,i}{\mid} s,a)^T \delta a.
    \end{align}
\end{linenomath*}
This approximation avoids the expensive re-evaluation of $p_T(s^{\prime}{\mid} s,a)$ (or $p(\bar{b}^{\prime -} {\mid} \bar{b},a)$ in POMDPs).
The LSE aggregation steps for $\eta^{\prime}$ and $\tilde{V}_{f}^{\prime}$ remain identical using the approximated $l(\boldsymbol{p}^{\prime})$.

\begin{proof}
    The update of $\boldsymbol{p}^{\prime}$ corresponds to recomputing it either via exact re-evaluation or via the linearized approximation.
    The linearized approximation is derived from a first-order Taylor expansion of $l(\boldsymbol{p}_{i})$ around $a$:
    \begin{linenomath*}
        \begin{align}
            l(\boldsymbol{p}^{\prime}_{i}) &= \log (p_T(s^{\prime,i}{\mid} s,\breve{a}))
            \\
            &\approx \log (p_T(s^{\prime,i}{\mid} s,a)) + \nabla_a \log p_T(s^{\prime,i}{\mid} s,a)^T (\breve{a} - a)
            \\
            &= l(\boldsymbol{p}_{i}) + \nabla_a \log p_T(s^{\prime,i}{\mid} s,a)^T \delta a.
        \end{align}
    \end{linenomath*}
    Therefore, the log-form update equations that follow match exactly the definitions:
    \begin{linenomath*}
        \begin{align}
            \eta^{\prime}(s,\breve{a})&=\sum_{i=1}^{M}\frac{\boldsymbol{n}_{i}\boldsymbol{p}_{i}^{\prime}}{\boldsymbol{q}_{i}}, \\
            \tilde{V}_{f}^{\prime}(s,\breve{a})&=\frac{1}{\eta^{\prime}(s,\breve{a})}\sum_{i=1}^{M}\frac{\boldsymbol{n}_{i}\boldsymbol{p}_{i}^{\prime}\boldsymbol{V}_{i}}{\boldsymbol{q}_{i}}.
        \end{align}
    \end{linenomath*}
\end{proof}

\begin{algorithm}[tb]
    \caption{AGMCTS - Full Pseudocode. Highlighted lines mark AGMCTS-specific steps.}
    \label{alg:AGMCTS_full}
    \textbf{procedure}\;\textsc{Simulate}$(s,d)$
    \begin{algorithmic}[1]
    \IF {\textsc{IsTerminal}$(s)$ OR $d=0$}
    \STATE $v\leftarrow$ \textsc{Rollout}$(s)$
    \COMMENT{\textit{Rollout until terminal state}}
    \STATE \begingroup \AGHighlightColor \textsc{UpdateTerminal}($s,v$)
    \COMMENT {\textit{Eq. \eqref{eq:posterior_terminal_update}}} \endgroup
    \STATE \textbf{return} $v$
    \ENDIF
    \STATE $a\leftarrow$ \textsc{ActionProgWiden}$(s)$
    \COMMENT{\textit{Vanilla or VPW}}
    \STATE \begingroup \AGHighlightColor $addBranch \leftarrow $ \textsc{ActionOpt} $(s,a,d)$ \endgroup
    \IF {\begingroup \AGHighlightColor $\absvalflat{\mathcal{S}_{sa}}\leq k_{o}\cdot n(s,a)^{\alpha_{o}}$ OR $addBranch$ \endgroup}
    \STATE $s^{\prime},r \sim G(s,a)$
    \begingroup \AGHighlightColor
    \COMMENT {$\bar{b}^{\prime -},\bar{b}^{\prime},r \sim G(\bar{b},a)$ \textit{in POMDPs}}
    \endgroup
    \STATE $\mathcal{S}_{sa}\leftarrow \mathcal{S}_{sa}\cup\left\{ s^{\prime}\right\}$ 
    \STATE $v\leftarrow\text{\textsc{Rollout}}(s^{\prime},d-1)$
    \ELSE
    \STATE $s^{\prime} \sim \textnormal{Unif}(\mathcal{S}_{sa})$, $r\leftarrow$ \textsc{Reward}$(s,a,s^{\prime})$
    \STATE $v\leftarrow\text{\textsc{Simulate}}(s^{\prime},d-1)$
    \ENDIF
    \STATE \begingroup \AGHighlightColor \textsc{UpdateMIS}($s,a,s^\prime,r$)
    \COMMENT {\textit{Eqs. \eqref{eq:action_backprop_update_1}, \eqref{eq:action_backprop_update_2}, \eqref{eq:posterior_nonterminal_update}}} \endgroup
    \end{algorithmic}
    \AGHighlightColor
    \textbf{procedure}\;$\text{\textsc{ActionOpt}}(s,a,d)$
    \begin{algorithmic}[1]
    \STATE $addBranch\leftarrow$ FALSE
    \IF {\textsc{ActionUpdateRule}$(s,a,d)$}
    \FORALL {$k=1,\dots,K_{\text{opt}}$} 
    \STATE $g_a^q \leftarrow \hat{\nabla}_{a}\tilde{Q}(s,a)$
    \COMMENT {\textit{Either of Eqs. \eqref{eq:mdp_grad_est_linearized_rpt}}-\eqref{eq:pomdp_state_grad_est}}
    \STATE $\breve{a} \leftarrow $\textsc{ClipNorm}$(\breve{a},T_{d_a}^{\max})$
    \STATE \textsc{ActionUpdate}$(s, a, \breve{a})$
    \COMMENT {\textit{Eq. \eqref{eq:branch_imp_ratio}-\eqref{eq:imp_ratio_linear_update}, \eqref{eq:posterior_nonterminal_update}}}
    \STATE $addBranch \leftarrow addBranch \lor (\forall s^{\prime,i}\in \mathcal{S}_{s\breve{a}}: \rho_{\breve{a}}^{i}(s^{\prime,i})\leq T_{\rho}^{add})$
    \ENDFOR
    \ENDIF
    \STATE \textbf{return} $addBranch$
    \end{algorithmic}
\end{algorithm}

\paragraph{Terminal State Backpropagation.}
Let $s$ be a terminal node.
Its value estimate is based only on rollouts.
Hence, for a new rollout value $v^{\prime}$, we perform a running average update
\begin{linenomath*}
    \begin{align}
        \hat{V}^{\prime}(s)=\hat{V}(s)+\frac{n^{\prime}(s)-n(s)}{n^{\prime}(s)_{+1}}(v^{\prime}-\hat{V}(s)),
        \label{eq:posterior_terminal_update_rpt}
    \end{align}
\end{linenomath*}
where $n^{\prime}(s)$ is $n(s)$ plus the number of rollouts (usually 1).

\begin{proof}
    This is a simple running average update for the mean estimator:
    \begin{linenomath*}
        \begin{align}
            \hat{V}(s)=\frac{1}{n(s)_{+1}}\sum_{i=1}^{n(s)_{+1}}v^{i},
        \end{align}
    \end{linenomath*}
    for all $v^{i}$ rollout values from the terminal node $s$.
\end{proof}

\paragraph{Non-Terminal State Backpropagation.}
Let $(s,a^j)$ be an updated node. Hence, we updated $n(s,a^j)$ to $n^{\prime}(s,a^j)$ and $\tilde{Q}(s,a^j)$ to $\tilde{Q}^{\prime}(s,a^j)$. We update $n^{\prime}(s)$ by \eqref{eq:visits_nodes_mcts_rpt} and perform
\begin{linenomath*}
    \begin{align}
            \hat{V}^{\prime}(s)=\frac{n(s)\hat{V}(s)+n^{\prime}(s,a^j)\tilde{Q}^{\prime}(s,a^j) 
            - n(s,a^j)\tilde{Q}(s,a^j)}{n^{\prime}(s)}.
        \label{eq:posterior_nonterminal_update_rpt}
    \end{align}
\end{linenomath*}

\begin{proof}
    Following the definition of the state value estimator \eqref{eq:state_value_estimator_rpt}, the updated estimator is:
    \begin{linenomath*}
        \begin{align}
            \hat{V}^{\prime}(s) = \sum_{i=1}^{\absval{\mathcal{A}_s}}\frac{n^{\prime}(s,a^{i})}{n^{\prime}(s)}\tilde{Q}^{\prime}(s,a^{i}).
        \end{align}
    \end{linenomath*}
    Therefore:
    \begin{linenomath*}
        \begin{align}
            & \hat{V}^{\prime}(s) - \frac{n(s)}{n^{\prime}(s)}\hat{V}(s)
            \nonumber \\
            &= \sum_{i=1}^{\absval{\mathcal{A}_s}}\frac{n^{\prime}(s,a^{i})}{n^{\prime}(s)}\tilde{Q}^{\prime}(s,a^{i}) - \frac{n(s)}{n^{\prime}(s)}\sum_{i=1}^{\absval{\mathcal{A}_s}}\frac{n(s,a^{i})}{n(s)}\tilde{Q}(s,a^{i})
            \\
            &= \frac{1}{n^{\prime}(s)}\sum_{i=1}^{\absval{\mathcal{A}_s}}(n^{\prime}(s,a^{i})\tilde{Q}^{\prime}(s,a^{i}) - n(s,a^{i})\tilde{Q}(s,a^{i})).
        \end{align}
    \end{linenomath*}
    All terms are equal except for the updated node $j$.
    Hence, we obtain
    \begin{linenomath*}
        \begin{align}
            & \hat{V}^{\prime}(s) - \frac{n(s)}{n^{\prime}(s)}\hat{V}(s)
            \nonumber \\
            &= \frac{1}{n^{\prime}(s)} \left(n^{\prime}(s,a^{j})\tilde{Q}^{\prime}(s,a^{j}) - n(s,a^{j})\tilde{Q}(s,a^{j})\right).
        \end{align}
    \end{linenomath*}
\end{proof}

\section{Area Formula Discussion}
\label{sec:appendix_area_formula_assumptions}

\begin{manualtheorem}{4.1}~\cite{Negro22}
If $f$ is locally Lipschitz, and it holds that $\operatorname{rank}(D_\xi f)=n_\xi$ a.e., then the induced probability measure over $s^\prime$ is absolutely continuous w.r.t. the Hausdorff measure $\mathcal{H}^{n_\xi}$ on $\mathbb{R}^{n_s}$, and its Radon-Nikodym derivative is
\begin{linenomath*}
    \begin{align}
        \textstyle p_T(s^{\prime}{\mid} s,a)=\sum_{\Xi^{*}}p_{\xi}(\xi^{*}{\mid} s,a)(J_{n_{\xi}}f(s,a,\xi^{*}))^{-1},
        \label{eq:area_formula_rpt}
    \end{align}
\end{linenomath*}
for $\Xi^{*}=\{\xi^{*}:s^{\prime}=f(s,a,\xi^{*})\}$, and $0$ when $\Xi^{*}=\emptyset$.
The $n_{\xi}$-dimensional Jacobian is given by the Cauchy-Binet formula: $J_{n_{\xi}}f(s,a,\xi)=\sqrt{\det((D_\xi f)^{\intercal} \cdot (D_\xi f))}$.
\end{manualtheorem}

We provide here a more detailed discussion of the assumptions in the Theorem and their implications for the transition density of the simulator.

\textbf{$f$ is locally Lipschitz.}
For every $\xi_0 \in \mathbb{R}^{n_\xi}$ there exist a neighborhood $U_{\xi_0}$ and a constant $L_{\xi_0}<\infty$ such that
\begin{linenomath*}
    \begin{align}
        \normflat{f(s,a,\xi_1)-f(s,a,\xi_2)}
        \leq
        L_{\xi_0}\normflat{\xi_1-\xi_2},
        \,\,
        \forall \xi_1,\xi_2 \in U_{\xi_0}.
    \end{align}
\end{linenomath*}
This is only a local regularity assumption in the noise variable, and is weaker than requiring a single global Lipschitz constant for all $\xi$. Example: the map $f(s,a,\xi) = s + \xi^3$ for $s,a,\xi\in\mathbb{R}$ is locally Lipschitz in $\xi$ but not globally Lipschitz.

\textbf{Full rank condition: $\operatorname{rank}(D_\xi f)=n_\xi$ a.e.}
For fixed $(s,a)$, the condition $\operatorname{rank}(D_\xi f)=n_\xi$ a.e. means that the noise coordinates used by the simulator generate $n_\xi$ independent infinitesimal directions in the successor state, except on a null set. Thus the assumption is not a restriction that the ambient transition be full-dimensional in $\mathbb{R}^{n_s}$, and it is compatible with $n_\xi<n_s$; it rules out redundant or inactive noise coordinates in the chosen generative representation. This makes the condition a property of the simulator parameterization rather than of the intrinsic MDP transition law. Indeed, let $\mathcal{M}_{s,a}=\operatorname{supp}(p_T(\cdot{\mid}s,a))$ be the supported transition manifold, with intrinsic dimension $m$, and consider the identity generative model that samples $s^\prime\sim p_T(\cdot{\mid}s,a)$ on $\mathcal{M}_{s,a}$ and returns $f^{\mathrm{id}}_{s,a}(s^\prime)=s^\prime$. In any smooth local chart of $\mathcal{M}_{s,a}$, this model has rank $m$ by construction, so a full-rank description exists on the supported manifold. The practical issue is whether the simulator exposes such non-redundant coordinates, together with the corresponding density w.r.t. $\mathcal{H}^{m}$, in a computable analytic form.

\section{AGMCTS Algorithm Details}
\label{sec:appendix_agmcts}

\begin{algorithm}[h]
    \caption{Action Progressive Widening (APW)}
    \label{alg:vanilla_pw}
    \textbf{procedure}\;\textsc{ActionProgWiden}$(s)$
    \vspace{-\baselineskip}
    \begin{algorithmic}[1]
    \IF {$\absval{\mathcal{A}_{s}} \le k_{a} \cdot n(s)^{\alpha_{a}}$}
        \STATE $a \sim \textnormal{Unif}(\mathcal{A})$
        \STATE $\mathcal{A}_{s} \leftarrow \mathcal{A}_{s} \cup \{a\}$
        \STATE Initialize $Q(s,a)$ and $n(s,a)$
        \RETURN $a$
    \ELSE
        \STATE \COMMENT{Select best action using UCB}
        \RETURN $\argmax_{a \in \mathcal{A}_{s}} Q(s,a) + c \sqrt{\frac{\log n(s)}{n(s,a)}}$
    \ENDIF
    \end{algorithmic}
\end{algorithm}

\begin{algorithm}[h]
    \caption{Voronoi Progressive Widening (VPW)}
    \label{alg:voronoi_pw}
    \textbf{procedure}\;\textsc{VoronoiProgWiden}$(s)$
    \vspace{-\baselineskip}
    \begin{algorithmic}[1]
    \IF {$\absval{\mathcal{A}_{s}} \le k_{a} \cdot n(s)^{\alpha_{a}}$}
        \STATE $a \leftarrow \textsc{VOO}(\mathcal{A}_{s}, \{Q(s,a')\}_{a' \in \mathcal{A}_{s}})$ 
        \STATE $\mathcal{A}_{s} \leftarrow \mathcal{A}_{s} \cup \{a\}$
        \STATE Initialize $Q(s,a)$ and $n(s,a)$
        \RETURN $a$
    \ELSE
        \STATE \COMMENT{Select best action using UCB}
        \RETURN $\argmax_{a \in \mathcal{A}_{s}} Q(s,a) + c \sqrt{\frac{\log n(s)}{n(s,a)}}$
    \ENDIF
    \end{algorithmic}
    \textbf{procedure}\;\textsc{VOO}$(\mathcal{A}_{s}, Q_{vals})$
    \begin{algorithmic}[1]
    \STATE $u \sim \text{Unif}[0,1]$
    \IF {$u \le \omega_{\textnormal{VOO}}$ \textbf{or} $\absval{\mathcal{A}_{s}} = 0$}
        \RETURN $a \sim \text{Unif}(\mathcal{A})$
        \COMMENT{Global Exploration}
    \ELSE
        \RETURN \textsc{BestVoronoiCell}$(\mathcal{A}_{s}, Q_{vals})$
        \COMMENT{Local Refinement}
    \ENDIF
    \end{algorithmic}
    \vspace{0.5\baselineskip}
    \textbf{procedure}\;\textsc{BestVoronoiCell}$(\mathcal{A}_{s}, Q_{vals})$
    \vspace{-\baselineskip}
    \begin{algorithmic}[1]
    \STATE $a^* \leftarrow \argmax_{a' \in \mathcal{A}_{s}} Q_{vals}(a')$
    \WHILE{true}
        \STATE $a \sim \mathcal{N}(a^*, \Sigma_{\textnormal{VOO}})$ \COMMENT{Gaussian Rejection Sampling}
        \IF {$\forall a_i \in \mathcal{A}_{s} \setminus \{a^*\}, D(a, a^*) \le D(a, a_i)$}
            \RETURN $a$
        \ENDIF
    \ENDWHILE
    \end{algorithmic}
\end{algorithm}

\begin{table*}[h]
    \centering
    \caption{
        Hyperparameters optimized by CE in the evaluations of each algorithm in the D-Continuous Light-Dark, Mountain Car, Hill Car and Lunar Lander domains. The different subsets of algorithms to which each parameter applies are indicated in the "Algorithms" column. The "Tuning" column indicates whether the parameter was optimized via Cross-Entropy (CE) or set manually. Algorithm categories: AGMCTS includes AG-DPW, AG-VPW, AG-PFT-DPW, AG-PFT-VPW; PFT includes PFT-DPW, PFT-VPW, AG-PFT-DPW, AG-PFT-VPW; VPW includes VPW, AG-VPW, PFT-VPW, AG-PFT-VPW, VOMCPOW.
        }
    \label{tab:hyperparameters_notations}
    \begin{tabular}{cccc}
        \toprule
        Param. Notation & Explanation & Algorithms & Tuning \\
        \midrule
        $c$ & UCT exploration bonus & All & CE \\
        $k_a$ & Action prog. widening factor & All & CE \\
        $\alpha_a$ & Action prog. widening power & All & CE \\
        $k_o$ & Obs. prog. widening factor & All & CE \\
        $\alpha_o$ & Obs. prog. widening power & All & CE \\
        $\eta_{\text{Adam}}$ & Adam step size (learning rate) & AGMCTS & CE \\
        $J$ & No. particles & PFT & Manual \\
        $K_{rollout}$ & No. states for rollout & PFT & Manual \\
        $\omega_{\textnormal{VOO}}$ & VOO global exploration prob. & VPW & Manual \\
        $\Sigma_{\textnormal{VOO}}$ & VOO local refinement cov. & VPW & Manual \\
        $K_{\text{opt}}$ & No. grad. iterations & AGMCTS & Manual \\
        $T_{d_a}^{\max}$ & Max. \textit{action update} step & AGMCTS & Manual \\
        $T_{\rho}^{add}$ & Imp. ratio add threshold & AGMCTS & Manual \\
        $T_{\rho}^{del}$ & Imp. ratio delete threshold & AGMCTS & Manual \\
        $K_{b}^{\nabla}$ & No. states for $\hat{\nabla}_a \log p_T$ & AGMCTS & Manual \\
        $n_{\max}^{\textnormal{sims}}$ & Max. simulation budget & All & Manual \\
        \bottomrule
    \end{tabular}
\end{table*}

\begin{table*}[h]
    \caption{Hyperparameters used in the evaluations of each algorithm in the 2D-Continuous Light-Dark domain.}
    \label{tab:2d_light_dark_hyperparameters_reformatted}
    \centering
    \small
    \begin{tabular}{l cc cc cc}
        \toprule
        & \multicolumn{2}{c}{\textbf{DPW Variants}} & \multicolumn{2}{c}{\textbf{AG Variants (Ours)}} & \multicolumn{2}{c}{\textbf{POMCP Variants}} \\
        \textbf{Param.} & PFT-DPW & PFT-VPW & AG-PFT-DPW & AG-PFT-VPW & POMCPOW & VOMCPOW \\
        \midrule
        $c$             & 1.01 & 1.02 & 1.96 & 3.16 & 0.86 & 1.45 \\
        $k_a$           & 7.68 & 7.16 & 5.70 & 4.29 & 0.46 & 0.36 \\
        $\alpha_a$      & 0.52 & 0.48 & 0.47 & 0.48 & 0.77 & 0.75 \\
        $k_o$           & 8.90 & 9.03 & 8.05 & 4.51 & 0.16 & \num{9.6e-2} \\
        $\alpha_o$      & 0.30 & 0.35 & 0.77 & 0.67 & 0.25 & 0.30 \\
        \midrule
        $\eta_{\text{Adam}}$ & - & - & \num{5.3e-8} & \num{4.28e-7} & - & - \\
        \bottomrule
    \end{tabular}    
\end{table*}

\begin{table*}[h]
    \caption{Hyperparameters used in the evaluations of each algorithm in the 3D-Continuous Light-Dark domain.}
    \label{tab:3d_light_dark_hyperparameters}
    \centering
    \small
    \begin{tabular}{l cc cc cc}
        \toprule
        & \multicolumn{2}{c}{\textbf{DPW Variants}} & \multicolumn{2}{c}{\textbf{AG Variants (Ours)}} & \multicolumn{2}{c}{\textbf{POMCP Variants}} \\
        \textbf{Param.} & PFT-DPW & PFT-VPW & AG-PFT-DPW & AG-PFT-VPW & POMCPOW & VOMCPOW \\
        \midrule
        $c$             & 1.70 & 1.28 & 2.52 & 1.60 & 1.04 & 1.28 \\
        $k_a$           & 11.76 & 7.29 & 5.71 & 4.57 & 0.85 & 0.45 \\
        $\alpha_a$      & 0.26 & 0.54 & 0.44 & 0.62 & 0.49 & 0.77 \\
        $k_o$           & 7.49 & 8.58 & 6.11 & 8.93 & 0.40 & 0.16 \\
        $\alpha_o$      & 0.33 & 0.63 & 0.63 & 0.22 & 0.33 & 0.28 \\
        \midrule
        $\eta_{\text{Adam}}$ & - & - & \num{3.8e-8} & \num{2.9e-7} & - & - \\
        \bottomrule
    \end{tabular}    
\end{table*}

\begin{table*}[h]
    \caption{Hyperparameters used in the evaluations of each algorithm in the 4D-Continuous Light-Dark domain.}
    \label{tab:4d_light_dark_hyperparameters}
    \centering
    \small
    \begin{tabular}{l cc cc cc}
        \toprule
        & \multicolumn{2}{c}{\textbf{DPW Variants}} & \multicolumn{2}{c}{\textbf{AG Variants (Ours)}} & \multicolumn{2}{c}{\textbf{POMCP Variants}} \\
        \textbf{Param.} & PFT-DPW & PFT-VPW & AG-PFT-DPW & AG-PFT-VPW & POMCPOW & VOMCPOW \\
        \midrule
        $c$             & 1.45 & 1.12 & 3.59 & 0.70 & 1.18 & 1.55 \\
        $k_a$           & 10.85 & 8.39 & 7.82 & 5.12 & 0.79 & 0.31 \\
        $\alpha_a$      & 0.29 & 0.46 & 0.22 & 0.62 & 0.47 & 0.82 \\
        $k_o$           & 7.00 & 6.74 & 0.96 & 6.21 & 0.27 & 0.14 \\
        $\alpha_o$      & 0.14 & 0.62 & 0.24 & 0.33 & 0.41 & 0.14 \\
        \midrule
        $\eta_{\text{Adam}}$ & - & - & \num{4.2e-8} & \num{2.8e-8} & - & - \\
        \bottomrule
    \end{tabular}
\end{table*}

\begin{table*}[h]
    \caption{Hyperparameters used in the evaluations of each algorithm in the Two-Agent 2D-Continuous Light-Dark domain.}
    \label{tab:collab_light_dark_hyperparameters}
    \centering
    \small
    \begin{tabular}{l cc cc cc}
        \toprule
        & \multicolumn{2}{c}{\textbf{DPW Variants}} & \multicolumn{2}{c}{\textbf{AG Variants (Ours)}} & \multicolumn{2}{c}{\textbf{POMCP Variants}} \\
        \textbf{Param.} & PFT-DPW & PFT-VPW & AG-PFT-DPW & AG-PFT-VPW & POMCPOW & VOMCPOW \\
        \midrule
        $c$             & 3.72 & 0.93 & 9.33 & \num{6.7e-3} & 4.12 & 1.76 \\
        $k_a$           & 9.64 & 9.24 & 0.14 & \num{7.5e-4} & 4.06 & 4.80 \\
        $\alpha_a$      & 0.08 & 0.49 & 0.26 & 1.00 & 0.30 & 0.52 \\
        $k_o$           & 2.89 & 8.14 & 10.81 & 18.99 & 3.97 & 1.70 \\
        $\alpha_o$      & 0.40 & 0.43 & 0.50 & 1.00 & 0.37 & 0.52 \\
        \midrule
        $\eta_{\text{Adam}}$ & - & - & \num{1.2e-2} & 0.90 & - & - \\
        \bottomrule
    \end{tabular}
\end{table*}

\begin{table*}[h]
    \caption{
        Performance (Mean $\pm$ SEM) and Runtimes in 2D-Continuous Light-Dark.
    }
    \label{tab:2d_light_dark_complete}
    \centering
    \small
    \begin{tabular}{l cc cc cc}
        \toprule
        \textbf{Simulations} & \multicolumn{2}{c}{\textbf{DPW Variants}} & \multicolumn{2}{c}{\textbf{AG Variants (Ours)}} & \multicolumn{2}{c}{\textbf{POMCP Variants}} \\
        $n^{\text{sims}}$ / $n_{\text{POMCPOW}}^{\text{sims}}$ & PFT-DPW & PFT-VPW & AG-PFT-DPW & AG-PFT-VPW & POMCPOW & VOMCPOW \\
        \midrule
        \textbf{50 / 400} & $1.87\pm0.11$ & $2.65\pm0.11$ & $3.33\pm0.10$ & $\mathbf{4.03}\pm0.09$ & $2.99\pm0.11$ & $ 3.69^{\dagger}\pm0.10 $ \\
        & \scriptsize 0.017s & \scriptsize 0.030s & \scriptsize 4.726s & \scriptsize 5.439s & \scriptsize 0.095s & \scriptsize 0.228s \\
        \midrule
        \textbf{89 / 712} & $3.08\pm0.11$ & $ 4.30^{\dagger}\pm0.09 $ & $4.00\pm0.09$ & $ 4.35^{\dagger}\pm0.09 $ & $3.84\pm0.10$ & $\mathbf{4.63}\pm0.10$ \\
        & \scriptsize 0.006s & \scriptsize 0.006s & \scriptsize 0.012s & \scriptsize 0.013s & \scriptsize 0.006s & \scriptsize 0.006s \\
        \midrule
        \textbf{158 / 1264} & $4.11\pm0.10$ & $5.10\pm0.09$ & $4.40\pm0.08$ & $4.43\pm0.09$ & $3.15\pm0.11$ & $\mathbf{5.56}\pm0.08$ \\
        & \scriptsize 0.011s & \scriptsize 0.011s & \scriptsize 0.032s & \scriptsize 0.029s & \scriptsize 0.018s & \scriptsize 0.016s \\
        \midrule
        \textbf{281 / 2248} & $4.69\pm0.09$ & $5.34\pm0.09$ & $4.59\pm0.08$ & $4.77\pm0.08$ & $5.16\pm0.09$ & $\mathbf{6.05}\pm0.08$ \\
        & \scriptsize 0.017s & \scriptsize 0.018s & \scriptsize 0.103s & \scriptsize 0.075s & \scriptsize 0.019s & \scriptsize 0.020s \\
        \midrule
        \textbf{500 / 4000} & $5.29\pm0.08$ & $\mathbf{5.98}\pm0.08$ & $4.86\pm0.08$ & $5.07\pm0.08$ & $ 5.73^{\dagger}\pm0.08 $ & $ 5.81^{\dagger}\pm0.08 $ \\
        & \scriptsize 0.027s & \scriptsize 0.027s & \scriptsize 0.316s & \scriptsize 0.191s & \scriptsize 0.033s & \scriptsize 0.033s \\
        \bottomrule
    \end{tabular}
\end{table*}

\begin{table*}[h]
    \caption{
        Performance (Mean $\pm$ SEM) and Runtimes in 3D-Continuous Light-Dark.
    }
    \label{tab:3d_light_dark_complete}
    \centering
    \small
    \begin{tabular}{l cc cc cc}
        \toprule
        \textbf{Simulations} & \multicolumn{2}{c}{\textbf{DPW Variants}} & \multicolumn{2}{c}{\textbf{AG Variants (Ours)}} & \multicolumn{2}{c}{\textbf{POMCP Variants}} \\
        $n^{\text{sims}}$ / $n_{\text{POMCPOW}}^{\text{sims}}$ & PFT-DPW & PFT-VPW & AG-PFT-DPW & AG-PFT-VPW & POMCPOW & VOMCPOW \\
        \midrule
        \textbf{50 / 566} & $ 1.10^{\dagger}\pm0.09 $ & $0.16\pm0.09$ & $\mathbf{1.29}\pm0.09$ & $0.56\pm0.09$ & $ 1.11^{\dagger}\pm0.09 $ & $ 1.21^{\dagger}\pm0.10 $ \\
        & \scriptsize 0.020s & \scriptsize 0.031s & \scriptsize 4.367s & \scriptsize 0.026s & \scriptsize 0.108s & \scriptsize 0.196s \\
        \midrule
        \textbf{89 / 1007} & $1.92\pm0.09$ & $1.13\pm0.09$ & $ 1.97^{\dagger}\pm0.09 $ & $ 2.01^{\dagger}\pm0.10 $ & $1.61\pm0.09$ & $\mathbf{2.33}\pm0.10$ \\
        & \scriptsize 0.013s & \scriptsize 0.015s & \scriptsize 0.021s & \scriptsize 3.673s & \scriptsize 0.011s & \scriptsize 0.012s \\
        \midrule
        \textbf{158 / 1788} & $2.55\pm0.09$ & $2.62\pm0.09$ & $2.51\pm0.09$ & $ 2.83^{\dagger}\pm0.09 $ & $2.39\pm0.09$ & $\mathbf{3.10}\pm0.10$ \\
        & \scriptsize 0.020s & \scriptsize 0.023s & \scriptsize 0.048s & \scriptsize 0.032s & \scriptsize 0.027s & \scriptsize 0.024s \\
        \midrule
        \textbf{281 / 3179} & $2.94\pm0.09$ & $3.48\pm0.09$ & $3.20\pm0.08$ & $3.50\pm0.08$ & $2.77\pm0.09$ & $\mathbf{4.26}\pm0.09$ \\
        & \scriptsize 0.035s & \scriptsize 0.040s & \scriptsize 0.125s & \scriptsize 0.063s & \scriptsize 0.031s & \scriptsize 0.035s \\
        \midrule
        \textbf{500 / 5657} & $3.52\pm0.08$ & $3.90\pm0.09$ & $3.34\pm0.08$ & $4.17\pm0.08$ & $3.20\pm0.08$ & $\mathbf{4.80}\pm0.08$ \\
        & \scriptsize 0.068s & \scriptsize 0.073s & \scriptsize 0.315s & \scriptsize 0.154s & \scriptsize 0.060s & \scriptsize 0.063s \\
        \bottomrule
    \end{tabular}
\end{table*}

\begin{table*}[h]
    \caption{
        Performance (Mean $\pm$ SEM) and Runtimes in 4D-Continuous Light-Dark.
    }
    \label{tab:4d_light_dark_complete}
    \centering
    \small
    \begin{tabular}{l cc cc cc}
        \toprule
        \textbf{Simulations} & \multicolumn{2}{c}{\textbf{DPW Variants}} & \multicolumn{2}{c}{\textbf{AG Variants (Ours)}} & \multicolumn{2}{c}{\textbf{POMCP Variants}} \\
        $n^{\text{sims}}$ / $n_{\text{POMCPOW}}^{\text{sims}}$ & PFT-DPW & PFT-VPW & AG-PFT-DPW & AG-PFT-VPW & POMCPOW & VOMCPOW \\
        \midrule
        \textbf{50 / 800} & $-0.28\pm0.07$ & $-0.89\pm0.06$ & $ -0.16^{\dagger}\pm0.07 $ & $-0.66\pm0.06$ & $\mathbf{0.07}\pm0.06$ & $ -0.10^{\dagger}\pm0.07 $ \\
        & \scriptsize 0.028s & \scriptsize 0.038s & \scriptsize 3.440s & \scriptsize 0.031s & \scriptsize 0.096s & \scriptsize 0.195s \\
        \midrule
        \textbf{89 / 1424} & $ 0.30^{\dagger}\pm0.08 $ & $ 0.34^{\dagger}\pm0.08 $ & $\mathbf{0.41}\pm0.07$ & $0.05\pm0.07$ & $0.09\pm0.07$ & $ 0.27^{\dagger}\pm0.08 $ \\
        & \scriptsize 0.031s & \scriptsize 0.029s & \scriptsize 0.038s & \scriptsize 3.322s & \scriptsize 0.020s & \scriptsize 0.020s \\
        \midrule
        \textbf{158 / 2528} & $1.03\pm0.08$ & $0.86\pm0.08$ & $ 1.09^{\dagger}\pm0.07 $ & $ 1.15^{\dagger}\pm0.08 $ & $1.01\pm0.07$ & $\mathbf{1.39}\pm0.08$ \\
        & \scriptsize 0.049s & \scriptsize 0.045s & \scriptsize 0.067s & \scriptsize 0.061s & \scriptsize 0.035s & \scriptsize 0.036s \\
        \midrule
        \textbf{281 / 4496} & $1.59\pm0.08$ & $1.43\pm0.08$ & $1.49\pm0.07$ & $\mathbf{2.43}\pm0.08$ & $1.23\pm0.07$ & $ 2.34^{\dagger}\pm0.08 $ \\
        & \scriptsize 0.078s & \scriptsize 0.079s & \scriptsize 0.143s & \scriptsize 0.132s & \scriptsize 0.056s & \scriptsize 0.064s \\
        \midrule
        \textbf{500 / 8000} & $1.98\pm0.08$ & $2.28\pm0.08$ & $2.10\pm0.07$ & $ 2.97^{\dagger}\pm0.08 $ & $1.61\pm0.08$ & $\mathbf{3.04}\pm0.08$ \\
        & \scriptsize 0.151s & \scriptsize 0.155s & \scriptsize 0.302s & \scriptsize 0.318s & \scriptsize 0.141s & \scriptsize 0.170s \\
        \bottomrule
    \end{tabular}
\end{table*}

\begin{table*}[h]
    \caption{
        Performance (Mean $\pm$ SEM) and Runtimes in Two-Agent 2D-Continuous Light-Dark.
    }
    \label{tab:collab_light_dark_complete}
    \centering
    \small
    \begin{tabular}{l cc cc cc}
        \toprule
        \textbf{Simulations} & \multicolumn{2}{c}{\textbf{DPW Variants}} & \multicolumn{2}{c}{\textbf{AG Variants (Ours)}} & \multicolumn{2}{c}{\textbf{POMCP Variants}} \\
        $n^{\text{sims}}$ / $n_{\text{POMCPOW}}^{\text{sims}}$ & PFT-DPW & PFT-VPW & AG-PFT-DPW & AG-PFT-VPW & POMCPOW & VOMCPOW \\
        \midrule
        \textbf{50 / 800} & $1.10\pm0.11$ & $-0.01\pm0.10$ & $\mathbf{2.14}\pm0.11$ & $ 1.82^{\dagger}\pm0.11 $ & $ 1.98^{\dagger}\pm0.10 $ & $1.66\pm0.10$ \\
        & \scriptsize 0.035s & \scriptsize 0.045s & \scriptsize 12.265s & \scriptsize 11.980s & \scriptsize 0.140s & \scriptsize 0.241s \\
        \midrule
        \textbf{89 / 1424} & $1.47\pm0.11$ & $0.43\pm0.10$ & $ 1.81^{\dagger}\pm0.10 $ & $ 1.58^{\dagger}\pm0.10 $ & $\mathbf{1.96}\pm0.10$ & $ 1.56^{\dagger}\pm0.10 $ \\
        & \scriptsize 0.036s & \scriptsize 0.037s & \scriptsize 0.057s & \scriptsize 0.056s & \scriptsize 0.020s & \scriptsize 0.021s \\
        \midrule
        \textbf{158 / 2528} & $1.44\pm0.11$ & $1.19\pm0.11$ & $\mathbf{2.31}\pm0.11$ & $ 2.17^{\dagger}\pm0.10 $ & $ 2.13^{\dagger}\pm0.10 $ & $ 2.27^{\dagger}\pm0.11 $ \\
        & \scriptsize 0.063s & \scriptsize 0.059s & \scriptsize 0.109s & \scriptsize 0.108s & \scriptsize 0.040s & \scriptsize 0.042s \\
        \midrule
        \textbf{281 / 4496} & $1.70\pm0.10$ & $1.55\pm0.11$ & $\mathbf{2.72}\pm0.11$ & $ 2.48^{\dagger}\pm0.10 $ & $2.28\pm0.10$ & $ 2.41^{\dagger}\pm0.11 $ \\
        & \scriptsize 0.088s & \scriptsize 0.096s & \scriptsize 0.214s & \scriptsize 0.208s & \scriptsize 0.065s & \scriptsize 0.076s \\
        \midrule
        \textbf{500 / 8000} & $1.91\pm0.10$ & $1.67\pm0.11$ & $\mathbf{2.84}\pm0.10$ & $1.18\pm0.10$ & $1.83\pm0.10$ & $ 2.57^{\dagger}\pm0.11 $ \\
        & \scriptsize 0.160s & \scriptsize 0.173s & \scriptsize 0.447s & \scriptsize 0.464s & \scriptsize 0.141s & \scriptsize 0.169s \\
        \bottomrule
    \end{tabular}
\end{table*}

The full AGMCTS algorithm description is outlined in Algorithm~\ref{alg:AGMCTS_full}. It is based on MCTS with DPW, with new components highlighted in blue.
We discuss several of the heuristics and optimizations that we found to be crucial for the performance of AGMCTS.

\subsection{Monte Carlo Gradient Estimation}
When estimating the action score gradient of the action-value function, we subsample the successor branches, or states of particle-beliefs when computing gradient estimates, greatly reducing computational cost.

\subsubsection{Transition Model Log-Probability}

Via automatic differentiation and the chain rule, we assume we can compute $\nabla_a \log p_T(s^{\prime} {\mid} s, a)$ for any given transition triplet $(s,a,s^{\prime})$.

In POMDPs, we do not compute exact propagated belief probabilities $p(\bar{b}^{\prime -} {\mid} \bar{b}, a)$, but rather approximate them via subsampling their gradient.
We estimate \eqref{eq:propagated_belief_grad} with $K_b^{\nabla}$ state particles via 
\begin{linenomath*}
    \begin{gather}
        \hat{\nabla}_a \log p(\bar{b}^{\prime -}{\mid} \bar{b},a) = \frac{J}{K_{b}^{\nabla}}\sum_{l=1}^{K_{b}^{\nabla}} \nabla_a \log p_T(s^{\prime -,j_l}{\mid} s^{j_l}, a),
        \label{eq:propagated_belief_grad_est}
    \end{gather}
\end{linenomath*}
where indices $j_l$ are sampled uniformly from $[1,\dots,J]$, where $J$ is the particle count of $\bar{b}$ and $\bar{b}^{\prime -}$. Since this is a Monte Carlo approximation of the sum \eqref{eq:propagated_belief_grad_rpt}, the estimator \eqref{eq:propagated_belief_grad_est} is unbiased.

\subsubsection{Action-Gradient Estimates of the Action-Value}

In the MDP case, we use the baseline function $B(s)=\hat{V}(s)$ when estimating the action-gradient of the action-value function, essentially estimating advantage gradients.
In the POMDP case, we use $B(\bar{b})=\hat{V}(\bar{b})$ similarly.

When linearizing the importance weights update in Equation \eqref{eq:imp_ratio_linear_update}, we have to compute $\nabla_{a}\log p_{T}(s^{\prime,i} {\mid} s,a)$ (or $\hat{\nabla}_{a}\log p(\bar{b}^{- \prime,i} {\mid} \bar{b},a)$) for all $i=1,\dots,\absvalflat{\mathcal{S}_{sa}}$ to compute their updated importance weight.
Since this is the most expensive term when computing $\hat{\nabla}\tilde{Q}(s,a)$, it makes sense to cache the transition log-gradient terms, and use all successor branches for gradient estimation.
Therefore, we sum over all successor states $s^{\prime}\in \mathcal{S}_{sa}$, weighted by the visitation count and the importance ratio:
\begin{linenomath*}
    \begin{align}
        &\hat{\nabla}_{a}\tilde{Q}(s,a)
        \nonumber \\
        &\begin{multlined}[b][0.8\columnwidth]
            =\frac{1}{\eta(s,a)}\sum_{i=1}^{\absval{\mathcal{S}_{sa}}} n(s^{\prime,i})_{+1} \rho_{a}^{i}(s^{\prime,i}) \cdot \bigg( \nabla_{a}\log p_{T}(s^{\prime,i} {\mid} s,a)
            \\
            \cdot(r(s,a,s^{\prime,i})+\gamma\hat{V}(s^{\prime,i}) - B(s))+\nabla_{a}r(s,a,s^{\prime,i}) \bigg),
        \end{multlined}
        \label{eq:mdp_grad_est_linearized_rpt}
    \end{align}
\end{linenomath*}
and for POMDPs:
\begin{linenomath*}
    \begin{align}
        &\hat{\nabla}_{a}\tilde{Q}(\bar{b},a)
        \nonumber \\
        &\begin{multlined}[b][0.8\columnwidth]
            =\frac{1}{\eta(\bar{b},a)}\sum_{i=1}^{\absval{\mathcal{S}_{sa}}} n(\bar{b}^{\prime,i})_{+1} \rho_{a}^{i}(\bar{b}^{- \prime,i}) \cdot \bigg( \hat{\nabla}_{a}\log p(\bar{b}^{- \prime,i} {\mid} \bar{b},a)
            \\
            \cdot(r(\bar{b},a,\bar{b}^{\prime,i})+\gamma\hat{V}(\bar{b}^{\prime,i}) - B(\bar{b}))+\nabla_{a}r(\bar{b},a,\bar{b}^{\prime,i}) \bigg).
        \end{multlined}
        \label{eq:pomdp_grad_est_linearized_rpt}
    \end{align}
\end{linenomath*}

When using direct state sampling based on Theorem~\ref{thm:action_grad_state_reward_rpt}, we sample $K_r^{\nabla}$ new successor states $s_{new}^{\prime,j_m}\sim p_T(\cdot{\mid} s,a)$, for $m=1,\dots,K_r^{\nabla}$, and we compute:
\begin{linenomath*}
    \begin{align}
        &\hat{\nabla}_{a}\hat{Q}(s,a)
        \nonumber \\
        &\begin{multlined}[b][0.9\columnwidth]
            =\frac{1}{K_{r}^{\nabla}} \big(\sum_{m=1}^{K_{r}^{\nabla}}\nabla_{a}\log p_T(s_{new}^{\prime,j_m} {\mid} s,a)\cdot 
            r(s,a,s_{new}^{\prime,j_m})
            \\
            +\nabla_{a}r(s,a,s_{new}^{\prime,j_m}) \big) 
        \end{multlined}
        \nonumber \\
        &\begin{multlined}[b][0.9\columnwidth]    
            +\frac{1}{\eta(s,a)}\sum_{i=1}^{\absval{\mathcal{S}_{sa}}} n(s^{\prime,i})_{+1} \rho_{a}^{i}(s^{\prime,i}) \cdot \nabla_{a}\log p_{T}(s^{\prime,i} {\mid} s,a)
            \\
            \cdot(\gamma\hat{V}(s^{\prime,i}) - B(s)),
        \end{multlined}
        \nonumber \\
        \label{eq:mdp_state_grad_est_linearized_rpt}
    \end{align}
\end{linenomath*}
In the POMDP case, we generate $K_r^{\nabla}$ posterior states by sampling from the belief $s^{j_m}\sim \bar{b}$, and sampling new states $s_{new}^{\prime,j_m}\sim p_T(\cdot{\mid} s^{j_m},a)$, for $m=1,\dots,K_r^{\nabla}$, where $J$ is the number of particles of $\bar{b}$:
\begin{linenomath*}
    \begin{align}
        &\hat{\nabla}_{a}\hat{Q}(s,a)
        \nonumber \\
        &\begin{multlined}[b][0.9\columnwidth]
            =\frac{1}{K_{r}^{\nabla}} \big(\sum_{m=1}^{K_{r}^{\nabla}}\nabla_{a}\log p_T(s_{new}^{\prime,j_m} {\mid} s^{j_m},a)\cdot 
            r(s^{j_m},a,s_{new}^{\prime,j_m})
            \\
            +\nabla_{a}r(s^{j_m},a,s_{new}^{\prime,j_m}) \big) 
        \end{multlined}
        \nonumber \\
        &\begin{multlined}[b][0.9\columnwidth]    
            +\frac{1}{\eta(s,a)}\sum_{i=1}^{\absval{\mathcal{S}_{sa}}} n(\bar{b}^{\prime,i})_{+1} \rho_{a}^{i}(\bar{b}^{- \prime,i}) \cdot \hat{\nabla}_{a}\log p(\bar{b}^{- \prime,i} {\mid} \bar{b},a)
            \\
            \cdot(\gamma\hat{V}(\bar{b}^{\prime,i}) - B(\bar{b})).
        \end{multlined}
        \nonumber \\
        \label{eq:pomdp_state_grad_est_linearized_rpt}
    \end{align}
\end{linenomath*}

\textit{Note: The following estimators describe exact importance weight updates. In practice, we have not used these in any of the reported experiments in Section \ref{sec:appendix_experiments}. We bring these estimators for completeness.}

When performing exact importance weight updates \eqref{eq:action_update_rpt}, we sample successor branches to compute the gradient, rather than computing over all successor states.
We sample $K_{O}^{\nabla}$ successor state indices $i_k$ from $[1,\dots,\absval{\mathcal{S}_{sa}}]$, weighted proportionally to the visitation count times the importance ratio, i.e. from the discrete distribution on $[1,\dots,\absval{\mathcal{S}_{sa}}]$ with weights $\propto n(s^{\prime,i})_{+1}\cdot \rho_{a}^{i}(s^{\prime,i})$.
Thus, the MC approximation of \eqref{eq:mdp_action_grad_rpt} becomes
\begin{linenomath*}
    \begin{align}
        \begin{multlined}[b][\columnwidth]
            \hat{\nabla}_{a}\tilde{Q}(s,a)=\frac{1}{K_{O}^{\nabla}}\sum_{k=1}^{K_{O}^{\nabla}}\nabla_{a}\log p_{T}(s^{\prime,i_{k}} {\mid} s,a)
            \\
            \cdot(r(s,a,s^{\prime,i_{k}})+\gamma\hat{V}(s^{\prime,i_{k}}) - B(s))+\nabla_{a}r(s,a,s^{\prime,i_{k}}),
        \end{multlined}
        \label{eq:mdp_grad_est}
    \end{align}
\end{linenomath*}
and for POMDPs:
\begin{linenomath*}
    \begin{align}
        \begin{multlined}[b][\columnwidth]
            \hat{\nabla}_{a}\tilde{Q}(\bar{b},a)=\frac{1}{K_{O}^{\nabla}}\sum_{k=1}^{K_{O}^{\nabla}} \hat{\nabla}_{a}\log p(\bar{b}^{- \prime,i_{k}} {\mid} \bar{b},a)
            \\
            \cdot(r(\bar{b},a,\bar{b}^{\prime,i_{k}})+\gamma\hat{V}(\bar{b}^{\prime,i_{k}}) - B(\bar{b}))+\nabla_{a}r(\bar{b},a,\bar{b}^{\prime,i_{k}}).
        \end{multlined}
        \label{eq:pomdp_grad_est}
    \end{align}
\end{linenomath*}

When using direct state sampling based on Theorem~\ref{thm:action_grad_state_reward_rpt}, we similarly sample $K_r^{\nabla}$ new successor states $s_{new}^{\prime,j_m}\sim p_T(\cdot{\mid} s,a)$, for $m=1,\dots,K_r^{\nabla}$, and we compute:
\begin{linenomath*}
    \begin{align}
        &\hat{\nabla}_{a}\hat{Q}(s,a)
        \nonumber \\
        &\begin{multlined}[b][0.9\columnwidth]
            =\frac{1}{K_{r}^{\nabla}} \big(\sum_{m=1}^{K_{r}^{\nabla}}\nabla_{a}\log p_T(s_{new}^{\prime,j_m} {\mid} s,a)\cdot 
            r(s,a,s_{new}^{\prime,j_m})
            \\
            +\nabla_{a}r(s,a,s_{new}^{\prime,j_m}) \big) 
        \end{multlined}
        \nonumber \\
        &\begin{multlined}[b][0.9\columnwidth]    
            + \frac{1}{K_{O}^{\nabla}}\sum_{k=1}^{K_{O}^{\nabla}}
            \nabla_a \log p_T(s^{\prime,i_{k}}{\mid} s, a)
            \cdot (\gamma \cdot (\hat{V}(s^{\prime,i_{k}})) - B(s) ).
        \end{multlined}
        \label{eq:mdp_state_grad_est}
    \end{align}
\end{linenomath*}

And again in the POMDP case, we generate $K_r^{\nabla}$ posterior states by sampling from the belief $s^{j_m}\sim \bar{b}$, and sampling new states $s_{new}^{\prime,j_m}\sim p_T(\cdot{\mid} s^{j_m},a)$, for $m=1,\dots,K_r^{\nabla}$, where $J$ is the number of particles of $\bar{b}$:
\begin{linenomath*}
\begin{align}
        &\hat{\nabla}_{a}\hat{Q}(\bar{b},a)
        \nonumber \\
        &\begin{multlined}[b][0.9\columnwidth]
            =\frac{1}{K_{r}^{\nabla}} \big(\sum_{m=1}^{K_{r}^{\nabla}}\nabla_{a}\log p_T(s_{new}^{\prime,j_m} {\mid} s^{j_m},a)\cdot 
            r(s^{j_m},a,s_{new}^{\prime,j_m})
            \\
            +\nabla_{a}r(s^{j_m},a,s_{new}^{\prime,j_m}) \big) 
        \end{multlined}
        \nonumber \\
        &\begin{multlined}[b][0.9\columnwidth]    
            + \frac{1}{K_{O}^{\nabla}}\sum_{k=1}^{K_{O}^{\nabla}}
            \hat{\nabla}_{a}\log p(\bar{b}^{- \prime,i_{k}}{\mid} \bar{b}, a)
            \cdot (\gamma \hat{V}(\bar{b}^{\prime,i_{k}}) - B(\bar{b}) ).
        \end{multlined}
        \label{eq:pomdp_state_grad_est}
    \end{align}
\end{linenomath*}

\subsection{Action Update Rule}

Since we compute advantage gradients, we only perform \textit{action update} when we have more than one successor state in order to have a meaningful estimate of the gradient.
In the case of a single successor, the advantage is always zero, leading to zero gradient estimates (unless the reward is action-dependent).
Hence, to save computational cost, we only perform \textit{action update} when $\absval{\mathcal{S}_{sa}}>=K_{\text{child}}^{\min}$, where we set $K_{\text{child}}^{\min}=2$ in all our experiments.

\subsection{Gradient Optimization}

We've found throughout our experiments that the Adam optimizer~\cite{Kingma15iclr} that normalizes the step size was crucial for consistent behavior due to the high variance of the gradient estimates.
We've fixed as a hyperparameter $K_{\text{opt}}$ the number of consecutive gradient optimization iterations before \textit{simulation}.
Together with the step size, this controls a trade-off between accuracy and computational complexity.
To further ensure moderate step-sizes, we also set a max-norm threshold on the final action update $\normflat{a - \breve{a}}<T_{d_a}^{\max}$, and scaling down the step norm to $T_{d_a}^{\max}$ if necessary.

In some domains we used an exponential decay on the step size after the Adam step size.
The calculation for the update of the accumulated action is done as:
\begin{linenomath*}
    \begin{align*}
        \delta_{Adam} &= \text{\tallet{A}DAM}(a, g_a^q), \\
        \lambda &= \max\{\num{0.999}^{T}, \num{0.1} \}, \\
        \breve{a} &= a + \lambda \delta_{Adam},
    \end{align*}
\end{linenomath*}
where $T$ is the number of gradient optimizations performed at this particular action node.

For the Adam optimizer, we used the standard hyperparameters $\beta_1=0.9$, $\beta_2=0.999$, $\epsilon=10^{-8}$. The learning rate $\eta_{\text{Adam}}$ was determined by cross entropy optimization (described in Section \ref{sec:appendix_experiments}), in conjunction with the optimization over the other algorithm hyperparameters.

\subsection{Thresholds For Adding/Deleting State Nodes}
To save computational overhead, we would like to ensure that we have relevant samples after several action updates.
After \textit{action update} to action $a$, we delete child state $s^{\prime,i}$ from $\mathcal{S}_{sa}$ if its importance ratio satisfies $\rho_{a}^{i}(s^{\prime,i})<T_{\rho}^{del}$.
Additionally, if $\rho_{a}^{i}(s^{\prime,i})<T_{\rho}^{add}$ for all $s^{\prime,i}\in \mathcal{S}_{sa}$, we force sampling a new posterior node, overriding the progressive widening limitation.
The deletion threshold avoids spending computation on states with low contribution to $\tilde{Q}(s,a)$, while the addition threshold ensures that at least one relevant sample is available after an action update.

\subsection{Action Sampling Heuristics}

In this section, we describe the action progressive widening strategies used to handle continuous action spaces.
We detail the vanilla Action Progressive Widening (APW) used in algorithms like POMCPOW~\cite{Sunberg18icaps}, and the Voronoi Progressive Widening (VPW) proposed by~\citeauthor{Lim21cdc}~\cite{Lim21cdc}.

\subsubsection{Action Progressive Widening}
Action Progressive Widening (APW) limits the branching factor of the search tree by progressively adding new action branches as the number of visits to the state node increases. New actions are sampled uniformly from the action space $\mathcal{A}$.

\subsubsection{Voronoi Progressive Widening}
VPW generalizes APW by utilizing Voronoi Optimistic Optimization (VOO)~\cite{Kim20aaai} to guide the generation of new actions. 
Instead of sampling uniformly, VPW uses the existing action-value estimates to sample new actions from the Voronoi cell of the current best action with high probability, balancing exploration and exploitation during the widening phase.

The \textsc{VOO} procedure samples a new action by first selecting the best existing action $a^* = \argmax_{a \in \mathcal{A}_{s}} Q(s,a)$, and then sampling a new candidate $a$ from a distribution centered at $a^*$ such that $a$ lies within the Voronoi cell of $a^*$. 
With a probability $\omega_{\textnormal{VOO}}$, it samples uniformly from $\mathcal{A}$ to ensure global exploration.
In our implementation, we used a Gaussian distribution centered at $a^*$ with covariance $\Sigma_{\textnormal{VOO}}$ for sampling new actions within the Voronoi cell.

\section{Experiments}
\label{sec:appendix_experiments}

The highest mean result in each row is bolded. A dagger ($\dagger$) indicates a mean that overlaps with a higher mean by at most 2 SEM.

\begin{table}[h]
    \caption{Manual hyperparameters in the Mountain/Hill Car domains.}
    \label{tab:mountain_hill_manual_hyperparameters}
    \centering
    \small
    \begin{tabular}{l c c}
        \toprule
        \textbf{Param.} & \textbf{Mountain MDP/POMDP} & \textbf{Hill MDP/POMDP} \\
        \midrule
        $J$              & - / 30 & - / 30 \\
        $J^{\text{PF}}$  & - / 200 & - / 200 \\
        $K_{rollout}$    & 1 / 5 & 1 / 5 \\
        $K_{\text{opt}}$ & 3 & 3 / 2 \\
        $T_{d_a}^{\max}$ & $\sigma_{\xi}$ & $\sigma_{\xi}$ \\
        $T_{\rho}^{add}$ & 1.0 / 0.99 & 1.0 / 0.99 \\
        $T_{\rho}^{del}$ & 0.5 / \num{1e-8} & 0.5 / 0.01 \\
        $K_{b}^{\nabla}$ & - / 3 & - / 3 \\
        \bottomrule
    \end{tabular}
\end{table}

\begin{table}[h]
    \caption{Hyperparameters used in the evaluations of each algorithm in the Mountain Car MDP domain.}
    \label{tab:mountain_car_mdp_hyperparameters}
    \centering
    \small
    \begin{tabular}{l cc cc}
        \toprule
        & \multicolumn{2}{c}{\textbf{DPW Variants}} & \multicolumn{2}{c}{\textbf{AG Variants (Ours)}} \\
        \textbf{Param.} & DPW & VPW & AG-DPW & AG-VPW \\
        \midrule
        $c$             & 112.20 & 116.80 & 0.0 & 39.90 \\
        $k_a$           & 6.13 & 2.09 & 5.02 & 9.08 \\
        $\alpha_a$      & 0.60 & 0.72 & 0.67 & \num{2.3e-2} \\
        $k_o$           & 0.24 & 0.28 & 0.20 & 3.38 \\
        $\alpha_o$      & 0.36 & 0.62 & 0.57 & 0.54 \\
        \midrule
        $\eta_{\text{Adam}}$ & - & - & \num{4.0e-4} & 0.11 \\
        \bottomrule
    \end{tabular}
\end{table}

\begin{table*}[h]
    \caption{Hyperparameters used in the evaluations of each algorithm in the Mountain Car POMDP domain.}
    \label{tab:mountain_car_pomdp_hyperparameters}
    \centering
    \small
    \begin{tabular}{l cc cc cc}
        \toprule
        & \multicolumn{2}{c}{\textbf{DPW Variants}} & \multicolumn{2}{c}{\textbf{AG Variants (Ours)}} & \multicolumn{2}{c}{\textbf{POMCP Variants}} \\
        \textbf{Param.} & PFT-DPW & PFT-VPW & AG-PFT-DPW & AG-PFT-VPW & POMCPOW & VOMCPOW \\
        \midrule
        $c$             & 119.06 & 139.54 & 43.36 & 46.97 & 144.46 & 70.89 \\
        $k_a$           & 5.40 & 8.96 & 3.11 & 3.10 & 3.93 & 5.23 \\
        $\alpha_a$      & 0.87 & 0.76 & \num{2.6e-2} & \num{2.9e-2} & 0.64 & 0.67 \\
        $k_o$           & 0.92 & 3.23 & 5.69 & 5.63 & 0.47 & 0.37 \\
        $\alpha_o$      & 0.48 & 0.30 & 0.46 & 0.58 & \num{9.8e-2} & 0.48 \\
        \midrule
        $\eta_{\text{Adam}}$ & - & - & \num{1.7e-2} & \num{1.4e-2} & - & - \\
        \bottomrule
    \end{tabular}
\end{table*}

\begin{table}[h]
    \caption{Hyperparameters used in the evaluations of each algorithm in the Hill Car MDP domain.}
    \label{tab:hill_car_mdp_hyperparameters}
    \centering
    \small
    \begin{tabular}{l cc cc}
        \toprule
        & \multicolumn{2}{c}{\textbf{DPW Variants}} & \multicolumn{2}{c}{\textbf{AG Variants (Ours)}} \\
        \textbf{Param.} & DPW & VPW & AG-DPW & AG-VPW \\
        \midrule
        $c$             & 177.99 & 135.07 & 169.92 & 173.43 \\
        $k_a$           & 6.73 & 3.79 & 6.66 & 1.28 \\
        $\alpha_a$      & 0.62 & 0.71 & 0.37 & 0.54 \\
        $k_o$           & 0.52 & 0.59 & 7.44 & 6.39 \\
        $\alpha_o$      & 0.26 & 0.72 & 0.32 & 0.26 \\
        \midrule
        $\eta_{\text{Adam}}$ & - & - & \num{4.6e-6} & \num{5.8e-5} \\
        \bottomrule
    \end{tabular}
\end{table}

\begin{table*}[h]
    \caption{Hyperparameters used in the evaluations of each algorithm in the Hill Car POMDP domain.}
    \label{tab:hill_car_pomdp_hyperparameters}
    \centering
    \small
    \begin{tabular}{l cc cc cc}
        \toprule
        & \multicolumn{2}{c}{\textbf{DPW Variants}} & \multicolumn{2}{c}{\textbf{AG Variants (Ours)}} & \multicolumn{2}{c}{\textbf{POMCP Variants}} \\
        \textbf{Param.} & PFT-DPW & PFT-VPW & AG-PFT-DPW & AG-PFT-VPW & POMCPOW & VOMCPOW \\
        \midrule
        $c$             & 162.86 & \num{5.2e-7} & 131.41 & 136.45 & 101.51 & 70.13 \\
        $k_a$           & 10.31 & 4.80 & 5.55 & 1.95 & 9.94 & 7.44 \\
        $\alpha_a$      & 0.56 & 0.49 & 0.30 & 0.55 & 0.18 & 0.36 \\
        $k_o$           & 0.85 & 8.15 & 9.88 & 13.31 & 8.56 & 6.81 \\
        $\alpha_o$      & 0.53 & 0.32 & 0.58 & 0.42 & 0.84 & 0.73 \\
        \midrule
        $\eta_{\text{Adam}}$ & - & - & \num{8.1e-6} & \num{1.5e-5} \\
        \bottomrule
    \end{tabular}
\end{table*}

\begin{table*}[h]
    \caption{
        Performance (Mean $\pm$ SEM) and Runtimes in Mountain Car MDP.
    }
    \label{tab:mountain_car_mdp_complete}
    \centering
    \small
    \begin{tabular}{l cc cc}
        \toprule
        \textbf{Simulations} & \multicolumn{2}{c}{\textbf{DPW Variants}} & \multicolumn{2}{c}{\textbf{AG Variants (Ours)}} \\
        $n^{\text{sims}}$ & DPW & VPW & AG-DPW & AG-VPW \\
        \midrule
        \textbf{50} & $-52.00\pm0.17$ & $-50.50\pm0.29$ & $-52.00\pm0.17$ & $\mathbf{-4.30}\pm1.17$ \\
        & \scriptsize 0.007s & \scriptsize 0.008s & \scriptsize 0.009s & \scriptsize 0.036s \\
        \midrule
        \textbf{89} & $-46.18\pm0.68$ & $-50.33\pm0.31$ & $-51.66\pm0.25$ & $\mathbf{7.26}\pm1.03$ \\
        & \scriptsize 0.004s & \scriptsize 0.004s & \scriptsize 0.007s & \scriptsize 0.015s \\
        \midrule
        \textbf{158} & $-47.46\pm0.56$ & $-50.86\pm0.27$ & $ 12.55^{\dagger}\pm1.03 $ & $\mathbf{16.16}\pm0.79$ \\
        & \scriptsize 0.006s & \scriptsize 0.006s & \scriptsize 0.056s & \scriptsize 0.027s \\
        \midrule
        \textbf{281} & $-45.71\pm0.64$ & $24.42\pm0.09$ & $\mathbf{29.68}\pm0.06$ & $21.27\pm0.57$ \\
        & \scriptsize 0.010s & \scriptsize 0.009s & \scriptsize 0.060s & \scriptsize 0.065s \\
        \midrule
        \textbf{500} & $24.24\pm0.38$ & $24.33\pm0.05$ & $\mathbf{29.97}\pm0.06$ & $25.94\pm0.12$ \\
        & \scriptsize 0.016s & \scriptsize 0.016s & \scriptsize 0.148s & \scriptsize 0.152s \\
        \bottomrule
    \end{tabular}
\end{table*}

\begin{table*}[h]
    \caption{
        Performance (Mean $\pm$ SEM) and Runtimes in Mountain Car POMDP.
    }
    \label{tab:mountain_car_pomdp_complete}
    \centering
    \small
    \begin{tabular}{l cc cc cc}
        \toprule
        \textbf{Simulations} & \multicolumn{2}{c}{\textbf{DPW Variants}} & \multicolumn{2}{c}{\textbf{AG Variants (Ours)}} & \multicolumn{2}{c}{\textbf{POMCP Variants}} \\
        $n^{\text{sims}}$ / $n_{\text{POMCPOW}}^{\text{sims}}$ & PFT-DPW & PFT-VPW & AG-PFT-DPW & AG-PFT-VPW & POMCPOW & VOMCPOW \\
        \midrule
        \textbf{50 / 274} & $ 24.64^{\dagger}\pm0.37 $ & $ 24.43^{\dagger}\pm0.38 $ & $ 24.96^{\dagger}\pm0.41 $ & $ 24.72^{\dagger}\pm0.44 $ & $ 24.80^{\dagger}\pm0.29 $ & $\mathbf{25.62}\pm0.36$ \\
        & \scriptsize 0.005s & \scriptsize 0.006s & \scriptsize 0.047s & \scriptsize 0.048s & \scriptsize 0.016s & \scriptsize 0.018s \\
        \midrule
        \textbf{89 / 487} & $ 24.54^{\dagger}\pm0.36 $ & $ 24.48^{\dagger}\pm0.35 $ & $ 25.56^{\dagger}\pm0.36 $ & $\mathbf{25.58}\pm0.37$ & $ 25.39^{\dagger}\pm0.31 $ & $ 25.46^{\dagger}\pm0.37 $ \\
        & \scriptsize 0.008s & \scriptsize 0.008s & \scriptsize 0.051s & \scriptsize 0.051s & \scriptsize 0.019s & \scriptsize 0.020s \\
        \midrule
        \textbf{158 / 865} & $24.17\pm0.38$ & $23.82\pm0.40$ & $ 25.21^{\dagger}\pm0.41 $ & $\mathbf{25.98}\pm0.34$ & $ 25.36^{\dagger}\pm0.33 $ & $ 25.38^{\dagger}\pm0.36 $ \\
        & \scriptsize 0.013s & \scriptsize 0.013s & \scriptsize 0.143s & \scriptsize 0.142s & \scriptsize 0.034s & \scriptsize 0.037s \\
        \midrule
        \textbf{281 / 1539} & $23.23\pm0.45$ & $23.14\pm0.44$ & $\mathbf{26.64}\pm0.26$ & $25.01\pm0.45$ & $24.75\pm0.40$ & $24.05\pm0.45$ \\
        & \scriptsize 0.023s & \scriptsize 0.024s & \scriptsize 0.376s & \scriptsize 0.366s & \scriptsize 0.063s & \scriptsize 0.069s \\
        \midrule
        \textbf{500 / 2739} & $23.20\pm0.43$ & $22.69\pm0.45$ & $\mathbf{26.96}\pm0.25$ & $ 26.27^{\dagger}\pm0.35 $ & $24.95\pm0.36$ & $24.11\pm0.44$ \\
        & \scriptsize 0.042s & \scriptsize 0.044s & \scriptsize 0.928s & \scriptsize 0.904s & \scriptsize 0.116s & \scriptsize 0.134s \\
        \bottomrule
    \end{tabular}
\end{table*}

\begin{table*}[h]
    \caption{
        Performance (Mean $\pm$ SEM) and Runtimes in Hill Car MDP.
    }
    \label{tab:hill_car_mdp_complete}
    \centering
    \small
    \begin{tabular}{l cc cc}
        \toprule
        \textbf{Simulations} & \multicolumn{2}{c}{\textbf{DPW Variants}} & \multicolumn{2}{c}{\textbf{AG Variants (Ours)}} \\
        $n^{\text{sims}}$ & DPW & VPW & AG-DPW & AG-VPW \\
        \midrule
        \textbf{50} & $ -92.87^{\dagger}\pm1.05 $ & $ -92.71^{\dagger}\pm1.06 $ & $\mathbf{-91.90}\pm1.12$ & $-98.05\pm0.56$ \\
        & \scriptsize 0.339s & \scriptsize 0.359s & \scriptsize 0.036s & \scriptsize 1.412s \\
        \midrule
        \textbf{89} & $-92.06\pm1.11$ & $-90.93\pm1.18$ & $\mathbf{56.27}\pm1.00$ & $25.66\pm2.16$ \\
        & \scriptsize 0.006s & \scriptsize 0.008s & \scriptsize 0.389s & \scriptsize 0.127s \\
        \midrule
        \textbf{158} & $-71.43\pm1.96$ & $-69.31\pm2.01$ & $-39.16\pm2.49$ & $\mathbf{39.03}\pm1.82$ \\
        & \scriptsize 0.023s & \scriptsize 0.022s & \scriptsize 0.123s & \scriptsize 0.221s \\
        \midrule
        \textbf{281} & $-66.67\pm2.08$ & $-67.08\pm2.07$ & $\mathbf{56.68}\pm0.97$ & $-6.36\pm2.55$ \\
        & \scriptsize 0.029s & \scriptsize 0.026s & \scriptsize 0.313s & \scriptsize 0.472s \\
        \midrule
        \textbf{500} & $-66.04\pm2.09$ & $-59.66\pm2.23$ & $\mathbf{56.34}\pm1.00$ & $44.56\pm1.62$ \\
        & \scriptsize 0.026s & \scriptsize 0.029s & \scriptsize 0.744s & \scriptsize 1.028s \\
        \bottomrule
    \end{tabular}
\end{table*}

\begin{table*}[h]
    \caption{
        Performance (Mean $\pm$ SEM) and Runtimes in Hill Car POMDP.
    }
    \label{tab:hill_car_pomdp_complete}
    \centering
    \small
    \begin{tabular}{l cc cc cc}
        \toprule
        \textbf{Simulations} & \multicolumn{2}{c}{\textbf{DPW Variants}} & \multicolumn{2}{c}{\textbf{AG Variants (Ours)}} & \multicolumn{2}{c}{\textbf{POMCP Variants}} \\
        $n^{\text{sims}}$ / $n_{\text{POMCPOW}}^{\text{sims}}$ & PFT-DPW & PFT-VPW & AG-PFT-DPW & AG-PFT-VPW & POMCPOW & VOMCPOW \\
        \midrule
        \textbf{50 / 274} & $-83.22\pm1.55$ & $-78.37\pm1.73$ & $48.63\pm1.34$ & $35.93\pm1.84$ & $\mathbf{53.83}\pm0.98$ & $21.96\pm2.18$ \\
        & \scriptsize 0.044s & \scriptsize 0.054s & \scriptsize 0.438s & \scriptsize 0.497s & \scriptsize 0.167s & \scriptsize 0.216s \\
        \midrule
        \textbf{89 / 487} & $-75.82\pm1.81$ & $-72.40\pm1.92$ & $-87.58\pm1.36$ & $-54.87\pm2.29$ & $\mathbf{54.69}\pm0.94$ & $26.43\pm2.11$ \\
        & \scriptsize 0.061s & \scriptsize 0.062s & \scriptsize 0.117s & \scriptsize 0.120s & \scriptsize 0.062s & \scriptsize 0.054s \\
        \midrule
        \textbf{158 / 865} & $-71.17\pm1.94$ & $-63.52\pm2.14$ & $\mathbf{55.10}\pm0.95$ & $15.29\pm2.30$ & $ 54.09^{\dagger}\pm1.00 $ & $27.83\pm2.08$ \\
        & \scriptsize 0.095s & \scriptsize 0.100s & \scriptsize 0.331s & \scriptsize 0.263s & \scriptsize 0.115s & \scriptsize 0.096s \\
        \midrule
        \textbf{281 / 1539} & $-46.51\pm2.39$ & $-60.41\pm2.19$ & $\mathbf{55.30}\pm0.93$ & $41.64\pm1.66$ & $ 54.56^{\dagger}\pm0.98 $ & $31.17\pm2.00$ \\
        & \scriptsize 0.192s & \scriptsize 0.175s & \scriptsize 0.764s & \scriptsize 0.563s & \scriptsize 0.189s & \scriptsize 0.164s \\
        \midrule
        \textbf{500 / 2739} & $-42.19\pm2.43$ & $-57.66\pm2.25$ & $\mathbf{56.37}\pm0.86$ & $40.18\pm1.72$ & $ 55.18^{\dagger}\pm0.94 $ & $30.55\pm2.02$ \\
        & \scriptsize 0.360s & \scriptsize 0.347s & \scriptsize 1.636s & \scriptsize 1.109s & \scriptsize 0.348s & \scriptsize 0.297s \\
        \bottomrule
    \end{tabular}
\end{table*}

\begin{table}[h]
    \caption{Manual hyperparameters in the Lunar Lander domains.}
    \label{tab:lunar_lander_manual_hyperparameters}
    \centering
    \small
    \begin{tabular}{l c c}
        \toprule
        \textbf{Param.} & \textbf{MDP} & \textbf{POMDP} \\
        \midrule
        $J$             & 1 & 150 \\
        $J^{\text{PF}}$  & - & 2000 \\
        $K_{rollout}$   & 1 & 5 \\
        $K_{\text{opt}}$ & 3 & 3 \\
        $T_{d_a}^{\max}$ & 0.01 & 0.01 \\
        $T_{\rho}^{add}$ & 0.9 & 0.9 \\
        $T_{\rho}^{del}$ & 0.001 & \num{1e-8} \\
        $K_{b}^{\nabla}$ & - & 3 \\
        \bottomrule
    \end{tabular}
\end{table}

\begin{table}[h]
    \caption{Hyperparameters used in the evaluations of each algorithm in the Lunar Lander MDP domain.}
    \label{tab:lunar_lander_mdp_hyperparameters}
    \centering
    \small
    \begin{tabular}{l cc cc}
        \toprule
        & \multicolumn{2}{c}{\textbf{DPW Variants}} & \multicolumn{2}{c}{\textbf{AG Variants (Ours)}} \\
        \textbf{Param.} & DPW & VPW & AG-DPW & AG-VPW \\
        \midrule
        $c$             & 60.50 & 58.96 & 61.54 & 61.10 \\
        $k_a$           & 1.43 & 1.50 & 1.87 & 1.79 \\
        $\alpha_a$      & 0.59 & 0.58 & 0.51 & 0.54 \\
        $k_o$           & 0.07 & 0.08 & 0.07 & 0.11 \\
        $\alpha_o$      & 0.29 & 0.69 & 0.91 & 0.74 \\
        \midrule
        $\eta_{\text{Adam}}$ & - & - & \num{5.5e-2} & \num{2.1e-3} \\
        \bottomrule
    \end{tabular}
\end{table}

\begin{table*}[h]
    \caption{Hyperparameters used in the evaluations of each algorithm in the Lunar Lander POMDP domain.}
    \label{tab:lunar_lander_pomdp_hyperparameters}
    \centering
    \small
    \begin{tabular}{l cc cc cc}
        \toprule
        & \multicolumn{2}{c}{\textbf{DPW Variants}} & \multicolumn{2}{c}{\textbf{AG Variants (Ours)}} & \multicolumn{2}{c}{\textbf{POMCP Variants}} \\
        \textbf{Param.} & PFT-DPW & PFT-VPW & AG-PFT-DPW & AG-PFT-VPW & POMCPOW & VOMCPOW \\
        \midrule
        $c$             & 60.87 & 57.80 & 50.99 & 50.19 & 71.02 & 74.80 \\
        $k_a$           & 3.67 & 2.81 & 2.41 & 3.47 & 2.69 & 3.05 \\
        $\alpha_a$      & 0.39 & 0.45 & 0.49 & 0.41 & 0.46 & 0.43 \\
        $k_o$           & 0.24 & 0.22 & 0.28 & 0.28 & 0.57 & 0.54 \\
        $\alpha_o$      & 0.65 & 0.59 & 0.49 & 0.80 & 0.28 & 0.29 \\
        \midrule
        $\eta_{\text{Adam}}$ & - & - & \num{1.2e-7} & \num{3.7e-7} & - & - \\
        \bottomrule
    \end{tabular}
\end{table*}

\begin{table*}[h]
    \caption{
        Performance (Mean $\pm$ SEM) and Runtimes in Lunar Lander MDP.
    }
    \label{tab:lunar_lander_mdp_complete}
    \centering
    \small
    \begin{tabular}{l cc cc}
        \toprule
        \textbf{Simulations} & \multicolumn{2}{c}{\textbf{DPW Variants}} & \multicolumn{2}{c}{\textbf{AG Variants (Ours)}} \\
        $n^{\text{sims}}$ & DPW & VPW & AG-DPW & AG-VPW \\
        \midrule
        \textbf{100} & $ 44.40^{\dagger}\pm1.03 $ & $39.23\pm1.97$ & $ 44.22^{\dagger}\pm1.41 $ & $\mathbf{44.63}\pm0.55$ \\
        & \scriptsize 0.019s & \scriptsize 0.083s & \scriptsize 0.022s & \scriptsize 0.086s \\
        \midrule
        \textbf{178} & $ 50.24^{\dagger}\pm0.99 $ & $ 47.46^{\dagger}\pm1.35 $ & $\mathbf{50.69}\pm0.48$ & $ 48.79^{\dagger}\pm1.00 $ \\
        & \scriptsize 0.002s & \scriptsize 0.002s & \scriptsize 0.004s & \scriptsize 0.203s \\
        \midrule
        \textbf{316} & $\mathbf{55.80}\pm0.41$ & $ 54.46^{\dagger}\pm0.41 $ & $ 54.00^{\dagger}\pm1.05 $ & $ 54.49^{\dagger}\pm0.42 $ \\
        & \scriptsize 0.008s & \scriptsize 0.007s & \scriptsize 0.239s & \scriptsize 0.017s \\
        \midrule
        \textbf{562} & $\mathbf{59.80}\pm0.37$ & $ 58.45^{\dagger}\pm0.35 $ & $ 59.31^{\dagger}\pm0.39 $ & $57.56\pm0.45$ \\
        & \scriptsize 0.005s & \scriptsize 0.005s & \scriptsize 0.046s & \scriptsize 0.059s \\
        \midrule
        \textbf{1000} & $\mathbf{63.20}\pm0.29$ & $60.94\pm0.34$ & $61.28\pm0.40$ & $60.28\pm0.42$ \\
        & \scriptsize 0.008s & \scriptsize 0.009s & \scriptsize 0.203s & \scriptsize 0.216s \\
        \bottomrule
    \end{tabular}
\end{table*}

\begin{table*}[h]
    \caption{
        Performance (Mean $\pm$ SEM) and Runtimes in Lunar Lander POMDP.
    }
    \label{tab:lunar_lander_pomdp_complete}
    \centering
    \small
    \begin{tabular}{l cc cc cc}
        \toprule
        \textbf{Simulations} & \multicolumn{2}{c}{\textbf{DPW Variants}} & \multicolumn{2}{c}{\textbf{AG Variants (Ours)}} & \multicolumn{2}{c}{\textbf{POMCP Variants}} \\
        $n^{\text{sims}}$ / $n_{\text{POMCPOW}}^{\text{sims}}$ & PFT-DPW & PFT-VPW & AG-PFT-DPW & AG-PFT-VPW & POMCPOW & VOMCPOW \\
        \midrule
        \textbf{100 / 875} & $21.99\pm4.67$ & $12.59\pm5.40$ & $20.93\pm4.90$ & $22.11\pm4.88$ & $\mathbf{42.13}\pm3.84$ & $ 41.96^{\dagger}\pm3.81 $ \\
        & \scriptsize 0.015s & \scriptsize 0.020s & \scriptsize 0.307s & \scriptsize 0.343s & \scriptsize 0.088s & \scriptsize 0.145s \\
        \midrule
        \textbf{178 / 1557} & $25.03\pm4.70$ & $ 30.36^{\dagger}\pm4.23 $ & $ 39.39^{\dagger}\pm3.48 $ & $ 37.07^{\dagger}\pm3.77 $ & $\mathbf{45.50}\pm3.82$ & $ 44.35^{\dagger}\pm3.86 $ \\
        & \scriptsize 0.018s & \scriptsize 0.019s & \scriptsize 0.043s & \scriptsize 0.041s & \scriptsize 0.025s & \scriptsize 0.028s \\
        \midrule
        \textbf{316 / 2764} & $36.11\pm3.78$ & $ 41.24^{\dagger}\pm3.31 $ & $ 45.49^{\dagger}\pm3.26 $ & $ 44.51^{\dagger}\pm3.13 $ & $\mathbf{51.65}\pm3.09$ & $ 49.52^{\dagger}\pm3.40 $ \\
        & \scriptsize 0.031s & \scriptsize 0.031s & \scriptsize 0.081s & \scriptsize 0.078s & \scriptsize 0.037s & \scriptsize 0.037s \\
        \midrule
        \textbf{562 / 4916} & $ 46.23^{\dagger}\pm2.75 $ & $41.27\pm3.59$ & $ 51.69^{\dagger}\pm2.63 $ & $ 47.34^{\dagger}\pm3.16 $ & $ 52.67^{\dagger}\pm3.15 $ & $\mathbf{56.09}\pm2.68$ \\
        & \scriptsize 0.053s & \scriptsize 0.053s & \scriptsize 0.184s & \scriptsize 0.186s & \scriptsize 0.075s & \scriptsize 0.081s \\
        \midrule
        \textbf{1000 / 8748} & $ 44.72^{\dagger}\pm3.33 $ & $ 44.05^{\dagger}\pm3.45 $ & $ 51.47^{\dagger}\pm2.97 $ & $ 52.00^{\dagger}\pm2.77 $ & $\mathbf{54.89}\pm2.96$ & $ 53.67^{\dagger}\pm3.05 $ \\
        & \scriptsize 0.103s & \scriptsize 0.102s & \scriptsize 0.504s & \scriptsize 0.508s & \scriptsize 0.136s & \scriptsize 0.144s \\
        \bottomrule
    \end{tabular}
\end{table*}

\subsection{Experimental Method}

In a high-level overview, we conducted the experiments for each domain in the following manner:
\begin{enumerate}
    \item Constant hyperparameters were defined for each algorithm, including maximum simulation budget $n_{\max}^{\textnormal{sims}}$, particle count $J$ (in POMDPs), and rollout policy.
    \item A cross-entropy (CE) optimization was performed over the UCB and DPW parameters for each algorithm, for 50 iterations. For AGMCTS, the hyperparameters were optimized jointly with Adam's step size parameter $\eta_{\text{Adam}}$.
    \item For the best performing hyperparameters of each algorithm of the 50 CE iterations, we ran 1000 simulations of the algorithm with the same seeds between algorithms, at the simulation budgets $n^{\textnormal{sims}}=[10^{-1}, 10^{-0.75}, 10^{-0.5}, 10^{-0.25}, 1.0]\cdot n_{\max}^{\textnormal{sims}}$, where $n_{\max}^{\textnormal{sims}}$ is the maximum simulation budget allowed for the domain.
\end{enumerate}

The simulation jobs were run on a distributed CPU cluster, using the Julia programming language with Distributed.jl multi-processing tools.
Each simulation was run on a single core, without any parallelization in the solver itself.
The cluster's machines feature the following processors:
\begin{enumerate}
    \item Intel Xeon Gold 6230 Processor {--} up to 3.9 GHz clock speed.
    \item Intel Xeon Platinum 8358 Processor {--} up to 3.4 GHz clock speed.
    \item AMD EPYC 9654 Processor {--} up to 3.7 GHz clock speed.
\end{enumerate}

The implementation of AGMCTS was written in the POMDPs.jl~\cite{egorov2017pomdps} framework.
We took the double progressive widening (DPW)~\cite{Couetoux11iclio} implementation from the MCTS.jl~\cite{JuliaPOMDP_MCTSjl} package, particle belief implementations from the ParticleFilters.jl~\cite{JuliaPOMDP_ParticleFiltersjl} package, and the POMCPOW~\cite{Sunberg18icaps} implementation from the POMCPOW.jl~\cite{JuliaPOMDP_POMCPOWjl} package.
The implementation of Voronoi progressive widening (VPW)~\cite{Lim21cdc} was adapted from the codebase of the original authors.

The following algorithms were compared in the MDP domains:
\begin{itemize}
    \item \textbf{DPW:} MCTS with Double Progressive Widening (DPW), where the action sampling is uniform via Action Progressive Widening (APW).
    \item \textbf{VPW:} MCTS with Voronoi Progressive Widening (VPW), where the action sampling is via Voronoi Optimistic Optimization (VOO).
    \item \textbf{AG-DPW:} AGMCTS with DPW and uniform action sampling via APW.
    \item \textbf{AG-VPW:} AGMCTS with VPW and action sampling via VOO.
\end{itemize}

The following algorithms were compared in the POMDP domains:
\begin{itemize}
    \item \textbf{PFT-DPW:} Particle Filter Tree (PFT) with Double Progressive Widening (DPW), where the action sampling is uniform via Action Progressive Widening (APW).
    \item \textbf{PFT-VPW:} Particle Filter Tree (PFT) with Voronoi Progressive Widening (VPW), where the action sampling is via Voronoi Optimistic Optimization (VOO).
    \item \textbf{AG-PFT-DPW:} AGMCTS with DPW and uniform action sampling via APW.
    \item \textbf{AG-PFT-VPW:} AGMCTS with VPW and action sampling via VOO.
    \item \textbf{POMCPOW:} Partially Observable Monte Carlo Planning with Observation Widening (POMCPOW), using uniform action sampling via APW.
    \item \textbf{VOMCPOW:} Variant of POMCPOW using action sampling via VOO.
\end{itemize}

For the action-gradient algorithms, automatic differentiation was mostly done via Enzyme.jl~\cite{Moses20nips}, and we used ForwardDiff.jl~\cite{Revels16arxiv} overloaded with SciMLSensitivity.jl for ODE solution sensitivity~\cite{Rackauckas20arxiv}.

The following reporting conventions apply to all scenarios and results presented in this Appendix. Algorithm-specific hyperparameters and their notations are summarized in Table~\ref{tab:hyperparameters_notations}, while domain-specific values are listed in the respective tables. 
Reported numbers are rounded to two significant digits. 
Tabulated performance metrics reflect $\pm 1$ standard error (SEM). The highest mean performance in each row is highlighted in \textbf{bold}, and a dagger ($\dagger$) indicates results with confidence intervals ($\pm 2$ SEM) overlapping with the best result.
The plots in Figure \ref{fig:mdp_pomdp_grid_results} depict $\pm 2$ standard deviations. 
Runtimes (in seconds) are indicated in smaller font below the performance values; we note that these runtimes are artificially high for small simulation counts due to Julia's pre-compilation overhead.

\subsubsection{Simulation Budget}

Since POMCPOW is a state-trajectory algorithm, it is incomparable in terms of simulation budgets to belief-trajectory algorithms like PFT-DPW and AGMCTS.
In the POMDP scenarios, we measured empirically the runtime of PFT-DPW at $n_{\max}^{\textnormal{sims}}$ and $J$, and set a larger simulation budget for POMCPOW $n_{\max,\text{POMCPOW}}^{\textnormal{sims}}$ to match the runtime of PFT-DPW.

\subsubsection{Rollout Budget}

In order to save computation time and not compute a partially observable rollout, the rollout computation for a new particle belief was done by drawing $K_{rollout}$ particles from the belief, and computing the rollout policy based on the mean state, while calculating the rollout estimate based on the mean of the returns of the $K_{rollout}$ particles.
For POMCPOW, since every new node is expanded initially only with a single particle, the rollout was computed with that particle.
This is a disadvantage for POMCPOW as it made the rollout returns less meaningful, however it seemed to have achieved consistently better results than PFT-DPW despite that.
In the MDP domains, we have set $K_{rollout}=1$ similarly to POMCPOW.

\subsubsection{Cross-Entropy (CE) Optimization}

The algorithm hyperparameters were first determined by cross entropy optimization (CE)~\cite{Rubinstein04book}, largely following~\cite{Botev13chapter}, using Gaussian distributions over the continuous DPW hyperparameters:
\begin{enumerate}
    \item UCT exploration bonus $c$.
    \item Action progressive widening parameters $k_a$ and $\alpha_a$.
    \item Observation progressive widening parameters $k_o$ and $\alpha_o$.
\end{enumerate}
For AGMCTS, we also optimized over the Adam learning rate $\eta_{\text{Adam}}$. To ensure that the obtained parameters result in algorithms that do not cancel out the action gradients, we constrained the CE search space to the following condition:
\begin{linenomath*}
    \begin{align*}
        \begin{multlined}[b][0.8\columnwidth]
            \eta_{\text{Adam}}\cdot K_{\text{opt}}\left(N-\text{ceil}\left(\left(\frac{K_{\text{child}}^{\min}}{k_{o}}\right)^{1/\alpha_{o}}\right)\right) 
            \geq 0.01\cdot T_{d_{a}}^{\max}
        \end{multlined}
    \end{align*}
\end{linenomath*}
This condition states that the number of action gradient optimization steps after $K^{\text{child}}_{\min}$, assuming all of the simulation budget has been put into a single action branch at the root, multiplied by the step size $\eta_{\text{Adam}}$, should be sufficient to reach at least $1\%$ of the maximum allowed action update distance $T_{d_{a}}^{\max}$.
CE parameter samples that did not satisfy this condition were rejected.
In the domains in which AGMCTS algorithms obtained worse performance than their non-gradient counterparts, we observed that cancelling this condition led to convergence to hyperparameters that effectively disabled the action gradients, and the performance matched that of the DPW/VPW counterpart algorithm.
This suggests that in these domains, the action gradients were not beneficial to the planning performance.

We used 150 parameter samples, 40 simulations for each parameter sample to determine its mean return, used 30 elite samples to fit the new distribution, and smoothing the new distribution's parameters:
\begin{linenomath*}
    \begin{gather*}
        \mu_{\text{new}} = \alpha_\mu \mu_{\text{new}} + (1-\alpha_\mu)\mu_{\text{old}}, \\
        \Sigma_{\text{new}} = \alpha_\Sigma \Sigma_{\text{new}} + (1-\alpha_\Sigma)\Sigma_{\text{old}},
    \end{gather*}
\end{linenomath*}
where we chose $\alpha_\mu=0.8, \alpha_\Sigma=0.5$.

We performed 50 CE iterations for each algorithm in each dimension, with reasonable initial guesses for the starting parameters based on the literature and manual experiments.
The hyperparameters chosen were those from the CE iteration achieving the maximal mean of mean returns over the 30 elite samples.
After choosing the parameters, we reported the mean return over 1000 scenarios, with $\pm2$ standard errors around the mean.

\subsection{$d$D-Continuous Light-Dark POMDP.}

\begin{figure}[t]
    \centering
    \makebox[0.9\columnwidth][c]{
    \begin{subfigure}[b]{0.48\columnwidth}
    \centering
    \includegraphics[trim={1.0cm 0.2cm 2cm 0.2cm},clip,width=1.0\columnwidth]{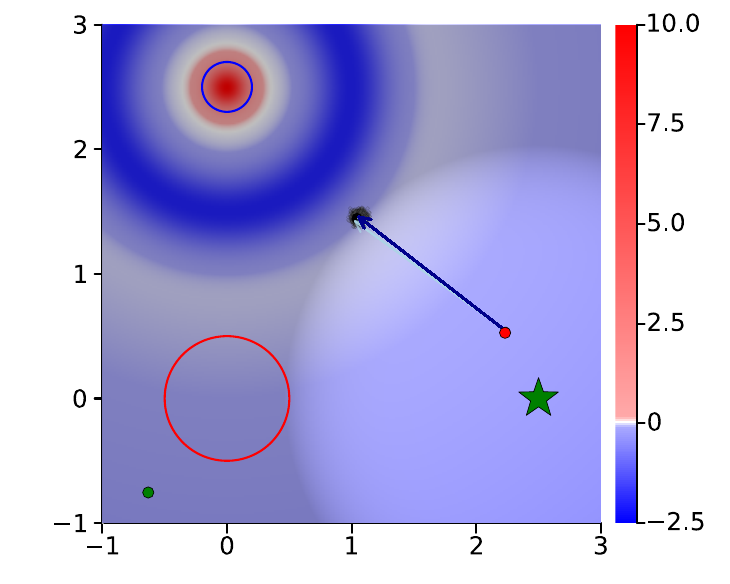}
    \caption{Middle of scenario, $t=5$.}
    \label{fig:sub1}
    \end{subfigure}
    \hfill
    \begin{subfigure}[b]{0.48\columnwidth}
    \centering
    \includegraphics[trim={2.7cm 0.2cm 0.3cm 0.2cm},clip,width=1.0\columnwidth]{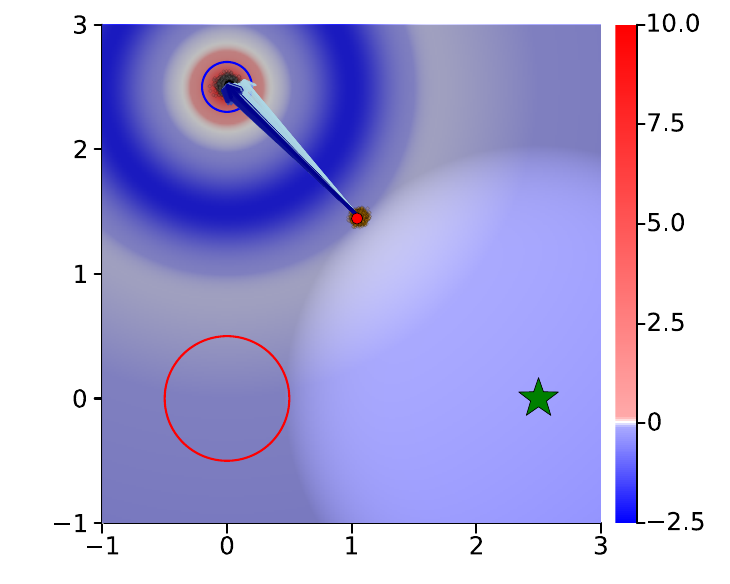}
    \caption{End of scenario, $t=6$.}
    \label{fig:sub2}
    \end{subfigure}
    }
    \caption
    {
    AGMCTS at 2D-Continuous Light-Dark. The agent's current state is the red dot, the current belief particles in orange, the next state in black, and the next belief particles in gray, the next observation is the green dot.
    The goal is the blue ring centered at $(0,2.5)$, $b_0$ is the red ring centered at $(0,0)$, and the beacon is the green star at $(2.5,0)$. 
    The reward function for the posterior state is the blue-red heatmap drawn in the background. The shaded area is where the observation noise is $\sigma_{O}>5$.
    The chosen action branch of AGMCTS is drawn as the blue arrows, getting darker in hue with each action update.
    We can see that the optimized actions point closer to the goal from the current belief's mean.
    }
    \label{fig:mis_update_zoom}
\end{figure}

\subsubsection{Problem Description}

In this domain, $\mathcal{S}=\mathcal{O}=\mathbb{R}^d$. The agent starts at a random position on the sphere of radius $r_0$ centered at the origin: $S^{d-1}_{r_0}$. The goal position is located at $s_{g}=(\boldsymbol{0}_{d-1},r_{g})$. A beacon is located to the side at $s_{b}=(r_{b},\boldsymbol{0}_{d-1})$. The constants we took are $r_a=1.5$, $r_g=r_b=2.5$, $r_0=0.5$. $T_{\textnormal{goal}}=0.2$ is the goal tolerance, and the POMDP terminates if $\normflat{s-s_{g}}<T_{\textnormal{goal}}$, or after $L=6$ time steps. We set $\gamma=0.99$.

The action space is $B_{r_a}(\boldsymbol{0}_{d})$. The transition model is a simple Gaussian with added noise: $s^\prime \sim \mathcal{N}(s+a,\sigma_{T}^2\mathbf{I})$ with $\sigma_{T}=0.025$. 

The observations are the agent's relative position to the beacon, with a Gaussian added noise that indreases with distance: $o\sim\mathcal{N}(s-s_b,(\sigma_{O}(\normflat{s-s_b}))^2\mathbf{I})$, with noise function $\sigma_{O}(x)=\min\{\sigma_{O}^{\max},k_{\sigma_{O}}(x+x^{\alpha_{\sigma_{O}}})\}$, for $\sigma_{O}^{\max}=15$, $k_{\sigma_{O}}=0.01$, $\alpha_{\sigma_{O}}=8$.

The reward function is $r(s,a,s^{\prime})=R_{goal}\exp(-\frac{1}{2}{(\frac{d}{0.5T})}^2)-R_{moat}\exp(-\frac{1}{2}{(\frac{d-5T}{T})}^2)-R_{dist}d^2$, where $d\mathbin{=}\normflat{s-s_{g}}$.
We took constants $R_{goal}\mathbin{=}10$, $R_{moat}\mathbin{=}2$, $R_{dist}\mathbin{=}0.02$.
This reward function only depends on the posterior state. The agent needs to carefully target the goal as the reward decreases quickly from the center, and advancing at random towards the goal will incur a penalty due to the moat.

To simulate more difficult domains, we include a noisy rollout policy.
The rollout heads directly to $s_g$, but the computed action has an added noise of $\mathcal{N}(\mathbf{0},\sigma_{r}^2\mathbf{I})$. In our experiments we set $\sigma_{r}=0.1$.

The implementation of $d$D-Continuous Light-Dark was written using POMDPs.jl~\cite{egorov2017pomdps}.
In this domain, it is straightforward to compute the transition density:
\begin{linenomath*}
    \begin{gather*}
        p_T(s^\prime{\mid} s,a) = \frac{1}{(2\pi\sigma_{T}^2)^{d/2}}\exp\left(-\frac{\normflat{s^\prime-(s+a)}^2}{2\sigma_{T}^2}\right).
    \end{gather*}
\end{linenomath*}

The gradient function $\nabla_a r$ was equal to $0$ in this domain, and $\nabla_a \log p_T$ was calculated using Enzyme.jl~\cite{Moses20nips}.
Enzyme.jl provided efficient automatic differentiation, yielding log-gradient computation times close to the transition probability.

\subsubsection{Evaluation}

We evaluated the algorithms on the $d$D-Continuous Light-Dark domain for $d=2,3,4$ dimensions.
PFT-DPW and AGMCTS were given a budget of $n_{\max}^{\textnormal{sims}}=500$ simulations. 
The particle count during planning was set to be dependent on the dimension:
\begin{linenomath*}
    \begin{gather*}
        J_{d=2}=64,\quad J_{d=3}=128, \quad
        J_{d=4}=256. \label{eq:light_dark_particle_count_planner}
    \end{gather*}
\end{linenomath*}
The particle filter used for inference between planning steps had a particle count $J^{\text{PF}}$ given by:
\begin{linenomath*}
    \begin{gather*}
        J^{\text{PF}}_{d=2}=256,\quad J^{\text{PF}}_{d=3}=512,  \quad
        J^{\text{PF}}_{d=4}=1024. \label{eq:light_dark_particle_count_inference}
    \end{gather*}
\end{linenomath*}
POMCPOW was given a simulation budget according to the formula
\begin{linenomath*}
    \begin{gather*}
        n_{\text{POMCPOW},d}^{\textnormal{sims}} = n_{\max}^{\textnormal{sims}} \cdot \sqrt{J_{d}}. \label{eq:light_dark_pomcpow_simulation_budget}
    \end{gather*}
\end{linenomath*}

Additional parameters that were manually set in all dimensions were:
\begin{linenomath*}
    \begin{gather*}
        T_{\rho}^{add} = 0.9,\quad T_{\rho}^{del} = \num{1e-8}, \quad K_{rollout}=10, 
        \\
        K_{\text{opt}}=3, \quad T_{d_a}^{\max}=0.05 \cdot \sigma_{T}, \quad K_{b}^{\nabla}=4 \\
        \omega_{\textnormal{VOO}} = 0.85, \quad \Sigma_{\textnormal{VOO}} = 0.05\cdot \boldsymbol{I}.
    \end{gather*}
\end{linenomath*}
The hyperparameters found by CE optimization for each algorithm in each dimension are summarized in Tables~\ref{tab:2d_light_dark_hyperparameters_reformatted}, \ref{tab:3d_light_dark_hyperparameters} and \ref{tab:4d_light_dark_hyperparameters}.
In this domain, exponential step size decay was used.
The importance weight updates were linearized according to Equation \eqref{eq:imp_ratio_linear_update}, along with linearized weight-updates with state-reward action-gradient estimates via Equation \eqref{eq:pomdp_state_grad_est_linearized_rpt}.
Performance results and runtimes are reported in Tables~\ref{tab:2d_light_dark_complete}, \ref{tab:3d_light_dark_complete} and \ref{tab:4d_light_dark_complete}.

\subsection{Two-Agent $d$D-Continuous Light-Dark POMDP.}

\subsubsection{Problem Description}
This domain is a multi-agent extension of the $d$D-Continuous Light-Dark domain described previously, and we specifically evaluate in $d=2$ dimensions due to computational constraints.
The state space and observation space are concatenated vectors of the individual agents, such that $\mathcal{S}=\mathcal{O}=\mathbb{R}^{2d}$, and the action space is the product of balls $\mathcal{A}=B_{r_a}(\boldsymbol{0}_N)\times B_{r_a}(\boldsymbol{0}_N)$.
The transition and reward models follow the same logic as the single-agent case, averaged over the number of agents. 
The transition dynamics are independent for each agent $i \in \{1, 2\}$, following $x^{(i)\prime} \sim \mathcal{N}(x^{(i)}+a^{(i)},\sigma_{T}^2\mathbf{I})$. 
The collective reward is the average of the individual rewards obtained by each agent, with the exception that the reward goal is only obtained for all agents if they are all within the goal threshold distance $T_{\textnormal{goal}}=0.2$.

The primary challenge in this domain lies in the coupled observation model. 
While the first agent observes the fixed beacon $x_b$, the second agent does not observe the beacon directly; instead, it observes its position relative to the first agent.
Let $x = [x^{(1)}, x^{(2)}]^T$. The observation vector $z = [z^{(1)}, z^{(2)}]^T$ is distributed as:
\begin{linenomath*}
    \begin{align*}
        z^{(1)} &\sim \mathcal{N}(x^{(1)} - x_b, \Sigma(x^{(1)}, x_b)) \\
        z^{(2)} &\sim \mathcal{N}(x^{(2)} - x^{(1)}, \Sigma(x^{(2)}, x^{(1)}))
    \end{align*}
\end{linenomath*}
where the covariance $\Sigma(u, v) = (\sigma_Z(\normflat{u-v}))^2\mathbf{I}$ uses the same distance-dependent noise function $\sigma_Z$ as the single-agent domain.
This structure necessitates collaboration: Agent 1 must localize itself relative to the beacon to become a reliable reference point for Agent 2, effectively acting as a mobile beacon.

\subsubsection{Evaluation}

We evaluated the algorithms for $d=2$ dimensions. All manually-chosen hyperparameters and particle counts followed the same as the single-agent experiments. 
The hyperparameters found via CE optimization for the collaborative setting are listed in Table~\ref{tab:collab_light_dark_hyperparameters}. 
Notably, AGMCTS required a significantly higher learning rate in this domain compared to the single-agent version.
Further hyperparameters are reported in Table~\ref{tab:collab_light_dark_hyperparameters}. Performance results and runtimes are reported in Table~\ref{tab:collab_light_dark_complete}.

\subsection{Mountain Car and Hill Car}

\subsubsection{Problem Description}

A car is placed randomly in a valley between two hills, and the goal is to reach the top of the right hill.
The car's action is its acceleration left or right, and must gain momentum in order to reach the goal.
In this domain, the state $s=(x,v)\in\mathbb{R}^2$ is the car's position and velocity, $\mathcal{A}=[-a_{\max},a_{\max}]\subset \mathbb{R}$ is the action space.

In the POMDP version, $z\in\mathbb{R}$ is the observation of the car's position with noise.
It is given by $z\sim\mathcal{N}(x,\sigma_{O}^2)$, where $\sigma_{O}=0.03$.

The car's position is bounded to $[x_{\min}, x_{\max}]$, and the velocity is bounded to $[-v_{\max}, v_{\max}]$.

If the car reaches the goal $x\geq x_{\max}$, it receives a reward of $100$. If $x<x_{\min}$, or $\absval{v} \geq v_{\max}$, a penalty reward of $-100$ is attained and the scenario terminates. Otherwise, the car obtains a reward of $-0.1$ at each time step that does not terminate.
The reward function is a function of the posterior state, i.e. $r(s,a,s^\prime)=r(s^\prime)$.
In Mountain Car, the scenario terminates after $L=200$ time steps, while in Hill Car it terminates after $L=30$ time steps.
The discount in both domains is $\gamma=0.99$.

The rollout policy chooses $a_{\max}$ when $v>0$, and $-a_{\max}$ otherwise.

In both domains, the generative transition model is given by a noise applied to the action at the input to the deterministic dynamics:
\begin{linenomath*}
    \begin{align*}
        \xi &\sim \mathcal{N}(0, \sigma_{\xi}^2), \\
        \tilde{a} &= \text{clip}(a + \xi, -a_{\max}, a_{\max}), \\
        s^\prime &= f(s, \tilde{a}, s^\prime),
    \end{align*}
\end{linenomath*}
where $\sigma_{\xi}=0.1$ for both domains.

When calculating the transition density in this domain, we do it based on the input noise to a deterministic simulator case via the Area Formula.
For each $(s,a,s^\prime)$ tuple we calculate the generating action noise $\xi$, depending on the scenario.
If the obtained $\xi$ is outside of the bounds (beyond numerical errors) $[L,R]$ where $L=-a_{\max} - a$ and $R=a_{\max} - a$, then the density is $0$. Otherwise, because of the clip operation, we calculate its density by the following mixture:
\begin{linenomath*}
    \begin{equation*}
        p(\xi{\mid} s,a) = \begin{cases}
            CDF(\mathcal{N}(0,\sigma_{\xi}^2), L), & \text{if } \xi \leq L \\
            1 - CDF(\mathcal{N}(0,\sigma_{\xi}^2), R), & \text{if } \xi \geq R \\
            PDF(\mathcal{N}(0,\sigma_{\xi}^2), \xi), & \text{otherwise}.
          \end{cases}
    \end{equation*}
\end{linenomath*}
Afterwards, we calculate the transition model density by the Area Formula according to the case of \emph{Input Noise to Simulator}.

\subsubsection{Mountain Car Transition Model}

In this domain, we set $x_{\min}=-1.5$, $x_{\max}=0.5$, $v_{\max}=0.05$, $a_{\max}=1.0$.
The deterministic dynamics are calculated by:
\begin{linenomath*}
    \begin{gather*}
        v^\prime = v + 0.001\cdot\tilde{a} - 0.0025\cos(3x), \\
        x^\prime = x + v^\prime.
    \end{gather*}
\end{linenomath*}
In order to calculate the transition density, for given $s,a,s^\prime$ we invert the dynamics to obtain $\xi$.

\subsubsection{Hill Car Transition Model}

In this domain, we set $x_{\min}=-1.0$, $x_{\max}=1.0$, $v_{\max}=2.5$, $a_{\max}=4.0$.
The deterministic dynamics are calculated by integrating the car's continuous-time dynamics with an ODE solver over a time period of $\Delta t = 0.1$.
The equations were integrated using the adaptive Tsitouras 5/4 Runge-Kutta method (Tsit5) \cite{Tsitouras11cma}, with a manually specified initial step size of $\delta t = 0.01$.

The continuous-time dynamics are given by:
\begin{linenomath*}
    \begin{align*}
        \dot{v} &= \left(\frac{a}{m} - g \cdot h^\prime(x) - v^2 \cdot h^\prime(x) \cdot h^{\prime\prime}(x)\right)\frac{1}{1 + h^\prime(x)^2}, \\
        \dot{x} &= v,
    \end{align*}
\end{linenomath*}
where $m=1$, $g=9.81$, and the hill-curve function $h(x)$ is given by:
\begin{linenomath*}
    \begin{align*}
        h(x) = \begin{cases}
            x^2 + x, & \text{if } x < 0 \\
            \frac{x}{\sqrt{1.0 + 5.0 * x^2}}, & \text{otherwise}.
          \end{cases}
    \end{align*}
\end{linenomath*}

Because the input noise is only based on the action, we calculate $\xi$ by caching simulator outputs $(s,a,\xi,s^\prime)$ tuples for which $s^\prime = f(s,\tilde{a})$.
For a new tuple $(s,a,s^\prime)$, we calculate $\xi^\prime=\tilde{a}-a$ for the cached $\tilde{a}$.
The jacobian of the dynamics is calculated by ForwardDiff.jl and SciMLSensitivity.jl overloaded gradient function for the ODE solver.

\subsubsection{Evaluation}

DPW, PFT-DPW and AGMCTS were given a simulation budget of $n_{\max}^{\textnormal{sims}}=500$ simulations.
POMCPOW simulation budget was set by the formula:
\begin{linenomath*}
    \begin{gather*}
        n_{pomcpow}^{\textnormal{sims}} = n_{\max}^{\textnormal{sims}} \cdot \num{0.08} \cdot J.
    \end{gather*}
\end{linenomath*}
For the VPW algorithms, we have used the following VOO hyperparameters:
\begin{linenomath*}
    \begin{gather*}
        \omega_{\textnormal{VOO}} = 0.85, \quad \Sigma_{\textnormal{VOO}} = 0.05.
    \end{gather*}
\end{linenomath*}
In these domains, we have not used exponential step size decay.
In all of the Mountain/Hill car domains, the importance weight updates were linearized according to Equation \eqref{eq:imp_ratio_linear_update}.
We used linearized weight-updates with state-reward action-gradient estimates via Equation \eqref{eq:mdp_state_grad_est_linearized_rpt} for the MDPs and Equation \eqref{eq:pomdp_state_grad_est_linearized_rpt} for the POMDPs.

Further hyperparameters are reported in Tables~\ref{tab:mountain_hill_manual_hyperparameters}, \ref{tab:mountain_car_mdp_hyperparameters}, \ref{tab:mountain_car_pomdp_hyperparameters}, \ref{tab:hill_car_mdp_hyperparameters}, and \ref{tab:hill_car_pomdp_hyperparameters}. Performance results and runtimes are reported in Tables~\ref{tab:mountain_car_mdp_complete}, \ref{tab:mountain_car_pomdp_complete}, \ref{tab:hill_car_mdp_complete}, and \ref{tab:hill_car_pomdp_complete}.

\subsection{Lunar Lander}

\subsubsection{Problem Description}

We used the Julia implementation provided by~\citeauthor{Mern21aaai}~[\citeyear{Mern21aaai}].

In this domain, a spacecraft vehicle must land safely on a flat surface from a high altitude.
The vehicle state is represented by a six dimensional tuple $(x,y,\theta,\dot{x},\dot{y},\omega)$, where $x$ and $y$ are the horizontal and vertical positions, $\theta$ is the orientation angle, $\dot{x}$ and $\dot{y}$ are the horizontal and vertical speeds, and $\omega$ is the angular rate.
The action space $\mathcal{A}\subset \mathbb{R}^3$ is a three-dimensional continuous space defined by the tuple $(F_x,T,\delta) \in [0,15]\times[-5,5]\times[-1,1]$. $T$ is the main thrust which acts along the vehicle's vertical axis through its center of mass. $F_x$ is the corrective thrust, which acts along a horizontal axis offset from the center of mass by a distance $\delta$.

The transition model is computed as an additive Gaussian noise to the output of a deterministic transition function.
The deterministic transition function is given by the following set of computations:
\begin{linenomath*}
    \begin{align*}
        f_x &= \cos(\theta) \cdot F_x - \sin(\theta) \cdot T, \\
        f_z &= \cos(\theta) \cdot T + \sin(\theta) \cdot F_x, \\
        \tau &= -\delta \cdot F_x, \\
        a_x &= \frac{f_x}{m},\quad a_z = \frac{f_z}{m},\quad \dot{\omega} = \frac{\tau}{I}, \\
        v_x' &= v_x + a_x \cdot \Delta t + \epsilon_1 \cdot Q_4, \\
        v_z' &= v_z + (a_z - 9.0) \cdot \Delta t + \epsilon_2 \cdot Q_5, \\
        \omega' &= \omega + \dot{\omega} \cdot \Delta t + \epsilon_3 \cdot Q_6, \\
        x' &= x + v_x \cdot \Delta t,\quad z' = z + v_z \cdot \Delta t,\quad \theta' = \theta + \omega \cdot \Delta t, \\
        s^{\prime} &= \begin{bmatrix} x' & z' & \theta' & v_x' & v_z' & \omega' \end{bmatrix}^T,
        \end{align*}
\end{linenomath*}
where the noise variables $\epsilon_i$ are sampled from a standard normal distribution, $Q_4=0.1$, $Q_5=0.1$, $Q_6=0.01$. We took the values $m=1$, $I=10$, and $\delta t=0.4$.
For computing the transition density, we used the Area Formula by solving the dynamics for the noise variables $\epsilon_i$ for given $(s,a,s^\prime)$ tuples.

In the POMDP variant, the vehicle makes noisy observations of its angular rate, horizontal speed, and above-ground height as measured by a distance sensor: $o=(\tilde{\omega}, \tilde{v_x}, \tilde{h})$, where $\tilde{\omega} \sim \mathcal{N}(\omega, 1.0^2)$, $\tilde{v_x} \sim \mathcal{N}(v_x, 0.01^2)$, and $\tilde{h} \sim \mathcal{N}(z / \cos(\theta), 0.1^2)$.

The reward function is defined as:
\begin{linenomath*}
    \begin{align*}
        r(s,a,s^\prime) = \begin{cases}
            -1000, & \text{if } x\geq 15 \vee \theta \geq 0.5 \\
            100 - x - \dot{y}^2, & \text{if } y \leq 1 \\
            -1, & \text{otherwise}.
        \end{cases}
    \end{align*}
\end{linenomath*}
The discount factor is set to $\gamma=0.99$.
The evaluation horizon is $L=35$ steps.

The initial vehicle state is sampled from a multivariate Gaussian with mean $\mu=(x=0,y=50,\theta=0,\dot{x}=0,\dot{y}=-10,\omega=0)$. The rollout policy was computed as a simple proportional rule: $a_{rollout}=(-0.1\cdot\dot{x},-0.1\cdot\dot{y},0)$.

\subsubsection{Evaluation}

DPW, PFT-DPW and AGMCTS were given a simulation budget of $n_{\max}^{\textnormal{sims}}=1000$ simulations.
POMCPOW simulation budget was set by the formula:
\begin{linenomath*}
    \begin{align*}
        n_{pomcpow}^{\textnormal{sims}} = n_{\max}^{\textnormal{sims}} \cdot \num{0.035} \cdot J.
    \end{align*}
\end{linenomath*}
For the VPW algorithms, we have used the following VOO hyperparameters:
\begin{linenomath*}
    \begin{gather*}
        \omega_{\textnormal{VOO}} = 0.9, \quad \Sigma_{\textnormal{VOO}} = \text{diag}(0.2, 0.5, 0.05).
\end{gather*}
\end{linenomath*}
In these domains, we have not used exponential step size decay.
In all of the Lunar Lander domains, the importance weight updates were linearized according to Equation \eqref{eq:imp_ratio_linear_update}.
We used linearized weight-updates with state-reward action-gradient estimates, as in Equation \eqref{eq:mdp_state_grad_est_linearized_rpt} for the MDPs and Equation \eqref{eq:pomdp_state_grad_est_linearized_rpt} for the POMDPs.

Further hyperparameters are reported in Tables~\ref{tab:lunar_lander_manual_hyperparameters}, \ref{tab:lunar_lander_mdp_hyperparameters}, and \ref{tab:lunar_lander_pomdp_hyperparameters}. Performance results and runtimes are reported in Tables~\ref{tab:lunar_lander_mdp_complete} and \ref{tab:lunar_lander_pomdp_complete}.